\documentclass[iicol,sn-apa]{sn-jnl}

\usepackage{graphicx}
\usepackage{multirow}
\usepackage{amsmath,amssymb,amsfonts}
\usepackage{amsthm}
\usepackage{mathrsfs}
\usepackage[title]{appendix}
\usepackage{xcolor}
\usepackage{textcomp}
\usepackage{manyfoot}
\usepackage{booktabs}
\usepackage{algorithm}
\usepackage{algorithmicx}
\usepackage{algpseudocode}
\usepackage{listings}

\usepackage{times}
\usepackage{epsfig}
\usepackage{graphicx}
\usepackage{amsmath}
\usepackage{amssymb}
\usepackage{xcolor}
\usepackage{wrapfig,lipsum,booktabs}
\usepackage{multirow}

\usepackage{subcaption}
\usepackage{verbatim}

\usepackage{pifont}
\usepackage{array}  

\usepackage{color, colortbl}
\usepackage{overpic}
\usepackage{lmodern}
\usepackage{anyfontsize}

\newcommand{\cmark}{\ding{51}}
\newcommand{\xmark}{\ding{55}}

\newcolumntype{x}[1]{
>{\centering\hspace{0pt}}p{#1}}

\def\eg{\emph{e.g}} 
\def\ie{\emph{i.e}}

\begin{document}

\title[Article Title]{{Active Stereo in the Wild through Virtual Pattern Projection}}

\author*[1,2]{\fnm{Luca} \sur{Bartolomei}}\email{luca.bartolomei5@unibo.it}

\author[1,2]{\fnm{Matteo} \sur{Poggi}}\email{m.poggi@unibo.it}

\author[2]{\fnm{Fabio} \sur{Tosi}}\email{fabio.tosi5@unibo.it}

\author[2]{\fnm{Andrea} \sur{Conti}}\email{andrea.conti35@unibo.it}

\author[1,2]{\fnm{Stefano} \sur{Mattoccia}}\email{stefano.mattoccia@unibo.it}

\affil[1]{\orgdiv{Advanced Research Center on Electronic System (ARCES)}}

\affil[2]{\orgdiv{Department of Computer Science and Engineering (DISI)}}

\affil[]{\orgname{University of Bologna}, \orgaddress{\street{Viale del Risorgimento 2}, \city{Bologna}, \postcode{40136}, \country{Italy}}}

\abstract{
This paper presents a novel general-purpose guided stereo paradigm that mimics the active stereo principle by replacing the unreliable physical pattern projector with a depth sensor.
It works by projecting virtual patterns consistent with the scene geometry onto the left and right images acquired by a conventional stereo camera, using the sparse hints obtained from a depth sensor, to facilitate the visual correspondence. Purposely, any depth sensing device can be seamlessly plugged into our framework, enabling the deployment of a virtual active stereo setup in any possible environment and overcoming the severe limitations of physical pattern projection, such as the limited working range and environmental conditions. Exhaustive experiments on indoor and outdoor datasets featuring both long and close range, including those providing raw, unfiltered depth hints from off-the-shelf depth sensors, highlight the effectiveness of our approach in notably boosting the robustness and accuracy of algorithms and deep stereo without any code modification and even without re-training. Additionally, we assess the performance of our strategy on active stereo evaluation datasets with conventional pattern projection. Indeed, in all these scenarios, our virtual pattern projection paradigm achieves state-of-the-art performance. 
The source code is available at: \url{https://github.com/bartn8/vppstereo}.
}


\maketitle

\section{Introduction}

Depth perception perception is pivotal in several downstream computer vision tasks, such as autonomous driving, robotics, and augmented reality. For this reason, this topic has received longstanding attention in the literature and industry through the deployment of different technologies. Nonetheless, achieving this goal with cameras has many advantages due to the low cost and potentially unbounded image resolution. Consequently, different setups exist to infer depth uniquely from standard imaging devices. At the core of them, either using multiple cameras or a moving one, there is the problem of determining the visual correspondence across images. However, this task turns intrinsically ambiguous, especially when dealing with textureless regions, repetitive patterns, and non-Lambertian materials. Moreover, the difficulty increases significantly when the task is performed for each image pixel. Despite these facts, as reported in the literature, learning-based techniques achieve compelling performance \citep{poggi2021synergies,STEREO_SURVEY_BENNAMOUN}. However, a severe downside is that these methods are prone to \textit{domain shift} issues that are, on the contrary, not present in traditional yet less accurate hand-crafted methods. Specifically, their dependency on training data yields significant performance deterioration when dealing with different data distributions, such as deploying a network trained with outdoor data in indoor environments or vice versa. Hence, fine-tuning the network in the target domain is crucial to adapt it to the new environment. Unfortunately, collecting new data in the target domain requires a significant effort. Moreover, since collecting training data is expensive, unpractical and time-consuming, the massive amount of labeled training data needed by deep networks is typically obtained by exploiting computer graphic frameworks to gather synthetic data and then fine-tuning in the target domain with a much smaller amount of real data, with the previously outlined challenge to cope with the synthetic-to-real domain issue.

\begin{figure*}
    \centering
    \includegraphics[trim=0cm 23cm 18cm 0cm, clip,width=1.0\linewidth]{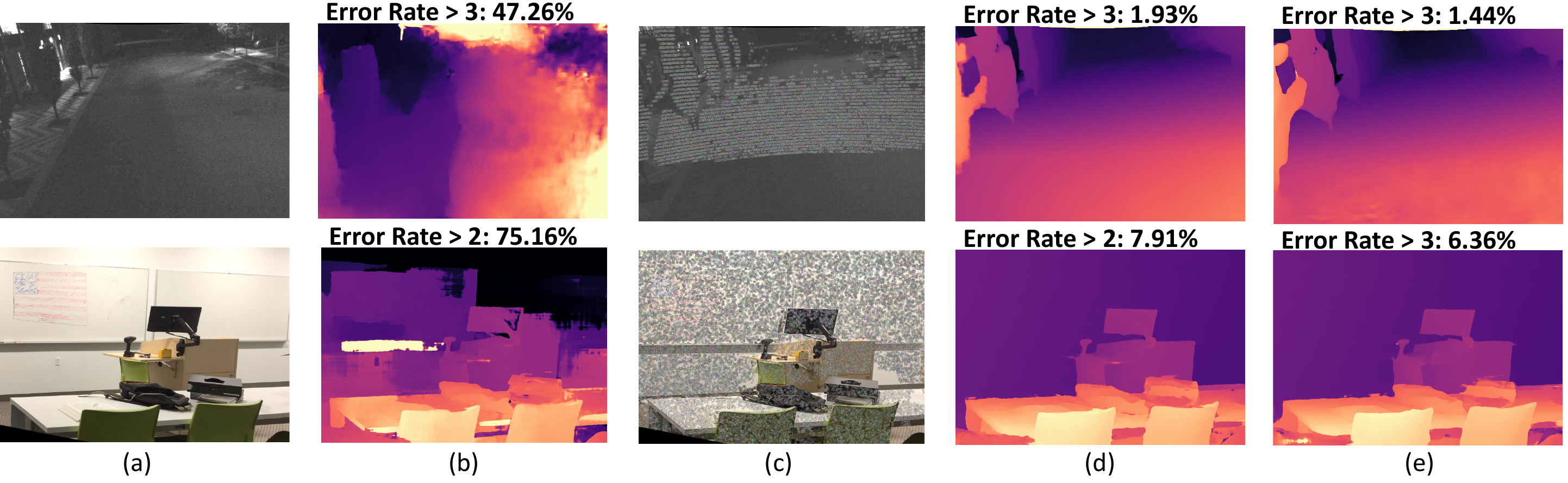}
   \captionof{figure}{\small\textbf{Virtual Pattern Projection for deep stereo.} 
   Either in challenging outdoor (top) or indoor (bottom) environments (a), stereo networks like RAFT-Stereo \citep{lipson2021raft} (top) or PSMNet \citep{chang2018psmnet} (bottom) struggle (b). By projecting a virtual pattern on images (c), the same networks dramatically improve their accuracy without retraining (d). Training the models to deal with the augmented images (e) further reduces errors.}
  \label{fig:teaser}
\end{figure*}

{An alternative strategy to infer depth relies on active sensing technologies, such as LiDAR (Light Detection and Ranging), ToF (Time of Flight), Radar (Radio Detection and Ranging), and structured or unstructured light devices. 
However, each of these technologies has specific limitations. LiDAR technology \citep{qiao2024survey} is quite reliable but features a much lower density than cameras, and increasing density is extremely expensive and challenging due to technological issues. ToF \citep{qiao2024survey} has similar limitations and is also ineffective in sunlight and at longer distances. Radar \citep{Radar_Monocular} is another popular active technology massively deployed in the automotive field to infer depth by measuring the time it takes to send and receive electromagnetic waves. Although it allows comparable range sensing, it provides sparser and noisier depth data and has a smaller vertical field of view.}
Finally, we mention active systems estimating depth from images; they rely on a structured \citep{Tutorial_structured_light} or unstructured \citep{INTEL_REAL_SENSE_UNSTRUCTURED} pattern projection to increase the robustness of pure passive-based sensing systems. Similarly to the latter approaches, they enable higher resolution than active sensors. However, their strength is also their main weakness, since they require a projector to ease visual correspondence with appropriate patterns projected into the sensed scene. Consequently, they struggle when the artificially injected visual cues are not visible, such as when the target distance falls just a few meters away or under sunlight. Moreover, since pattern projection typically occurs in the IR spectrum, they require at least one additional RGB camera. Finally, multiple projectors might interfere with each other, especially using structured light projection systems.

The complementary strengths and weaknesses of active and passive technologies mentioned above led to their joint deployment in many fields to improve depth perception. For instance, almost all autonomous driving prototypes leverage heterogeneous sensor suites, including cameras and LiDARs. Similarly, most smartphones/tablets and AR devices, such as Apple Vision Pro and Meta Quest Pro, have cameras coupled with ToF depth sensors. For these reasons, researchers have proposed strategies to exploit the synergy between active and passive depth sensing \citep{poggi2019guided,LIDARSTEREONET,wang20193d}. The shared main trait of these sensor-fusion methods consists of modifying their internal behaviour to account for the sparse depth seeds and to guide them to a better solution.

In this paper, we follow a completely different path, initially proposed in \citep{Bartolomei_2023_ICCV}, to exploit the synergy between passive and active sensing by acting directly on the vanilla input images acquired by the cameras before any processing occurs. Specifically, to ease the visual correspondence problem exploiting the availability of sparse depth seeds, we coherently hallucinate the stereo pair acquired by a conventional camera to facilitate the visual correspondence task performed by any stereo network/algorithm, as if a \textit{virtual} pattern projector were present in the scene. {Our hallucination procedure is an entirely hand-crafted algorithm, thus eliminating any possible domain shift issues within the virtual projector.} Fig. \ref{fig:teaser} showcases the improvements yielded by our proposal. As for sensor fusion methods in the literature  \citep{poggi2019guided,LIDARSTEREONET}, the only requirements are a calibrated stereo rig and the sparse yet accurate depth measurements registered with the reference camera. However, in contrast to other sensor fusion approaches, the motivation behind our proposal shares similarities with active methods based on unstructured pattern projections \citep{KONOLIGE_ACTIVE_STEREO,INTEL_REAL_SENSE_UNSTRUCTURED}, though implemented through a completely different setup and paradigm. Neglecting a physical pattern projector in favour of a depth sensor eliminates all the severe limitations mentioned above by simply selecting the device best suited for the specific target application domain. 

An exhaustive experimental evaluation with various stereo datasets, including those featuring the availability of raw data as gathered by off-the shelf depth sensors without any filtering, supports the following claims:

\begin{itemize}
    \item Even with a meager number of sparse depth seeds (e.g., less than 1\% of the whole image content), our virtual pattern paradigm outperforms state-of-the-art sensor fusion methods by a large margin;
    \item As shown in Fig. \ref{fig:teaser}, it has a compelling ability to tackle domain shift issues, even with networks trained only with synthetic data and without additional training or fine-tuning;
    \item Acting before any processing occurs, it can be seamlessly deployed with any stereo algorithm or deep network without modifications and benefit from future progress in the field.
\end{itemize}

\begin{figure*}
    \centering
    \includegraphics[width=0.95\linewidth]{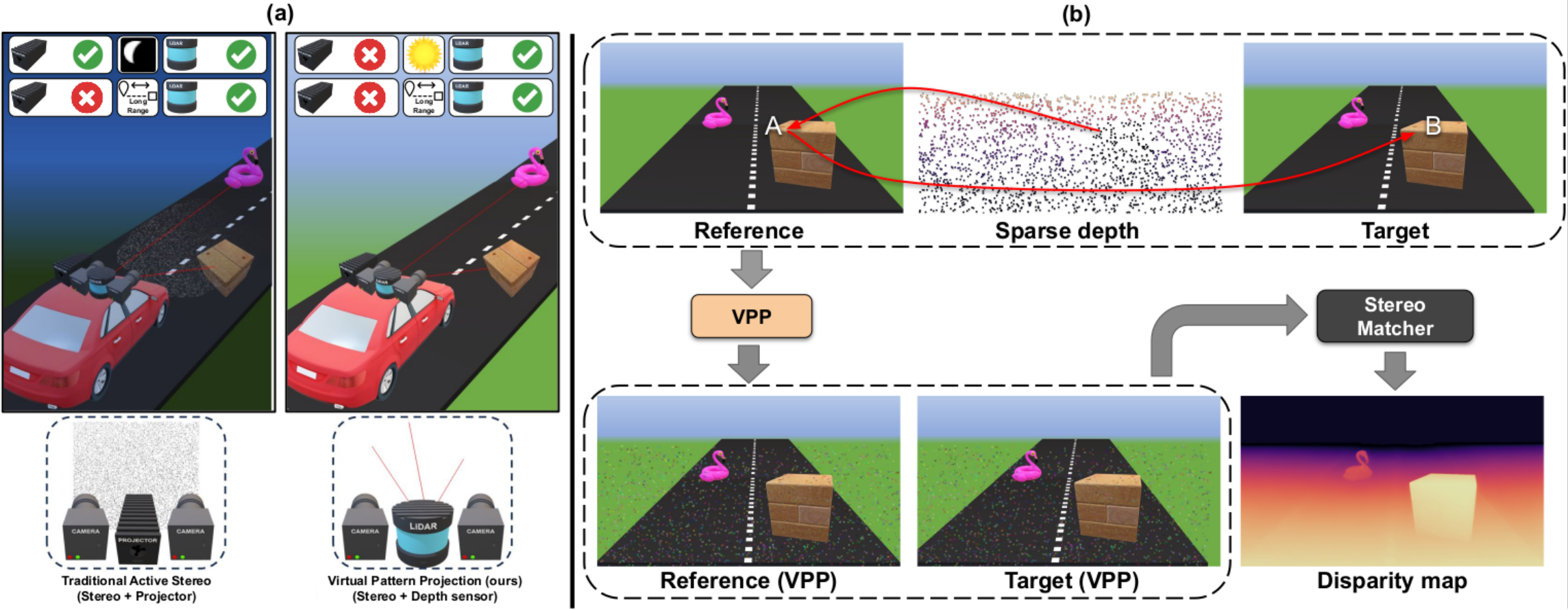}
    
    \caption{\textbf{Overview of the Virtual Pattern Projection framework.} \textbf{(a)} Unlike conventional active stereo systems, where the physically projected patterns are visible only in certain light conditions and within a limited distance, our setup overcomes these constraints using a standard stereo camera and a depth sensor. \textbf{(b)} The Virtual Pattern Projection (VPP) paradigm coherently hallucinates the vanilla stereo pair acquired by a standard camera exploiting the sparse depth seeds gathered by a depth sensor. The fictitiously patterned stereo pair in output can be seamlessly processed by any stereo matcher, either handcrafted or learning-based.} 
    \label{fig:VPP_method}
\end{figure*}

    Moreover, in contrast to active stereo systems, using a depth sensor in place of a pattern projector: 
\begin{itemize}
    \item As depicted in Fig. \ref{fig:VPP_method} (a), it can work under sunlight, indoors and outdoors, at long and close ranges, and without any additional processing cost for the selected stereo matcher; 
    \item It is more effective even in the specific application domain of projector-based systems and potentially less expensive;       
    \item It does not require additional hardware (e.g., additional RGB or IR cameras), as depth estimation is performed in the same target visual spectrum; 
    \item The virtual projection paradigm can be tailored on the fly to adapt to the image content and is agnostic to dynamic objects and ego-motion.
\end{itemize}

For the reasons mentioned above, we believe that our \textit{Virtual Pattern Projection} (VPP) paradigm has the potential to become a standard component for depth perception and pave the way to foster exciting progress in the field. 

\section{Related Work}

This section summarizes the literature relevant to our work concerning passive and active stereo and methods to infer depth jointly using cameras and sparse depth seeds.

\textbf{Stereo Matching.} Conventional stereo algorithms, deeply investigated in the past decades and thoroughly surveyed by \cite{scharstein2002taxonomy}, infer dense disparity maps from stereo pairs mainly using priors and handcrafted features \citep{Secaucus_1994_ECCV, veksler2005stereo, yang2008stereo, yang2010constant, liang2011hardware, taniai2014graph, kolmogorov2004energy, hirschmuller2007stereo, boykov2001fast}. Nonetheless, learning-based approaches have profoundly modified the traditional paradigm, enabling them to outperform previous methods by a large margin starting from pioneering work \citep{zbontar2016stereo} that initially acted on a single phase of conventional stereo pipelines. Eventually, more sophisticated end-to-end stereo networks became the most effective solution for disparity estimation. These networks, thoroughly reviewed by \cite{poggi2021synergies} and \cite{STEREO_SURVEY_BENNAMOUN}, fall into 2D and 3D architectures. The former relies on an encoder-decoder design \citep{mayer2016large, Pang_2017_ICCV_Workshops, Liang_2018_CVPR, saikia2019autodispnet, song2018edgestereo, yang2018segstereo, yin2019hierarchical, Tankovich_2021_CVPR} inspired by U-Net \citep{ronneberger2015u}. The latter creates a cost volume from features extracted by the image pair and estimates disparity through 3D convolutions \citep{Kendall_2017_ICCV, chang2018psmnet, khamis2018stereonet, zhang2019ga, cheng2019learning, cheng2020hierarchical,  duggal2019deeppruner, yang2019hierarchical, wang2019anytime, guo2019group, Shen_2021_CVPR}, at the cost of a much higher memory and runtime requirements. More recent deep stereo networks \citep{lipson2021raft, li2022practical,zhao2023high,zeng2023parameterized,xu2023iterative,wang2024selective} exploit the RAFT \citep{teed2020raft} iterative refinement paradigm initially proposed for optical flow, or Vision Transformers \citep{li2021revisiting,guo2022context} to catch broader contextual information from images. Unfortunately, to achieve compelling performance, learning-based methods need massive annotated data for training and yield limited generalization capabilities with out-of-domain distributions \citep{zhang2019domaininvariant,cai2020matchingspace,aleotti2021neural,zhang2022revisiting,liu2022graftnet,chuah2022itsa,watson2020stereo, Tosi_2023_CVPR}. To mitigate these issues, self-supervised techniques allow to train deep stereo models without ground-truth annotations leveraging photometric losses on stereo pairs or videos \citep{SsSMnet2017,Tonioni_2019_CVPR,Tonioni_2019_learn2adapt, lai2019bridging,wang2019unos,chi2021feature}, traditional algorithms and confidence \citep{CONFIDENCE_PAMI} measures  \citep{Tonioni_2017_ICCV, aleotti2020reversing}. Other methods perform continuous self-supervised adaptation from input images to cope with domain-shift issues \citep{Tonioni_2019_CVPR, Poggi2021continual, Poggi_2024_CVPR}.

\textbf{Active Stereo Matching.} 
Active stereo systems rely on a projector, typically operating in the IR spectrum, to infer depth by injecting patterns onto the sensed scene employing two principles: structured \citep{Tutorial_structured_light} and unstructured \citep{INTEL_REAL_SENSE_UNSTRUCTURED,KONOLIGE_ACTIVE_STEREO} light.
According to different strategies, former methods work by analyzing the behaviour of known patterns encoding depth in a single shot \citep{Kinect_review_PAMI,CONNECTING_THE_DOTS, GIGADEPTH,ACTIVE_STEREO_FANELLO} or by temporal projection of known patterns \citep{INTEL_REAL_SENSE_STRUCTURED_PAMI}. However, temporal projection is problematic when dealing with moving objects or camera ego-motion; thus, a much higher frame rate is sometimes used to soften this constraint \citep{INTEL_REAL_SENSE_STRUCTURED_PAMI}.
In active systems that leverage structured light, depth triangulation typically relies on the pattern projector serving as one of the two camera viewpoints, although some exceptions exist \citep{scharstein2014high,liu2023joastereo}. Due to its relevance, methods to optimize the projected patterns were proposed \citep{PATTERN_OPTIMIZAtION_2018,PATTERN_OPTIMIZAtION_ECCV_2018, PATTERN_OPTIMIZAtION_2020}.
Conversely, unstructured light methods rely on a stereo camera and take advantage of the additional texture injected into the scene through a random pattern projection \citep{grunnet2018projectors}, sometimes conceived beforehand to increase local distinctiveness \citep{KONOLIGE_ACTIVE_STEREO}, to enrich the scene and thus ease stereo matching, especially in challenging regions with ambiguous/repetitive features or lacking distinctive details \citep{COMPARATIVE_ACTIVE_SENSING_SHORT_RANGE}.
On the other hand, most of unstructured light proposals infer depth in a single shot \citep{zhang2018activestereonet,INTEL_REAL_SENSE_UNSTRUCTURED,KONOLIGE_ACTIVE_STEREO,ACTIVE_STEREO_FANELLO}, but, assuming static scenes, some methods can increase accuracy when exposed to a temporal pattern \citep{zhang2003spacetime,BOOSTER_PAMI,ACTIVE_STEREO_CVPR_2022}.
Regardless of the working principle, pattern projection is constrained to shorter distances up to a few meters away and unfeasible under sunlight. Moreover, methods based on structured projection require carefully designed projectors suffering thermal drifts \citep{KINECT_THERMAL} among other issues \citep{KINECT_1_VS_2}, and multiple projectors in the same area interfere. Despite these severe limitations, enriching the original visual content with artificial texture yields dramatic improvement over passive stereo. Consequently, active stereo is a lively research field \citep{ACTIVE_STEREO_FANELLO,zhang2018activestereonet,ACTIVE_STEREO_CVPR_2022,liu2023joastereo}.

\textbf{Image-Guided Methods.} An alternative strategy for depth estimation aims to integrate sparse depth seeds gathered from an active sensor with visual information. According to this principle, two main strategies exist in the literature.  

\textit{Depth Completion.} It aims to estimate dense depth maps at the resolution of the single RGB camera in the setup, exploiting the visual content and sparse depth seeds gathered by a depth sensor. Many methods have been proposed for this purpose, ranging from traditional methods based on interpolation and optimization \citep{camplani2012efficient,shen2013layer,lu2014depth}, to learning-based approaches \citep{ma2019self, cheng2018depth, park2020non, hu2021penet, chen2019learning, lin2022dynamic, zhang2023completionformer}. However, despite the compelling results, these approaches suffer in areas without depth hints, are sensitive to the number of depth seeds used for training, and generalize poorly to out-of-domain data distributions. Purposely, \cite{conti2023wacv} proposed a method to deal with the first two mentioned issues by relying more heavily on the image content. In contrast, \cite{bartolomei2024revisiting} faced domain-shift issues by casting depth completion as a fictitious stereo correspondence problem deploying the virtual projection paradigm proposed in this paper.

\textit{Guided Stereo Matching.} Sharing the motivation of the previous strategy, they aim to infer depth by fusing sparse depth seeds gathered by active sensors with registered stereo pairs exploiting the complementary strengths of the two sensing modalities. In particular, sparse depth hints can provide additional cues that complement stereo and improve robustness in challenging regions, especially those featuring poor texture. 
In early works, \cite{Badino2011IntegratingLI} use dynamic programming to integrate LiDAR data into a conventional stereo algorithm, while \cite{gandhi2012high} propose fusing ToF data and stereo pairs with an efficient seed-growing algorithm. In contrast, more recent works exploit sparse depth seeds either by concatenating them as input to CNN-based architectures \citep{LIDARSTEREONET,choe2021volumetric, park2018high, zhang2020listereo,xu2023expanding} or by guiding cost aggregation from existing cost volumes \citep{wang20193d, poggi2019guided, huang2021s3, zhang2022lidar} which require modifications to the internal structure of the networks.
In contrast, our proposal follows an entirely different direction by acting outside the network before processing images. We increase the match distinctiveness, thus facilitating the visual correspondence of any stereo method by coherently hallucinating the input vanilla stereo pair with virtual patterns generated according to the sparse depth seeds provided by a depth sensor before feeding them to an unmodified stereo matcher.

\section{Virtual Pattern Projection (VPP)}

Densely matching points across images is an inherently challenging task at the core of many vision problems. The strategy adopted by active stereo is potentially disruptive, but its dependence on a physical pattern projector severely limits its practical effectiveness. In light of this evidence, given a setup with a stereo camera and the availability of sparse depth seeds registered with the reference frame of the rig, our proposal aims to facilitate visual correspondence by virtually hallucinating the vanilla input images coherently with the sparse depth information. In contrast to projector-based systems, it removes all the mentioned limitations and works regardless of the technology used to infer such depth points. Nonetheless, in the most typical current embodiment, these points come from an active depth sensor -- including cheaper devices providing a modest yet accurate number of depth seeds suffices (e.g., even less than 1\% of the whole image content).

\subsection{Virtual Projection Principle}

Despite having the same goal and processing patterned images, our method initially reverses the basic unstructured projection principle of active stereo to compensate for the missing physical projection using instead the input sparse depth points to hallucinate the vanilla stereo pair coherently. Specifically, our crucial observation is that, given a rectified stereo rig and a pool of sparse depth points registered with the reference image of the stereo rig, each of these depth points implicitly locates two corresponding pixels in the reference and target image of the stereo camera. Relying on this evidence, we coherently augment the visual appearance of both pixels in the two images, hallucinating them identically and making them as distinctive as possible from their neighbours. This novel paradigm allows us to conceive an imaginary setup with a real stereo camera and a fictitious projector capable of projecting consistent patterns onto the input vanilla stereo pair where depth seeds are available. 
Our key insight is that, even when applied very sparsely, our peculiar virtual image patterning can significantly enhance matching by enabling more meaningful local and global reasoning over the entire image. Besides, by selecting a depth sensor or any other means to infer reliable sparse seeds suited for the target sensing environment, we eliminate the shortcomings of active stereo based on unstructured light while retaining its advantages. Additionally, our proposal enables processing in the same visual target domain, hence not requiring additional cameras.  

Fig. \ref{fig:VPP_method} (b) outlines the central idea behind our virtual projection principle. The envisioned setup consists of a calibrated stereo camera and a depth sensor providing sparse yet accurate depth points registered with the stereo reference camera through an initial calibration. 
We observe that through the known stereo camera geometry, the depth $z(x,y)$ of each input seed, turned in a disparity value, implicitly locates two corresponding points $I_{L}(x,y)$ and $I_{R}(x',y)$ in the input images. Specifically, by knowing the focal length $f$ and the baseline $b$ of the stereo camera, the disparity corresponding to $z(x,y)$ can be obtained as $d(x,y) = \frac{b \cdot f}{z(x,y)}$ \citep{SZELISKI_BOOK}. Given a pixel $I_{L}(x,y)$, such a disparity value represents the offset to obtain the position along the same epipolar line of the corresponding point $I_{R}(x',y)$ in the target image with $x' = x - d(x,y)$. Fig. \ref{fig:VPP_method} (b) illustrates this evidence, showing how two corresponding pixels A, B in the two input images can be located by knowing the depth of A. Relying on this, we can coherently hallucinate the two vanilla input images where depth seeds are available to facilitate stereo correspondence. 

\subsection{Virtual Patterning}

For our purposes, a proper image hallucination should render corresponding points identical. Accordingly, we propose two augmenting strategies named 
\textit{random} and \textit{histogram-based} pattern projection. Both operate on corresponding points $(x,y)$ and $(x',y)$ located along the epipolar line as outlined before, applying the same pattern operator $\mathcal{P}(x,x',y)$ onto the two images:  

\begin{equation}
    \begin{split}
        I_L(x,y) \leftarrow \mathcal{P}(x,x',y)\\
        I_R(x',y) \leftarrow \mathcal{P}(x,x',y)
    \end{split} 
    \label{eq:stereo_pattern}
\end{equation}

The difference between the two strategies is how the operator $\mathcal{P}(x,x',y)$ generates the virtual pattern superimposed to input images. Fig. \ref{fig:Duck_pattern} outlines the possible patterns we will discuss in the remainder.
Moreover, since $x'$ is unlikely to be an integer, we apply $\mathcal{P}(x,x',y)$ to $I_R(\lfloor x'\rfloor,y)$ and $I_R(\lceil x'\rceil,y)$ using a weighted splatting according to $\beta = x'-\lfloor x'\rfloor$.

\begin{equation}
    \begin{split}
        I_R(\lfloor x'\!\rfloor,y) \leftarrow \beta I_R(\lfloor x'\!\rfloor,y) + (1-\beta) \mathcal{P}(x,x'\!\!,y) \\
        I_R(\lceil x'\!\rceil,y) \leftarrow (1-\beta) I_R(\lceil x'\!\rceil,y) + \beta \mathcal{P}(x,x'\!\!,y)
    \end{split}
    \label{eq:stereo_pattern_splatting}
\end{equation}

\subsubsection{Random Patterning}

To better face visual correspondence, the patterns should increase similarity across corresponding points in the two images and improve local distinctiveness \citep{Distinctiveness_ICCV_07,Manduchi1999Distinctiveness}, at least along epipolar lines. 
Consequently, a pattern like the one in Fig. \ref{fig:Duck_pattern} (i) would be suboptimal.
Hence, to better face this issue, a first method uses an operator $\mathcal{P}(x,x',y)$ that projects a random value picked from a uniform distribution $\mathcal{U}$ as follows: 

\begin{equation}
    \mathcal{P}(x,x',y)\sim\mathcal{U}(0,255)
    \label{eq:method_3}
\end{equation}

Fig. \ref{fig:Duck_pattern} (ii) reports a possible outcome of this strategy whose extension to multi-channel images, as for any method proposed, is straightforward by applying the same patterning strategy on each channel. 
This strategy might generate more meaningful patterns with an almost negligible computing overhead. However, inherently, it does not grant a complete fulfilment of distinctiveness requirements, as depicted on the head and a portion of the background Fig. \ref{fig:Duck_pattern} (ii). In contrast, a patterning operator that analyses the image content could constantly improve distinctiveness.

\subsubsection{Histogram-based Patterning}

\begin{figure}
    \centering
    \includegraphics[trim=4.2cm 8.5cm 8cm 2.1cm, clip, width=0.9\linewidth]{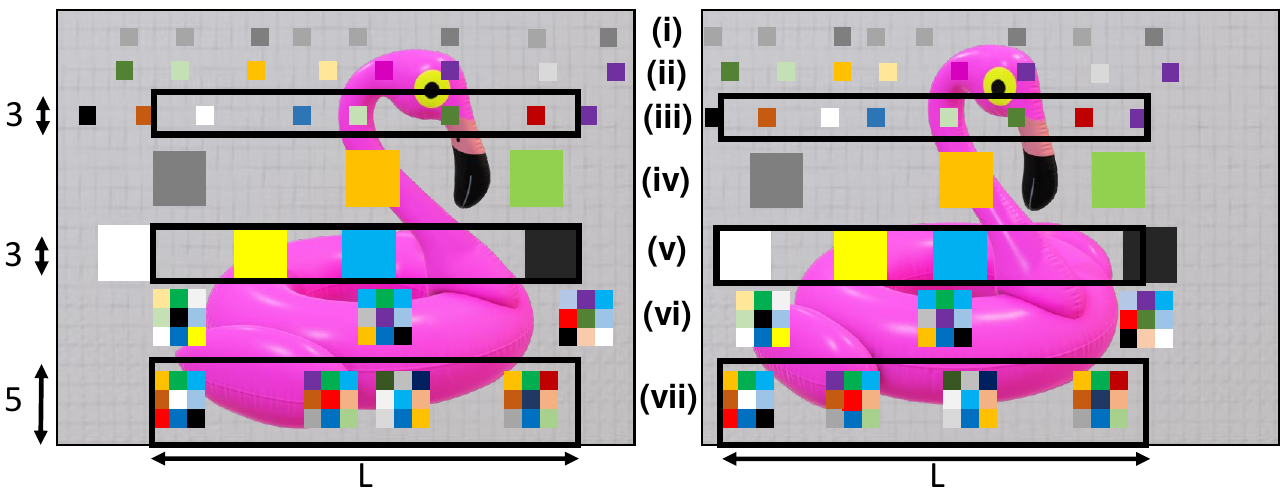}
    \caption{\textbf{Virtual patterns.} 
    From top to bottom: (i) indistinctive, (ii) randomly generated, (iii) distinctive, (iv) randomly generated patch-based uniform, (v) distinctive patch-based uniform, (vi) randomly generated patch-based, and (vii) distinctive patch-based patterns. In all experiments, we use a search area of length $L$, set to 64 and height $(2+N)$ where $N$ represents the vertical size of the virtual patches.
    }
    \label{fig:Duck_pattern}
\end{figure}

In light of the previous discussion, patterns superimposed onto images should stand out from the background and be unambiguous within nearby pixels along the same horizontal scanline, as shown in Fig. \ref{fig:Duck_pattern} (iii). 

Purposely, for each new pixel needing a virtual pattern, we use a histogram-based operator $\mathcal{P}(x,x',y)$, to select the optimal pattern by analyzing the scanline content in the two images subject to hallucination. Specifically, for a pixel $(x,y)$ in the reference image, we consider two windows of height 3 and length $L$ centred on it and on $(x',y)$ in the target image. Then, the histograms computed over the two areas are merged into a single one and the operator $\mathcal{P}(x,x',y)$ picks from it the value that maximizes the distance from any other entry in the histogram $\texttt{hdist}(i)$, with $\texttt{hdist}(i)$ returning the minimum distance from a filled bin in the sum histogram $\mathcal{H}$

\begin{equation}
    \begin{split}
        \texttt{hdist}(i) = & \big\{\min\{ |i-i_l|,\,|i-i_r| \},\\
        & i_l\in[0,i[:\mathcal{H}(i_l)>0,\\
        & i_r\in]i,255]:\mathcal{H}(i_r)>0 \big\}
    \end{split}
    \label{eq:method_2_1}
\end{equation}
When every bin in $\mathcal{H}$ is not empty, we select the value with the minimum occurrences.

\subsection{Advanced Virtual Patterns}

We will now extend the strategies described so far to account for the image content, spatial locality, distance, and occlusions.

\subsubsection{Blending image content and virtual patterns}

Although a virtual pattern facilitates matching for traditional algorithms \citep{hirschmuller2007stereo}, it might hamper a deep stereo network not used to deal with such data.
Hence, to soften this issue, we merge the image content with virtual patterns using an alpha-blending \citep{SZELISKI_BOOK} tuned according to a hyperparameter $\alpha \in [0,1]$:    

\begin{equation}
    \begin{split}
        I_L(x,y) \leftarrow (1-\alpha) I_L(x,y) + \alpha \mathcal{P}(x,x',y)\\
        I_R(x',y) \leftarrow (1-\alpha) I_R(x',y) + \alpha \mathcal{P}(x,x',y)
    \end{split}
    \label{eq:stereo_pattern_blending}
\end{equation}

\subsubsection{Spatial Locality}

The pointwise patterning strategy proposed can be extended to nearby points to further enrich the overall visual appearance. To this aim, we apply previous methods to patches (e.g., $3 \times 3$, $5 \times 5$), implicitly assuming that the disparity within these areas does not change. 
Fig. \ref{fig:Duck_pattern}, in (iv) and (v), depicts a possible outcome of the operator $\mathcal{P}(x,x',y)$ acting on a region of length $L$, that picks a uniform color within patches according to, respectively, random sampling or a histogram based selection. 
Nonetheless, we can generate additional and more distinctive patterns through (vi) random sampling or (vii) and a histogram-base selection being performed for every pixel in the patch independently, as depicted in Fig. \ref{fig:Duck_pattern}.
However, since naively setting a constant size squared window might be sub-optimal, we propose two additional strategies in the remainder.

\begin{figure}[t]
    \centering
    \includegraphics[width=0.9\linewidth]{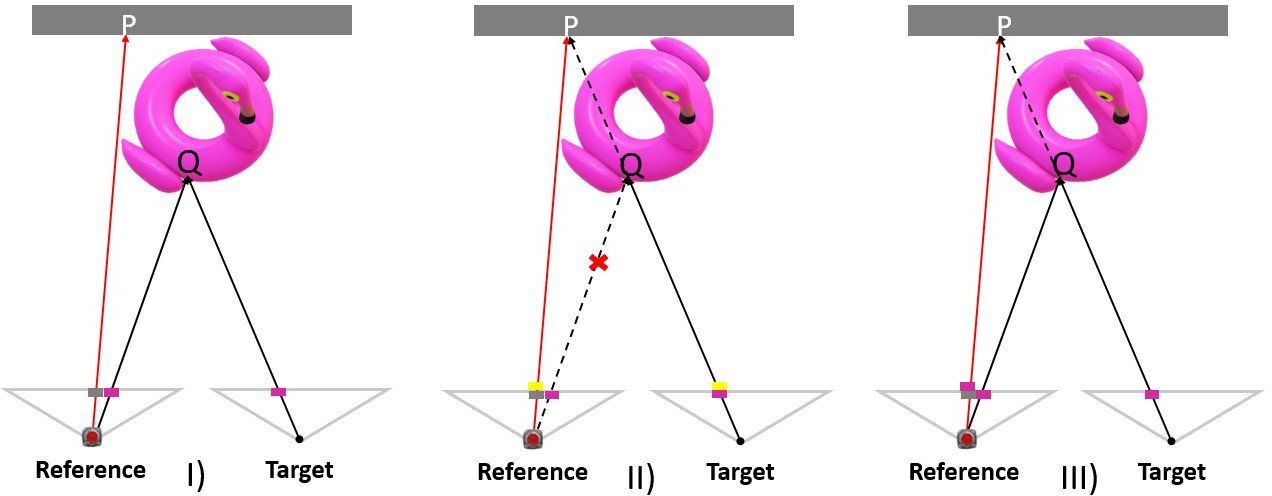}
    \caption{\textbf{Handling occlusions.} I) P is framed by the reference camera and depth sensor but occluded in the target camera (``NO" projection), II) Projection of the same pattern (yellow) onto the two input images according to depth in the background (``BKGD" projection), III) Projection of the foreground image content (Q) from the target image to the background in the reference image (``FGD" projection).}
    \label{fig:Duck_occlusion}
\end{figure}

\textbf{Distance-based patterning} Since things appear smaller in the images at farther distances, a large virtual patch might cover most of a small object when it is far from the camera, preventing the stereo algorithm from inferring its possibly different disparity values.
To deal with this issue, we propose a dynamic patch size that exploits the available input sparse depth seeds. Considering a disparity hint $d(x,y)$, a disparity search range $\left[ D_\text{min}, D_\text{max} \right]$, and a maximum patch size $N_\text{max}$, we adjust the patch size $N(x,y)$ as follows:

\begin{equation}
    N(x,y) = \left\lfloor \left(\frac{d(x,y)-D_\text{min}}{D_\text{max}-D_\text{min}}\right)^{\frac{1}{\phi}} \cdot \left( N_\text{max}-1 \right)  + 1  \right\rceil
    \label{eq:patch_distance}
\end{equation}
where $\lfloor\cdot\rceil$ is the rounding operator, $\phi$ models the mapping curve and $D_\text{min}, D_\text{max}$ are given respectively by the nearest and the farthest hint in the frame.

\textbf{Adaptive patch.} 
Regardless of the distance from the camera, a fixed patch shape becomes problematic near depth discontinuities and with thin objects, although widely used in stereo and other tasks \citep{SZELISKI_BOOK}. 

As detailed next, following \cite{bartolomei2024revisiting}, we adapt the patch shape according to the reference image content to address this challenge.

Given a local window $\mathcal{N}(x,y)$ centered around the disparity hint $d(x,y)$, for each pixel $(u,v)\in\mathcal{N}(x,y)$ we estimate the spatial $S(x,y,u,v)=(x-u)^2+(y-v)^2$ and color $C(x,y,u,v)=\left|I_L(u,v)-I_L(x,y)\right|$ agreement with the reference hint point $(x,y)$ and feed them into model $W_\text{c}(x,y,u,v)$:

\begin{equation}
    W_\text{c}(x,y,u,v)=\exp\left(\frac{S(x,y,u,v)}{-2\sigma_s^2}+\frac{C(x,y,u,v)}{-2\sigma_c^2}\right)
    \label{eq:adaptive_weight_color}
\end{equation}

where $\sigma_s$ and $\sigma_c$ control the impact of the spatial and color contribution as in a bilateral filter.  For each pixel $(u,v)$ within a patch centred in $(x,y)$, we apply the virtual projection only if $W(x,y,u,v)$ exceeds a threshold $t_w$.
Additionally, we store $W(x,y,u,v)$ values in a proper data structure to handle overlapping pixels between two or more patches -- \ie, we perform virtual projection only for the pixel with the highest score.

\subsubsection{Occlusions}

Even assuming a depth sensor perfectly aligned with the reference camera -- although not feasible \citep{Conti_confidence_IROS_2022} -- a stereo setup generates
occlusions by construction, given the different positions of the two cameras. Consequently, we cannot consistently project the pattern between the two views for these areas. Considering this evidence, detecting depth seeds in occluded regions is crucial to avoid projecting the same pattern on the occluded and occluder pixels on the reference and target image, as depicted in Fig. \ref{fig:Duck_occlusion} I).
To tackle this problem, we propose a simple yet effective heuristic to classify sparse input disparity hints warped onto the target image according to the difference in disparity and spatial distance from other input seeds.

\begin{figure}[t]
    \centering
    \includegraphics[trim=0cm 0cm 0cm 0.7cm,clip,width=\linewidth]{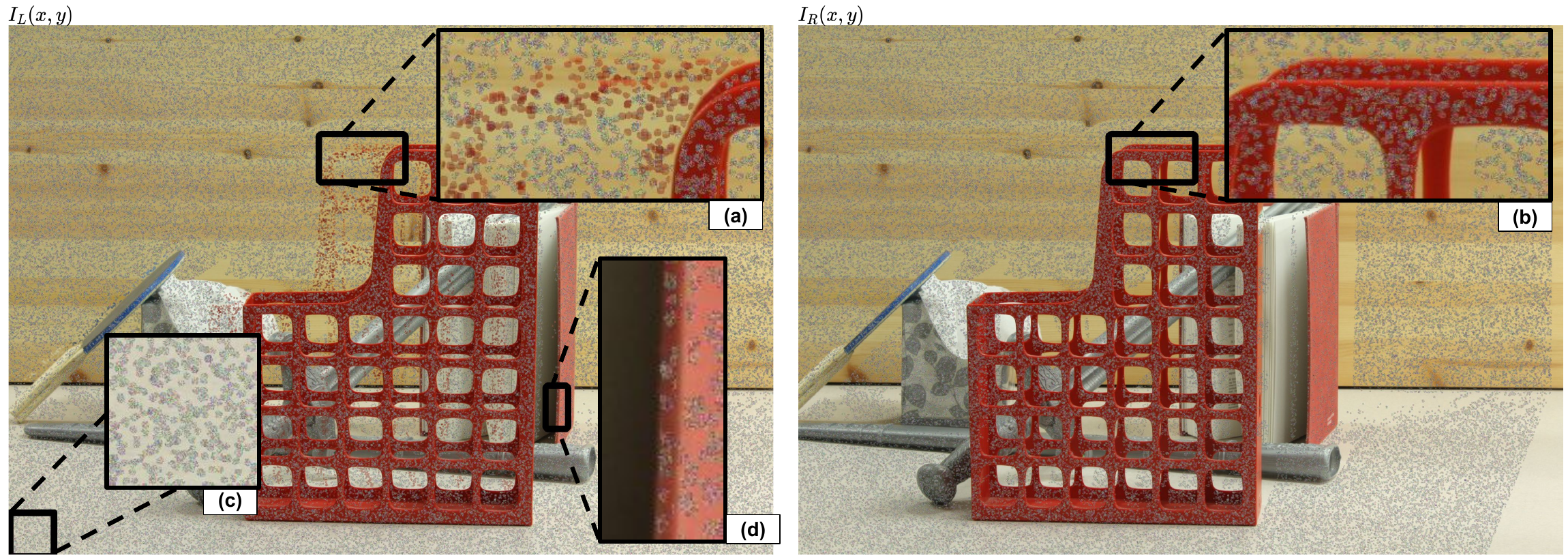}
    \caption{\textbf{VPP in action.} Hallucinated stereo pair using pattern (vi); zoomed-in view with FGD-Projection (a), corresponding area (b) with sub-pixel splatting and left border occlusion projection (c). Adaptive patches guarantee the preservation of depth discontinuities and thin details (d).}
    \label{fig:pattern_qualitative1}
\end{figure}

Specifically, we warp disparity $d(x,y)$ into an image-like grid $W$ at coordinates $(x',y)$; when multiple warped disparities collide at the exact location $(x',y)$, we keep the largest one.
Then, each $(x_o,y_o)$ in $W$ is classified as occluded if the following inequality holds for at least one neighbor $W(x,y)$ within an $r_x \times r_y$ patch:  
\begin{equation}
    W(x,y)-W(x_o,y_o) - \lambda (\gamma\lvert x-x_o \rvert + (1-\gamma) \lvert y-y_o \rvert ) > t    
    \label{eq:heuristic_maskocc}
\end{equation}
with 
$\lambda,\gamma,r_x,r_y,t$ hyper-parameters. Finally, we warp back occluded points to update the binary occlusion mask $\Omega$. 

When a disparity hint $(x,y)$ turns occluded ($\Omega(x,y)=1$), we could neglect projection on both images of the stereo pair (``NO" projection strategy). Hence, in the proximity of occlusions, this strategy can refrain from projecting the same pattern on the foreground (in the target image) and background (in the reference) -- as would occur with the strategy named ``BKGD" projection, Fig. \ref{fig:Duck_occlusion} II) -- to reduce ambiguity.
Nonetheless, to better exploit the few sparse seeds available at best, we follow a third strategy: we do not project when $\Omega(x,y)=1$ and instead replace the original content in $(x,y)$ on the reference image with the content at $(x',y)$ in the target image.
This approach -- named ``FGD" projection, Fig. \ref{fig:Duck_occlusion} III) -- does not alter the appearance of the correct foreground match (rays originating from Q), yet triggers the stereo matcher to search for a second correspondence with the same point $(x',y)$ in the target image, i.e., with pixel $(x,y)$ originating from P.   
Additionally, points close to the left border of the reference image would project patterns outside the target image. Although this fact would be irrelevant for traditional algorithms, we still project there (``FGD") to avoid artefacts in the predictions by deep stereo networks. Fig. \ref{fig:pattern_qualitative1} shows an example of a stereo pair hallucinated according to the ``FGD" strategy.

\begin{table}[t]
\centering
\renewcommand{\tabcolsep}{5pt}
\scalebox{0.75}{
\begin{tabular}{|ccc|ccc|}
\hline
&\multirow{2}{*}{Patch} & \multirow{2}{*}{Setting} & \multicolumn{3}{c|}{Error Rate (\%) $>2$} \\
& & & RAFT-Stereo 
& PSMNet 
& rSGM \\
\hline\hline
(A) & \xmark & No-VPP & 11.7 & 29.6 & 26.4\\
\hline
(B) &$3\times3$ & Fixed patch size 
& 5.2 & \bf 15.2 & 11.9 \\
\hline
(C) &$7\times7$ & Distance-based patch & 5.2 & \bf 15.2 & 12.6 \\
\rowcolor{yellow}
(D) &$7\times7$ & Bilateral based patch & \bf 4.9 & \bf 15.2 & \bf 11.0 \\
(E) &$7\times7$ & (C)+(D) & 5.2 & \bf 15.2 & 12.4 \\
\hline
\end{tabular}
}
\caption{
\textbf{Ablation study on adaptive patches.}
Results on Midd-A. Training on SceneFlow.
}
\label{tab:ablation_adaptive}
\end{table}

\section{Experimental Results}

This section describes our exhaustive experimental evaluation, including implementation details, datasets, and results analysis.

\begin{table*}
\centering
\renewcommand{\tabcolsep}{8pt}
\scalebox{0.58}{
\begin{tabular}{|l|cc|rrrr|r|rrrr|r|rrrr|r|}

\multicolumn{3}{c}{} & \multicolumn{5}{c}{Midd-14} & \multicolumn{5}{c}{Midd-21} & \multicolumn{5}{c}{ETH3D} \\ 
\hline
\multirow{2}{*}{Model} & \multicolumn{2}{c|}{Depth Points} & \multicolumn{4}{c|}{Error Rate (\%)} & avg. & \multicolumn{4}{c|}{Error Rate (\%)} & avg. & \multicolumn{4}{c|}{Error Rate (\%)} & avg. \\
 & Train & Test & $>1$ & $>2$ & $>3$ & $>4$ & (px) & $>1$ & $>2$ & $>3$ & $>4$ & (px) & $>1$ & $>2$ & $>3$ & $>4$ & (px) \\
 \hline\hline
rSGM 
& \xmark & \xmark 
& 44.82 & 32.00 & 27.21 & 24.48 & 13.19 
& 49.62 & 33.10 & 26.12 & 22.16 & 7.69 
& 16.58 & 8.02 & 5.35 & 4.12 & 0.93 \\
\hline
rSGM-{\em gd} 
& \xmark & \cmark 
& 32.47 & 18.80 & 14.64 & 12.77 & 9.44 
& 34.94 & 17.27 & 11.7 & 9.34 & 4.05 
& 7.26 & 2.40 & 1.53 & 1.19 & 0.47 \\
rSGM-{\em vpp} & \xmark & \cmark 
& \bf 14.35 & \bf 10.31 & \bf 9.05 & \bf 8.37 & \bf 8.18 
& \bf 10.98 & \bf 6.61 & \bf 5.51 & \bf 4.95 & \bf 2.47 
& \bf 0.62 & \bf 0.43 & \bf 0.38 & \bf 0.34 & \bf 0.20 \\
\hline\hline

PSMNet 
& \xmark & \xmark 
& 48.52 & 31.11 & 24.29 & 20.58 & 10.05 
& 47.25 & 28.03 & 20.19 & 15.88 & 4.52 
& 19.73 & 6.48 & 4.09 & 3.11 & 0.88 \\
\hline
PSMNet-{\em gd} 
& \xmark & \cmark 
& 48.24 & 30.66 & 24.03 & 20.47 & 10.50 
& 46.61 & 27.18 & 19.60 & 15.59 & 4.49 
& 18.96 & 5.91 & 3.56 & 2.85 & 0.84 \\
PSMNet-{\em vpp} & \xmark & \cmark 
& \bf 25.75 & \bf 15.75 & \bf 12.83 & \bf 11.50 & \bf 6.73 
& \bf 21.32 & \bf 10.21 & \bf 6.81 & \bf 5.29 & \bf 1.58 
& \bf 11.58 & \bf 2.43 & \bf 1.65 & \bf 1.53 & \bf 0.62 \\
\hline
PSMNet-{\em gd}-{\em ft} 
& \cmark & \cmark 
& 33.82 & 15.61 & \bf 10.74 & \bf 8.77 & \bf 4.52 
& 35.76 & 13.77 & 7.55 & \bf 5.21 & \bf 1.71 
& 12.28 & 2.43 & 0.86 & 0.61 & 0.54 \\
PSMNet-{\em vpp}-{\em ft} & \cmark & \cmark 
& \bf 23.68 & \bf 13.88 & 11.09 & 9.87 & 6.29 
& \bf 20.44 & \bf 9.90 & \bf 6.69 & 5.26 & 1.72 
& \bf 2.92 & \bf 1.43 & \bf 1.23 & \bf 1.14 & \bf 0.39 \\
\hline
PSMNet-{\em gd}-{\em tr}
& \cmark & \cmark 
& 25.17 & 12.61 & \bf 9.13 & \bf 7.59 & \bf 3.84 
& 23.67 & 9.63 & 5.75 & 4.15 & \bf 1.33 
& 4.79 & \bf 0.85 & \bf 0.54 & \bf 0.42 & 0.33 \\
PSMNet-{\em vpp}-{\em tr} & \cmark & \cmark 
& \bf 20.79 & \bf 12.07 & 9.59 & 8.44 & 4.40 
& \bf 18.03 & \bf 8.22 & \bf 5.35 & \bf 4.14 & 1.49 
& \bf 1.77 & 1.12 & 0.98 & 0.91 & \bf 0.28 \\
LidarStereoNet 
& \cmark & \cmark 
& 32.72 & 16.30 & 12.10 & 10.34 & 4.48  
& 27.32 & 11.67 & 7.58 & 5.88 & 1.80    
& 10.39 & 1.05 & 0.47 & 0.30 & 0.45 \\ 
\hline\hline
CCVNorm 
& \cmark & \cmark 
& 30.22 & 12.49 & 7.54 & 5.58 & 2.27
& 20.88 & 7.63 & 4.47 & 3.28 & 1.14
& 17.63 & 4.69 & 2.16 & 1.28 & 0.66 \\ 
\hline\hline
RAFT-Stereo 
& \xmark & \xmark 
& 24.24 & 15.66 & 12.47 & 10.63 & 3.85 
& 20.06 & 10.27 & 7.19 & 5.52 & 1.31 
& 2.62 & 1.22 & 0.88 & 0.70 & 0.27 \\
\hline
RAFT-Stereo-{\em gd} 
& \xmark & \cmark 
& 24.14 & 11.58 & 7.85 & 6.08 & 2.87  
& 20.38 & 8.31 & 5.09 & 3.65 & 1.11 
& 5.32 & 1.68 & 1.06 & 0.76 & 0.41 \\
RAFT-Stereo-{\em vpp} & \xmark & \cmark 
& \bf 7.54 & \bf 5.09 & \bf 4.20 & \bf 3.68 & \bf 2.18 
& \bf 6.58 & \bf 4.00 & \bf 3.09 & \bf 2.55 & \bf 0.67 
& \bf 1.19 & \bf 0.75 & \bf 0.57 & \bf 0.47 & \bf 0.14 \\
\hline
RAFT-Stereo-{\em gd}-{\em ft} & 
\cmark & \cmark 
& 15.22 & 8.02 & 5.97 & 5.05 & 2.67
& 15.51 & 6.76 & 4.63 & 3.63 & 1.21 
& 2.52 & 1.28 & 1.00 & 0.77 & 0.29 \\
RAFT-Stereo-{\em vpp}-{\em ft} & \cmark & \cmark 
& \bf 6.46 & \bf 4.16 & \bf 3.42 & \bf 3.04 & \bf 1.85 
& \bf 5.93 & \bf 3.59 & \bf 2.82 & \bf 2.40 & \bf 0.76 
& \bf 0.94 & \bf 0.68 & \bf 0.58 & \bf 0.54 & \bf 0.13 \\
\hline
RAFT-Stereo-{\em gd}-{\em tr} & 
\cmark & \cmark 
& 6.39 & \bf 3.29 & \bf 2.35 & \bf 1.93 & \bf 0.91 
& 6.45 & 3.14 & 2.25 & \bf 1.82 & 0.64 
& 0.94 & \bf 0.59 & \bf 0.46 & \bf 0.38 & 0.14 \\
RAFT-Stereo-{\em vpp}-{\em tr} & \cmark & \cmark 
& \bf 5.12 & 3.71 & 3.19 & 2.90 & 1.26 
& \bf 4.62 & \bf 2.82 & \bf 2.20 & 1.85 & \bf 0.62 
& \bf 0.78 & 0.61 & 0.54 & 0.50 & \bf 0.12 \\
\hline
\end{tabular}}

\caption{\textbf{Comparison with existing methods.} Results on Midd-14, Midd-21, ETH3D. Training on SceneFlow.}
\label{tab:Results_Middlebury_ETH}
\end{table*}

\subsection{Implementation and Experimental Settings}

All virtual pattern variants depicted in Fig. \ref{fig:Duck_pattern} from (ii) to (vii) are implemented in Python and Numba with sub-pixel disparities splatted on adjacent pixels in the right view. Concerning hyper-parameters, we set $\lambda=2,\gamma=0.4375,t=1$ $r_x \times r_y = 9 \times 7$ for the occlusion detection heuristic, following our previous investigation \citep{Bartolomei_2023_ICCV}, $\sigma_s=2,\sigma_c=1,t_w=0.001$ for adaptive patch and $\phi=0.3$ for distance-based pattering.
To assess the effectiveness of VPP, we run several experiments to compare it with existing approaches that combine sparse depth points with stereo algorithms and networks. In particular, we select the Guided Stereo matching framework \citep{poggi2019guided} and LidarStereoNet \citep{LIDARSTEREONET} as competitors being source code available. 

Since both are implemented over the PSMNet \citep{chang2018psmnet} architecture, we apply VPP to the same model for a direct and fair comparison. However, the original PSMNet weights used in \citep{poggi2019guided} yield poor generalization results. Hence, we retrain it following the original protocol \citep{chang2018psmnet}, i.e., for 10 epochs on SceneFlow  (specifically, using FlyingThings \citep{mayer2016large}) with a constant learning rate equal to $1e^{-3}$. For fairness, we retrain LidarStereoNet using the same protocol. Unless otherwise specified, we train all the networks according to this protocol.
Moreover, since Guided Stereo \citep{poggi2019guided} represents our closest competitor -- i.e., it does not require any architectural change to the stereo model -- we implement it over a more modern network, RAFT-Stereo \citep{lipson2021raft}, and compare it against our VPP proposal using the same network. 
To conclude, we extend this comparison with an implementation of the Semi-Global-Matching method \citep{hirschmuller2007stereo}, i.e., rSGM \citep{spangenberg2014large}, as a representative of traditional stereo algorithms. We run it by setting the maximum disparity to 192 disparity $P1=11$, adaptive $P2$ \citep{banz2012evaluation} (with $P2_\text{min}=17$, $P2_\alpha=0.5$, $P2_\gamma=35$), applying sub-pixel refinement, left-right check and a speckle filter to remove outliers, then filling holes with background interpolation \citep{Menze2015CVPR}.

\subsection{Evaluation Datasets \& Protocol}

We conduct experiments on seven indoor/outdoor datasets, including two featuring passive/active images. We refer to our codebase for the images selected from each dataset.

\textbf{Middlebury}. The Middlebury dataset \citep{scharstein2014high} is a high-resolution stereo dataset featuring indoor scenes, captured under controlled lighting conditions with accurate ground-truth obtained using structured light. We evaluate our approach on three different splits: the \textit{Additional} 13 scenes in Middlebury 2014 (Midd-A), the 15 scenes from Middlebury 2014 training set (Midd-14), 
and the 24 scenes from Middlebury 2021 (Midd-21). Results are evaluated at full resolution, with PSMNet running at half resolution. 

\textbf{KITTI 2015 \citep{Menze2015CVPR}.} This real-world stereo dataset depicts autonomous driving scenarios, captured at a resolution of approximately 1280$\times$384 pixels with sparse ground-truth depth maps collected using a LiDAR sensor. It provides 200 stereo pairs annotated with ground-truth, including independently moving objects such as cars. 
Among the 200 samples, we select 142 stereo pairs for which raw LiDAR measurements are provided \citep{LIDARSTEREONET}. This allows for evaluating the effectiveness of VPP and competitors with sparse points obtained by a real {noisy} sensor, rather than simulated.

\begin{figure}[t]
    \centering
    \includegraphics[width=0.9\linewidth]{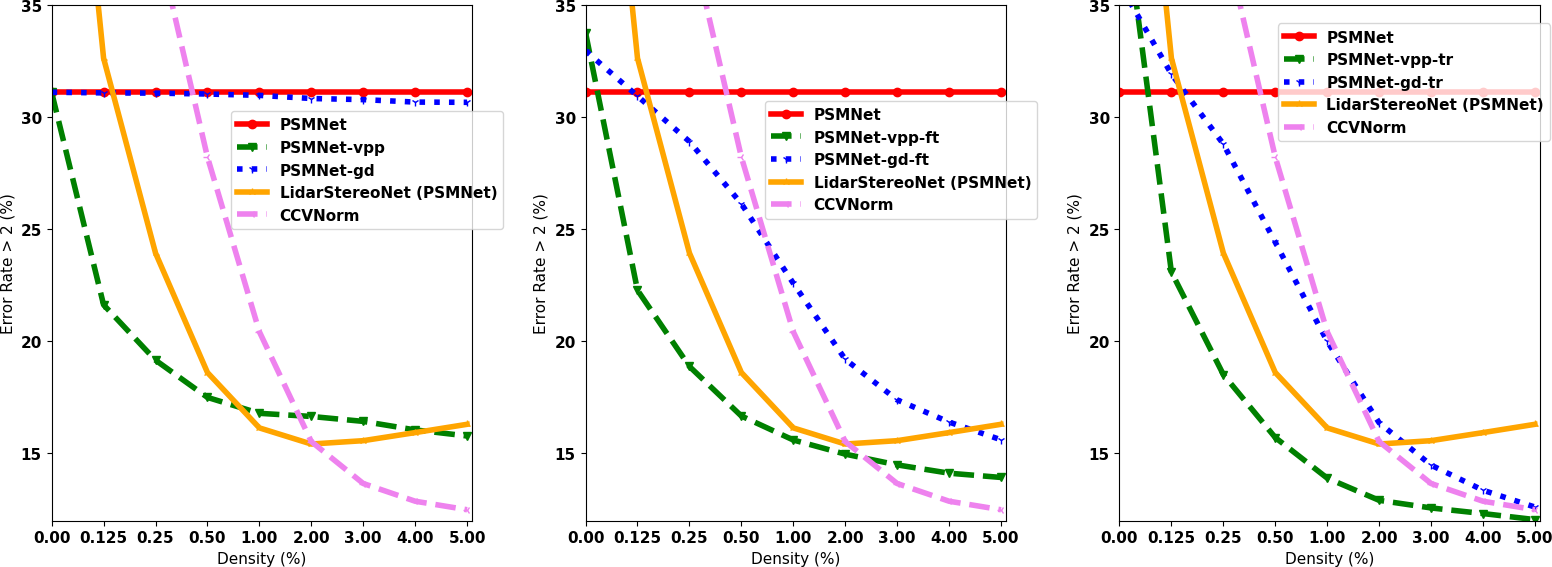}
    \includegraphics[width=0.9\linewidth]{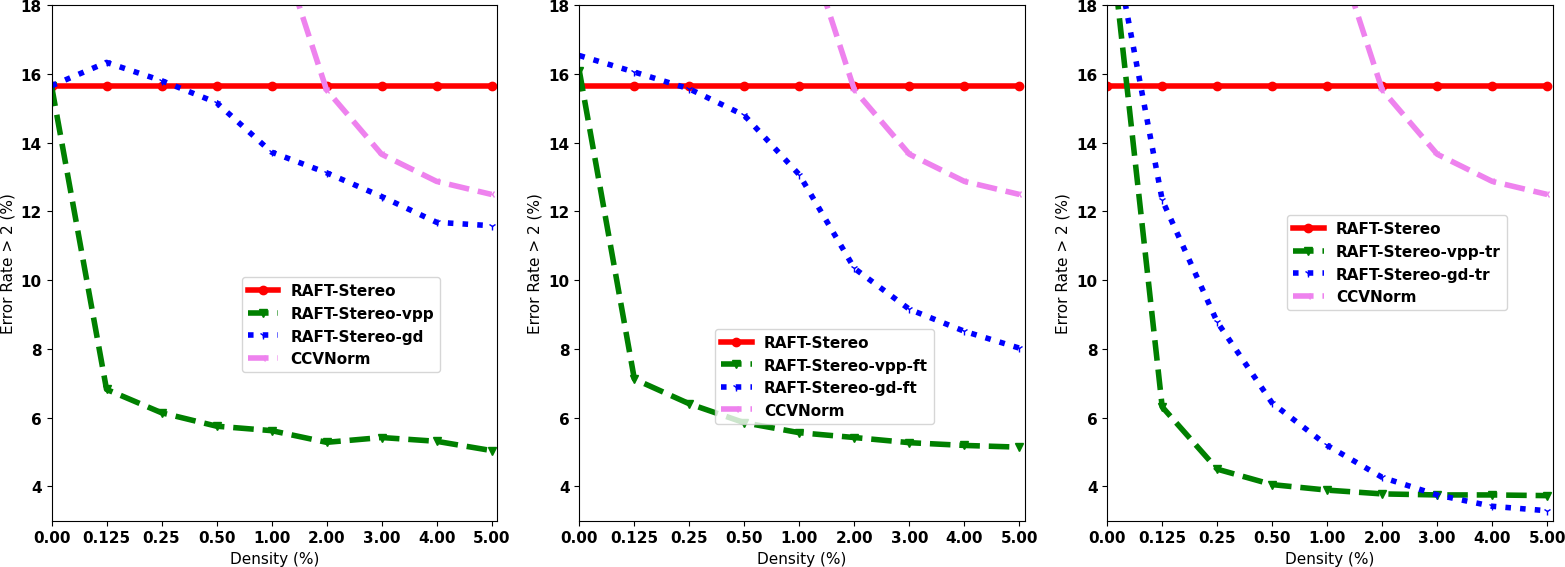}
    \caption{\textbf{Depth sparsity vs accuracy.} Results on Midd-14 \citep{scharstein2014high} by VPP and competitors with different amounts of depth points. Networks trained on synthetic data \citep{mayer2016large}.}
    \label{fig:raft_density_curve}
\end{figure}

\textbf{ETH3D \citep{schoeps2017cvpr}.} This dataset collects indoor and outdoor scenes, with a total of 27 grayscale low-resolution stereo pairs and corresponding ground-truth disparity maps.

\textbf{DSEC \citep{Gehrig21ral}.} An outdoor stereo dataset focused on autonomous driving with RGB and event cameras, captured using a wide baseline (50 cm) at $1440\times1080$ resolution. Ground-truth disparity, consisting of 26384 maps organized into 41 sequences, is obtained using a LiDAR-Inertial-Odometry (LIO) algorithm that accumulates the 16-line LiDAR scans. We split them into three different test sets with increasing challenging illumination conditions -- \ie, \textit{Day}, \textit{Afternoon} and \textit{Night} splits. We also developed a pipeline to extract de-skewed raw {noisy} LiDAR scans aligned with the reference image and to subsample our splits with a final number of 200 stereo frames for each split. 

\begin{table}[t]
\centering
\scalebox{0.667}{
\begin{tabular}{|l|cc|rrrr|r|}

\multicolumn{3}{c}{} & \multicolumn{5}{c}{Midd-14} \\ 
\hline
\multirow{2}{*}{Model} & \multicolumn{2}{c|}{Depth Points} & \multicolumn{4}{c|}{Error Rate (\%)} & avg. \\
 & Train & Test & $>1$ & $>2$ & $>3$ & $>4$ & (px) \\
 \hline\hline
rSGM 
& \xmark & \xmark &
44.82 & 32.00 & 27.21 & 24.48 & 13.19 \\
rSGM-{\em gd} & \xmark & \cmark &
32.47 & 18.80 & 14.64 & 12.77 & 9.44 \\

rSGM-{\em vpp} & \xmark & \cmark & 
14.35 & 10.31 & 9.05 & 8.37 & 8.18\\
rSGM-{\em gd}-{\em vpp} & \xmark & \cmark & 
\bf 13.96 & \bf 9.98 & \bf 8.74 & \bf 8.06 & \bf 7.91 \\
\hline\hline
PSMNet 
& \xmark & \xmark & 48.52 & 31.11 & 24.29 & 20.58 & 10.05 \\
PSMNet-{\em gd} & \xmark & \cmark & 48.24 & 30.66 & 24.03 & 20.47 & 10.50 \\

PSMNet-{\em vpp} & \xmark & \cmark & 
\bf 25.75 & \bf 15.75 & \bf 12.83 & \bf 11.50 & \bf 6.73  \\
PSMNet-{\em gd}-{\em vpp} & \xmark & \cmark & 
26.92 & 16.14 & 13.21 & 11.87 & 6.91  \\

\hline\hline
{LidarStereoNet} 
& \cmark & \cmark & 32.72 & 16.30 & 12.10 & 10.34 & \bf 4.48 \\ 
LidarStereoNet{\em -vpp} & \cmark & \cmark & \bf 30.00 & \bf 15.63 & \bf 11.91 & \bf 10.32 & 4.62 \\ 
\hline\hline
{CCVNorm} 
& \cmark & \cmark & 30.22 & \bf 12.49 & \bf 7.54 & \bf 5.58 & \bf 2.27 \\
CCVNorm{\em -vpp} & \cmark & \cmark & \bf 29.29 & 12.66 & 7.94 & 5.98 & 2.34 \\
\hline\hline

RAFT-Stereo 
& \xmark & \xmark & 24.24 & 15.65 & 12.48 & 10.62 & 3.87 \\
RAFT-Stereo-{\em gd} & \xmark & \cmark & 24.14 & 11.58 & 7.85 & 6.08 & 2.87 \\
RAFT-Stereo-{\em vpp} & \xmark & \cmark & 
\bf 7.54 & \bf 5.09 & \bf 4.20 & \bf 3.68 & \bf 2.18 \\
RAFT-Stereo-{\em gd}-{\em vpp} & \xmark & \cmark & 
12.08 & 6.12 & 4.52 & 3.71 & 2.20 \\

\hline
\end{tabular}}
\caption{\textbf{Combining VPP with existing methods.} Results on Midd-14 \citep{scharstein2014high} with networks trained on synthetic data \citep{mayer2016large}.
}
\label{tab:Results_all_together}
\end{table}

\begin{table}[t]
\centering
\scalebox{0.667}{
\begin{tabular}{|l|cc|rrrr|r|}

\multicolumn{3}{c}{} & \multicolumn{5}{c}{Midd-21} \\ 
\hline
\multirow{2}{*}{Model} & \multicolumn{2}{c|}{Depth Points} & \multicolumn{4}{c|}{Error Rate (\%)} & avg. \\
 & Train & Test & $>1$ & $>2$ & $>3$ & $>4$ & (px) \\

\hline\hline
PSMNet 
& \xmark & \xmark & 44.75 & 24.98 & 17.31 & 13.37 & 3.42 \\
\hline
PSMNet-{\em gd} & \xmark & \cmark & 44.40 & 24.58 & 16.98 & 13.09 & 3.40 \\
PSMNet-{\em vpp} & \xmark & \cmark & 
\bf 21.19 & \bf 10.56 & \bf 7.13 & \bf 5.52 & \bf 1.55\\
\hline
PSMNet-{\em gd}-{\em ft} & \cmark & \cmark & 40.47 & 16.06 & 8.56 & 5.83 & 1.78 \\
PSMNet-{\em vpp}-{\em ft} & \cmark & \cmark & 
\bf 20.65 & \bf 10.08 & \bf 6.78 & \bf 5.29 & \bf 1.61  \\
\hline
PSMNet-{\em gd}-{\em tr} & \cmark & \cmark & 23.15 & 9.44 & 5.60 & \bf 4.04 & \bf 1.29 \\
PSMNet-{\em vpp}-{\em tr} & \cmark & \cmark & 
\bf 18.08 & \bf 8.40 & \bf 5.52 & 4.29 & 1.52 \\

LidarStereoNet 
& \cmark & \cmark & 25.08 & 10.09 & 6.47 & 4.94 & 1.86 \\ 
\hline\hline
CCVNorm 
& \cmark & \cmark & 20.54 & 7.45 & 4.31 & 3.12 & 1.06 \\
\hline\hline
RAFT-Stereo 
& \xmark & \xmark & 
19.22 & 9.38 & 6.28 & 4.68 & 1.26 \\
\hline
RAFT-Stereo-{\em gd} & \xmark & \cmark & 
18.82 & 6.95 & 4.06 & 2.88 & 1.05 \\
RAFT-Stereo-{\em vpp} & \xmark & \cmark & 
\bf 6.13 & \bf 3.52 & \bf 2.67 & \bf 2.20 & \bf 0.67  \\
\hline
RAFT-Stereo-{\em gd}-{\em ft} & \cmark & \cmark & 14.87 & 6.09 & 3.99 & 3.09 & 1.02 \\
RAFT-Stereo-{\em vpp}-{\em ft} & \cmark & \cmark & 
\bf 6.18 & \bf 3.74 & \bf 2.95 & \bf 2.51 & \bf 0.70 \\
\hline
RAFT-Stereo-{\em gd}-{\em tr} & \cmark & \cmark & 6.04 & 2.91 & \bf 2.08 & \bf 1.70 & 0.58 \\
RAFT-Stereo-{\em vpp}-{\em tr} & \cmark & \cmark & 
\bf 4.36 & \bf 2.66 & 2.09 & 1.77 & \bf 0.52 \\
\hline
\end{tabular}}
\caption{\textbf{Fine-tuned models.}
Results on Midd-21, after training PSMNet \citep{chang2018psmnet}, LidarStereoNet \citep{LIDARSTEREONET} and CCVNorm \citep{wang20193d} on synthetic data \citep{mayer2016large} and fine-tuning on Midd-14 \citep{scharstein2014high}. We deployed the official Middlebury weights for RAFT-Stereo \citep{lipson2021raft}.}
\label{tab:Results_Middlebury_2021}
\end{table}

\textbf{M3ED \citep{Chaney_2023_CVPR}.} This dataset collects 57 indoor/outdoor scenes acquired using a portable multi-sensor device mounted on three different vehicles -- \ie, a drone, a quadruped robot, and a car. In contrast to KITTI and DSEC, the 64-line LiDAR provides both denser raw, noisy LiDAR scans and LIO-accumulated ground-truth maps. At the same time, the grayscale stereo rig has a shorter baseline (12 cm) and a slightly smaller resolution ($1280\times800$), cropped to $950\times525$ after our pre-processing pipeline that rectifies stereo frames and ground-truth, projects raw LiDAR scans onto the reference image, and generates a subset of 200 frames for the three splits we define --\ie, \textit{Outdoor Day}, \textit{Outdoor Night}, and \textit{Indoor}.

\begin{table}[t]
\centering
\scalebox{0.7}{
\begin{tabular}{|l|cc|rrrr|r|}

\multicolumn{3}{c}{} & \multicolumn{5}{c}{KITTI 142} \\ 
\hline
\multirow{2}{*}{Model} & \multicolumn{2}{c|}{Depth Points} & \multicolumn{4}{c|}{Error Rate (\%)} & avg. \\
 & Train & Test & $>1$ & $>2$ & $>3$ & $>4$ & (px) \\

\hline\hline
rSGM 
& \xmark & \xmark 
& 30.27 & 12.24 & 7.30 & 5.27 & 1.35  \\
\hline
rSGM-{\em gd} & \xmark & \cmark 
& 20.87 & 7.72 & 4.90 & 3.73 & \bf 1.05 \\
rSGM-{\em vpp} & \xmark & \cmark 
& \bf 14.20 & \bf 5.59 & \bf 4.18 & \bf 3.48 & 1.06  \\

\hline\hline
PSMNet 
& \xmark & \xmark & 32.50 & 11.70 & 6.40 & 4.51 & 1.32 \\
\hline
PSMNet-{\em gd} & \xmark & \cmark & 32.59 & 11.94 & 6.71 & 4.81 & 1.35 \\
PSMNet-{\em vpp} & \xmark & \cmark & 
\bf 19.92 & \bf 6.75 & \bf 4.43 & \bf 3.50 & \bf 1.09  \\
\hline
PSMNet-{\em gd}-{\em ft} & \cmark & \cmark & 30.39 & 9.79 & 5.18 & 3.69 & 1.27 \\
PSMNet-{\em vpp}-{\em ft} & \cmark & \cmark & 
\bf 20.77 & \bf 6.55 & \bf 4.11 & \bf 3.21 & \bf 1.10   \\
\hline
PSMNet-{\em gd}-{\em tr} & \cmark & \cmark & 25.84 & 8.30 & 4.73 & 3.47 & 1.17\\
PSMNet-{\em vpp}-{\em tr} & \cmark & \cmark & 
\bf 16.44 & \bf 5.94 & \bf 4.09 & \bf 3.29 & \bf 1.07 \\

LidarStereoNet 
& \cmark & \cmark & 33.01 & 10.16 & 5.13 & 3.57 & 1.33 \\ 
\hline\hline

CCVNorm 
& \cmark & \cmark & 18.98 & 6.78 & 4.50 & 3.52 & 1.17 \\
\hline\hline

RAFT-Stereo 
& \xmark & \xmark & 24.57 & 8.74 & 5.09 & 3.66 & 1.10 \\
\hline
RAFT-Stereo-{\em gd} & \xmark & \cmark & 32.50 & 12.61 & 7.10 & 4.95 & 1.33 \\
RAFT-Stereo-{\em vpp} & \xmark & \cmark & 
\bf 14.27 & \bf 6.35 & \bf 4.55 & \bf 3.63 & \bf 0.94 \\
\hline
RAFT-Stereo-{\em gd}-{\em ft} & \cmark & \cmark & 23.90 & 8.37 & 5.01 & 3.72 & 1.13 \\
RAFT-Stereo-{\em vpp}-{\em ft} & \cmark & \cmark & 
\bf 13.03 & \bf 5.62 & \bf 4.04 & \bf 3.27 & \bf 0.91 \\
\hline
RAFT-Stereo-{\em gd}-{\em tr} & \cmark & \cmark & 15.51 & 6.30 & 4.22 & 3.27 & 0.95 \\
RAFT-Stereo-{\em vpp}-{\em tr} & \cmark & \cmark & 
\bf 11.35 & \bf 4.92 & \bf 3.65 & \bf 3.00 & \bf 0.89  \\

\hline

\end{tabular}}
\caption{\textbf{Experiments on KITTI.} Results on the 142 split \citep{LIDARSTEREONET} with raw LiDAR. Networks trained on synthetic data \citep{mayer2016large}.}
\label{tab:Results_KITTI_2015}
\end{table}

\begin{table*}
\centering
\renewcommand{\tabcolsep}{6pt}
\scalebox{0.47}{
\begin{tabular}{|l|l|cc|rrrr|r|rrrr|r|rrrr|r|rrrr|r|}

\multicolumn{4}{c}{} & \multicolumn{5}{c}{Midd-14} & \multicolumn{5}{c}{Midd-21} & \multicolumn{5}{c}{ETH3D} & \multicolumn{5}{c}{KITTI 142}\\ 
\hline
 \multirow{2}{*}{Model} & & \multicolumn{2}{c|}{Depth Points} & \multicolumn{4}{c|}{Error Rate (\%)} & avg. & \multicolumn{4}{c|}{Error Rate (\%)} & avg. & \multicolumn{4}{c|}{Error Rate (\%)} & avg. & \multicolumn{4}{c|}{Error Rate (\%)} & avg. \\
  & Model name & Train & Test & $>1$ & $>2$ & $>3$ & $>4$ & (px) & $>1$ & $>2$ & $>3$ & $>4$ & (px) & $>1$ & $>2$ & $>3$ & $>4$ & (px) & $>1$ & $>2$ & $>3$ & $>4$ & (px) \\
\hline\hline
RAFT-Stereo 
& Middlebury & \xmark & \xmark 
& 17.77 & 9.74 & 7.17 & 5.85 & 1.69 
& 19.22 & 9.38 & 6.28 & 4.70 & 1.26 
& 2.75 & 1.41 & 1.01 & 0.86 & 0.29 
& 24.95 & 7.81 & 4.39 & 3.19 & 1.05 \\ 
RAFT-Stereo{\em -vpp} & Middlebury & \xmark & \cmark 
& \bf 6.36 & \bf 3.88 & \bf 3.09 & \bf 2.65 & \bf 1.01 
& \bf 6.13 & \bf 3.52 & \bf 2.67 & \bf 2.20 & \bf 0.67 
& \bf 1.10 & \bf 0.75 & \bf 0.63 & \bf 0.56 & \bf 0.14 
& \bf 14.67 & \bf 5.42 & \bf 3.78 & \bf 2.98 & \bf 0.88 \\ 
\hline\hline
RAFT-Stereo 
& ETH3D & \xmark & \xmark 
& 24.70 & 16.23 & 13.09 & 11.36 & 4.83 
& 20.26 & 10.19 & 7.09 & 5.48 & 1.39 
& 2.61 & 1.26 & 0.94 & 0.76 & 0.27 
& 24.42 & 8.63 & 4.99 & 3.64 & 1.09 \\
RAFT-Stereo{\em -vpp} & ETH3D & \xmark & \cmark 
& \bf 7.55 & \bf 5.04 & \bf 4.13 & \bf 3.60 & \bf 2.42 
& \bf 6.44 & \bf 3.83 & \bf 2.96 & \bf 2.51 & \bf 0.67 
& \bf 1.06 & \bf 0.66 & \bf 0.51 & \bf 0.42 & \bf 0.12 
& \bf 14.04 & \bf 6.19 & \bf 4.43 & \bf 3.53 & \bf 0.93 \\
\hline\hline
GMStereo 
& Sceneflow & \xmark & \xmark 
& 49.95 & 28.71 & 20.19 & 15.66 & 3.77 
& 41.10 & 21.56 & 14.32 & 10.68 & 2.39 
& 6.24 & 2.68 & 1.82 & 1.18 & 0.42 
& 29.98 & 10.25 & 5.42 & 3.77 & 1.20 \\
GMStereo{\em -vpp}$^*$  & Sceneflow & \xmark & \cmark 
& \bf 28.17 & \bf 13.59 & \bf 9.90 & \bf 8.25 & \bf 3.36 
& \bf 17.67 & \bf 8.31 & \bf 5.71 & \bf 4.46 & \bf 1.33 
& \bf 2.77 & \bf 1.74 & \bf 1.37 & \bf 1.13 & \bf 0.30 
& \bf 19.02 & \bf 7.10 & \bf 4.60 & \bf 3.49 & \bf 1.03 \\
\hline\hline
GMStereo 
& Mixdata & \xmark & \xmark 
& 31.36 & 14.11 & 8.98 & 6.63 & \bf 1.80 
& 30.44 & 11.07 & 6.20 & 4.29 & 1.32 
& \bf 1.76 & \bf 0.50 & \bf 0.36 & \bf 0.31 & \bf 0.30 
& 16.28 & 3.82 & 2.10 & 1.40 & 0.71 \\
GMStereo{\em -vpp}$^*$ & Mixdata & \xmark & \cmark 
& \bf 20.76 & \bf 9.67 & \bf 7.05 & \bf 5.85 & 1.92 
& \bf 17.86 & \bf 7.13 & \bf 4.53 & \bf 3.40 & \bf 1.07 
& 1.83 & 1.19 & 1.01 & 0.90 & 0.33 
& \bf 13.00 & \bf 3.71 & 2.24 & 1.59 & \bf 0.68 \\
\hline\hline
CFNet 
& Sceneflow & \xmark & \xmark 
& 43.78 & 29.85 & 24.32 & 21.13 & \bf 11.90 
& 39.75 & 22.44 & 16.08 & 12.81 & 11.24 
& 7.03 & 4.06 & 3.12 & 2.65 & 1.14 
& 25.49 & 9.25 & 5.25 & 3.83 & 1.11 \\
CFNet{\em -vpp}$^*$ & Sceneflow & \xmark & \cmark 
& \bf 23.54 & \bf 16.58 & \bf 13.86 & \bf 12.29 & 12.50 
& \bf 14.70 & \bf 8.23 & \bf 6.18 & \bf 5.09 & \bf 1.36 
& \bf 2.46 & \bf 1.93 & \bf 1.74 & \bf 1.61 & \bf 0.71 
& \bf 14.22 & \bf 6.27 & \bf 4.44 & \bf 3.49 & \bf 0.95 \\
\hline\hline
CFNet 
& Middlebury & \xmark & \xmark 
& 24.46 & 13.26 & 9.46 & 7.48 & \bf 2.21 
& 30.31 & 15.03 & 10.32 & 7.85 & 1.90 
& \bf 1.32 & \bf 0.53 & \bf 0.39 & \bf 0.33 & \bf 0.22 
& 11.71 & 3.20 & \bf 1.83 & \bf 1.23 & 0.60 \\
CFNet{\em -vpp}$^*$ & Middlebury & \xmark & \cmark 
& \bf 14.41 & \bf 9.59 & \bf 7.93 & \bf 7.04 & 2.85 
& \bf 12.54 & \bf 7.12 & \bf 5.38 & \bf 4.44 & \bf 1.10 
& 1.84 & 1.23 & 1.11 & 1.06 & 0.30 
& \bf 9.24 & \bf 3.02 & 1.88 & 1.35 & \bf 0.55 \\
\hline\hline
HSMNet 
& Middlebury & \xmark & \xmark 
& 32.73 & 17.50 & 12.25 & 9.57 & \bf 2.55 
& 36.77 & 18.12 & 11.97 & 8.97 & 2.27 
& 10.26 & 3.05 & 1.78 & 1.24 & 0.58 
& 31.69 & 12.34 & 6.74 & 4.55 & \bf 1.27 \\
HSMNet{\em -vpp} & Middlebury & \xmark & \cmark 
& \bf 18.90 & \bf 9.78 & \bf 7.17 & \bf 5.94 & 2.62 
& \bf 17.84 & \bf 8.61 & \bf 5.83 & \bf 4.48 & \bf 1.25 
& \bf 6.06 & \bf 2.19 & \bf 1.43 & \bf 1.14 & \bf 0.55 
& \bf 26.56 & \bf 10.54 & \bf 6.60 & \bf 4.90 & 1.32 \\
\hline\hline
CREStereo 
& ETH3D & \xmark & \xmark 
& 16.59 & 8.97 & 6.19 & 4.77 & 1.47 
& 19.07 & 9.08 & 6.33 & 4.95 & 1.24 
& 1.33 & \bf 0.60 & \bf 0.45 & \bf 0.34 & 0.19 
& 22.94 & 7.94 & 4.80 & 3.64 & 1.09 \\
CREStereo{\em -vpp}$^*$ & ETH3D & \xmark & \cmark 
& \bf 7.20 & \bf 4.84 & \bf 3.95 & \bf 3.44 & \bf 1.38 
& \bf 6.41 & \bf 3.82 & \bf 3.00 & \bf 2.53 & \bf 0.68 
& \bf 1.21 & 0.81 & 0.66 & 0.56 & \bf 0.18 
& \bf 13.93 & \bf 6.15 & \bf 4.43 & \bf 3.54 & \bf 0.96 \\
\hline\hline

{LEAStereo} 
& Sceneflow & \xmark & \xmark 
& 59.66 & 41.39 & 32.95 & 28.05 & 14.42 
& 52.03 & 32.31 & 23.22 & 18.18 & 3.73 
& 14.38 & 6.89 & 4.58 & 3.57 & 1.95 
& 46.43 & 22.28 & 13.57 & 9.70 & 1.92 \\
LEAStereo{\em -vpp} & Sceneflow & \xmark & \cmark 
& \bf 24.47 & \bf 14.83 & \bf 11.74 & \bf 10.26 & \bf 7.96 
& \bf 20.21 & \bf 10.17 & \bf 7.02 & \bf 5.56 & \bf 1.77 
& \bf 2.03 & \bf 1.06 & \bf 0.83 & \bf 0.70 & \bf 0.24 
& \bf 17.48 & \bf 7.21 & \bf 5.03 & \bf 4.02 & \bf 1.12 \\
\hline
\hline
{LEAStereo} 
& KITTI12 & \xmark & \xmark 
& 62.59 & 47.10 & 39.79 & 35.63 & 18.74 
& 56.79 & 35.90 & 26.62 & 21.46 & 5.45 
& 21.54 & 7.97 & 4.79 & 3.79 & 2.98 
& 22.49 & 6.07 & 3.00 & 2.00 & 0.87 \\
LEAStereo{\em -vpp} & KITTI12 & \xmark & \cmark 
& \bf 28.51 & \bf 16.30 & \bf 12.98 & \bf 11.56 & \bf 8.57 
& \bf 24.46 & \bf 12.15 & \bf 8.70 & \bf 7.17 & \bf 2.21 
& \bf 8.60 & \bf 1.82 & \bf 1.18 & \bf 0.93 & \bf 0.43 
& \bf 12.12 & \bf 3.66 & \bf 2.28 & \bf 1.64 & \bf 0.68 \\
\hline\hline
HITNet 
& Sceneflow & \xmark & \xmark 
& 42.67 & 26.74 & 20.49 & 17.07 & 5.88 
& 39.21 & 22.34 & 16.02 & 12.52 & 3.53 
& 18.22 & 7.95 & 4.08 & 3.25 & 0.74 
& 29.92 & 10.76 & 5.94 & 4.22 & 1.23 \\
HITNet{\em -vpp}$^*$ & Sceneflow & \xmark & \cmark 
& \bf 24.03 & \bf 13.62 & \bf 10.20 & \bf 8.52 & \bf 3.42 
& \bf 19.35 & \bf 9.96 & \bf 7.07 & \bf 5.67 & \bf 1.59 
& \bf 11.82 & \bf 3.25 & \bf 1.32 & \bf 1.15 & \bf 0.47 
& \bf 19.56 & \bf 7.00 & \bf 4.47 & \bf 3.43 & \bf 1.02 \\
\hline\hline
CoEx 
& Sceneflow & \xmark & \xmark 
& 50.56 & 40.47 & 36.52 & 34.11 & 22.19 
& 41.36 & 27.48 & 22.60 & 20.01 & 9.26 
& 16.01 & 4.67 & 2.76 & 2.08 & 0.68 
& 35.52 & 12.40 & 6.38 & 4.41 & 1.35 \\
CoEx{\em -vpp}$^*$ & Sceneflow & \xmark & \cmark 
& \bf 32.47 & \bf 27.31 & \bf 25.49 & \bf 24.38 & \bf 17.92 
& \bf 19.30 & \bf 14.01 & \bf 12.48 & \bf 11.71 & \bf 6.75 
& \bf 12.20 & \bf 2.91 & \bf 1.50 & \bf 1.24 & \bf 0.57 
& \bf 31.17 & \bf 10.14 & \bf 5.47 & \bf 4.02 & \bf 1.30 \\
\hline\hline
ELFNet 
& Sceneflow & \xmark & \xmark
& 61.96 & 44.79 & 36.02 & 30.83 & 15.51 
& 41.16 & 25.81 & 19.78 & 16.32 & 5.80 
& 24.91 & 11.90 & 7.69 & 6.11 & 3.53 
& 32.36 & 13.71 & 8.19 & 5.93 & 1.65 \\
ELFNet{\em -vpp}$^*$ & Sceneflow & \xmark & \cmark 
& \bf 40.86 & \bf 25.14 & \bf 18.89 & \bf 15.71 & \bf 7.79 
& \bf 18.80 & \bf 11.32 & \bf 9.04 & \bf 7.90 & \bf 3.42 
& \bf 3.94 & \bf 1.54 & \bf 1.30 & \bf 1.15 & \bf 0.38 
& \bf 16.44 & \bf 6.93 & \bf 4.97 & \bf 4.11 & \bf 1.25 \\
\hline\hline
PCWNet 
& Sceneflow & \xmark & \xmark 
& 50.72 & 31.59 & 24.19 & 20.28 & 8.59 
& 43.15 & 22.96 & 16.02 & 12.45 & 3.25 
& 6.13 & 2.58 & 1.78 & 1.49 & 0.48 
& 28.83 & 9.63 & 5.19 & 3.73 & 1.18 \\
PCWNet{\em -vpp}$^*$ & Sceneflow & \xmark & \cmark 
& \bf 27.09 & \bf 13.46 & \bf 9.87 & \bf 8.37 & \bf 5.72 
& \bf 22.04 & \bf 9.50 & \bf 6.06 & \bf 4.57 & \bf 1.41 
& \bf 2.25 & \bf 1.33 & \bf 1.06 & \bf 0.93 & \bf 0.34 
& \bf 16.90 & \bf 6.29 & \bf 4.18 & \bf 3.29 & \bf 0.99 \\
\hline\hline
PCWNet 
& KITTI & \xmark & \xmark 
& 55.80 & 38.31 & 31.52 & 27.85 & 12.65 
& 45.76 & 27.13 & 20.03 & 16.35 & 4.30 
& 15.47 & 2.78 & 1.36 & 1.01 & 0.58 
& 20.70 & 5.44 & 2.65 & 1.77 & 0.83 \\
PCWNet{\em -vpp}$^*$ & KITTI & \xmark & \cmark 
& \bf 28.43 & \bf 15.97 & \bf 12.73 & \bf 11.30 & \bf 7.12 
& \bf 24.46 & \bf 12.62 & \bf 9.10 & \bf 7.60 & \bf 2.31 
& \bf 13.58 & \bf 1.50 & \bf 0.79 & \bf 0.70 & \bf 0.45 
& \bf 12.04 & \bf 3.60 & \bf 2.24 & \bf 1.66 & \bf 0.68 \\
\hline\hline
{PCVNet} 
& Sceneflow & \xmark & \xmark 
& 29.16 & 20.57 & 17.07 & 14.81 & 7.01 
& 23.59 & 13.00 & 9.63 & 7.77 & 2.20 
& 4.21 & 2.07 & 1.40 & 1.05 & 0.41 
& 24.46 & 7.99 & 4.61 & 3.35 & 1.06 \\
PCVNet{\em -vpp} & Sceneflow & \xmark & \cmark 
& \bf 10.75 & \bf 7.35 & \bf 6.16 & \bf 5.47 & \bf 2.76 
& \bf 7.64 & \bf 4.31 & \bf 3.24 & \bf 2.64 & \bf 0.72 
& \bf 1.38 & \bf 0.86 & \bf 0.68 & \bf 0.56 & \bf 0.21 
& \bf 15.86 & \bf 6.16 & \bf 4.24 & \bf 3.32 & \bf 0.94 \\
\hline\hline
DLNR 
& Middlebury & \xmark & \xmark 
& 12.54 & 6.43 & 4.54 & 3.55 & 1.11 
& 15.40 & 6.77 & 4.51 & 3.43 & 1.00 
& 13.32 & 10.95 & 9.77 & 8.76 & 3.32 
& 24.58 & 8.15 & 4.78 & 3.50 & 1.11 \\
DLNR{\em -vpp} & Middlebury & \xmark & \cmark 
& \bf 5.08 & \bf 3.14 & \bf 2.54 & \bf 2.23 & \bf 0.80 
& \bf 5.35 & \bf 2.99 & \bf 2.24 & \bf 1.84 & \bf 0.59 
& \bf 6.99 & \bf 5.99 & \bf 5.53 & \bf 5.22 & \bf 2.37 
& \bf 14.32 & \bf 5.85 & \bf 4.20 & \bf 3.37 & \bf 0.94 \\
\hline\hline
NMRF 
& Sceneflow & \xmark & \xmark
& 33.85 & 21.26 & 16.65 & 14.20 & 5.88 
& 30.68 & 16.23 & 11.28 & 8.83 & 2.53 
& 4.35 & 2.42 & 1.87 & 1.54 & 0.42 
& 25.12 & 8.98 & 5.01 & 3.71 & 1.15 \\
NMRF{\em -vpp}$^*$ & Sceneflow & \xmark & \cmark
& \bf 16.90 & \bf 11.03 & \bf 9.13 & \bf 8.19 & \bf 4.86 
& \bf 12.46 & \bf 6.65 & \bf 4.87 & \bf 3.95 & \bf 1.16 
& \bf 2.05 & \bf 1.47 & \bf 1.25 & \bf 1.13 & \bf 0.34 
& \bf 15.25 & \bf 6.41 & \bf 4.48 & \bf 3.59 & \bf 1.01 \\
\hline\hline
NMRF 
& KITTI & \xmark & \xmark
& 45.55 & 27.97 & 22.01 & 18.93 & 7.64 
& 39.02 & 20.19 & 14.27 & 11.32 & 3.29 
& 20.47 & 11.18 & \bf 6.75 & \bf 5.29 & 1.96 
& \bf 4.48 & \bf 1.05 & \bf 0.60 & \bf 0.42 & \bf 0.37 \\
NMRF{\em -vpp}$^*$ & KITTI & \xmark & \cmark
& \bf 24.86 & \bf 14.42 & \bf 11.78 & \bf 10.51 & \bf 5.32 
& \bf 19.16 & \bf 9.80 & \bf 7.30 & \bf 6.20 & \bf 1.95 
& \bf 16.65 & \bf 10.64 & 7.28 & 5.36 & \bf 1.83 
& 8.21 & 2.03 & 1.19 & 0.87 & 0.50 \\
\hline

\end{tabular}}
\caption{\textbf{VPP with off-the-shelf networks.} Results on Midd-14, Midd-21, ETH3D and KITTI. Entries marked with $^*$ use $\alpha=0.2$ for blending.
}
\label{tab:roundtable}
\end{table*}

\textbf{SIMSTEREO \citep{jospin2022active}.} A synthetic indoor/outdoor passive/active dataset with dense ground-truth; RGB and IR stereo pairs have $640\times480$ resolution, with the latter modality exposing the active pattern. The authors utilized a photo-realistic renderer to generate 515 frames with overlapping RGB and IR modalities: we utilize the 103 frames from the predefined test subset. We simulate the availability of sparse depth seeds using ground-truth sampling and add a Kinect noise model \cite{handa2014benchmark} perturbating the depth proportionally to the square of the distance from the optical centre. To select a percentage of disparity hints comparable to the number of IR dots of the original dataset, we utilize the SIMSTEREO SDK to render the pattern on a plane; we then count the blobs and, finally, we randomly sample 2\% of points from the ground-truth to get $\sim6000$ noisy disparity hints -- slightly smaller than the number of IR blobs.

\textbf{M3ED-active.} 
After manual inspection of M3ED, we found that the indoor sequences collected by the quadruped robot expose an active pattern projected in the scene, appearing once every two frames. 
Accordingly, we selected two splits:  (\textit{Passive Indoor}, counting 200 frames where the active pattern is absent; and 
\textit{Active Indoor}), made of the 200 frames with visible pattern closest in time to those in the passive split. Both splits also provide raw, noisy LiDAR scans and ground-truth maps. 
Although the images in the two splits do not coincide, the tiny timeshift between adjacent passive and active frames -- $<0.1$s -- introduces negligible changes in the geometry of the scene, allowing for a fair enough comparison between methods running on the two different splits distinctly. For fairness, we manually mask part of the ground-truth and raw, noisy LiDAR scans to match regions with projected patterns.

\textbf{Evaluation Protocol.} We compute the percentage of pixels with a disparity error higher than a certain threshold $\tau$  compared to the ground-truth. 
We report error rates by varying $\tau$ to 1, 2, 3, and 4, together with the average disparity error (\textit{avg}). We evaluate our computed disparity maps across occluded and non-occluded regions with valid ground truth disparity unless otherwise noted.

\begin{table*}[t]
\centering
\renewcommand{\tabcolsep}{8pt}
\scalebox{0.53}{
\begin{tabular}{|l|l|cc|rrrr|r|rrrr|r|rrrr|r|}

\multicolumn{4}{c}{} & \multicolumn{5}{c}{DSEC Day} & \multicolumn{5}{c}{DSEC Afternoon} & \multicolumn{5}{c}{DSEC Night} \\ 
\hline
 \multirow{2}{*}{Model} &  & \multicolumn{2}{c|}{Depth Points} & \multicolumn{4}{c|}{Error Rate (\%)} & avg. & \multicolumn{4}{c|}{Error Rate (\%)} & avg. & \multicolumn{4}{c|}{Error Rate (\%)} & avg. \\
  & Model name & Train & Test & $>1$ & $>2$ & $>3$ & $>4$ & (px) & $>1$ & $>2$ & $>3$ & $>4$ & (px) & $>1$ & $>2$ & $>3$ & $>4$ & (px) \\
 \hline\hline
rSGM 
& - & \xmark & \xmark 
& 30.70 & \bf 8.57 & 3.46 & 2.07 & 1.00 
& 38.39 & 13.55 & 6.37 & 3.96 & 1.30 
& 43.08 & 17.45 & 8.91 & 5.65 & 1.58 \\
rSGM{\em -vpp} & - & \xmark & \cmark 
& \bf 29.83 & \bf 8.57 & \bf 3.33 & \bf 1.89 & \bf 0.98 
& \bf 35.82 & \bf 12.43 & \bf 5.68 & \bf 3.44 & \bf 1.21 
& \bf 38.45 & \bf 14.47 & \bf 6.94 & \bf 4.27 & \bf 1.38 \\
\hline\hline
RAFT-Stereo
& Sceneflow & \xmark & \xmark 
& 25.68 & 6.90 & 3.25 & 2.26 & 0.93 
& 27.83 & 9.46 & 4.64 & 3.06 & 1.06 
& 37.96 & 16.21 & 9.24 & 6.34 & 1.53 \\
RAFT-Stereo{\em -vpp} & Sceneflow & \xmark & \cmark 
& \bf 22.86 & \bf 6.00 & \bf 2.99 & \bf 2.13 & \bf 0.86 
& \bf 24.28 & \bf 8.08 & \bf 4.06 & \bf 2.73 & \bf 0.96 
& \bf 30.31 & \bf 12.46 & \bf 7.00 & \bf 4.80 & \bf 1.25 \\
\hline\hline
 RAFT-Stereo
 & Middlebury & \xmark & \xmark 
& 25.51 & 6.20 & 2.79 & 2.06 & 0.93 
& 23.65 & 7.66 & 4.10 & 3.00 & 1.03 
& 30.32 & 12.42 & 7.71 & 5.97 & 1.57 \\
RAFT-Stereo{\em -vpp} & Middlebury & \xmark & \cmark 
& \bf 23.43 & \bf 5.59 & \bf 2.67 & \bf 1.99 & \bf 0.88 
& \bf 21.48 & \bf 7.00 & \bf 3.85 & \bf 2.86 & \bf 0.97 
& \bf 24.20 & \bf 9.98 & \bf 6.18 & \bf 4.75 & \bf 1.27 \\
\hline\hline
 RAFT-Stereo
 & ETH3D & \xmark & \xmark 
& 24.99 & 6.57 & 3.08 & 2.15 & 0.91 
& 27.10 & 8.77 & 4.24 & 2.80 & 1.03 
& 35.30 & 14.25 & 7.92 & 5.46 & 1.44 \\
RAFT-Stereo{\em -vpp} & ETH3D & \xmark & \cmark 
& \bf 22.31 & \bf 5.76 & \bf 2.86 & \bf 2.05 & \bf 0.85 
& \bf 23.51 & \bf 7.55 & \bf 3.77 & \bf 2.55 & \bf 0.94 
& \bf 27.79 & \bf 11.01 & \bf 6.18 & \bf 4.30 & \bf 1.18 \\
\hline\hline
PSMNet 
& Sceneflow & \xmark & \xmark 
& 40.51 & \bf 15.25 & \bf 6.70 & \bf 3.82 & 1.32 
& 44.53 & 19.39 & 9.76 & 5.97 & 1.58 
& 48.12 & 21.91 & 11.61 & 7.42 & 1.88 \\
PSMNet{\em -vpp} & Sceneflow & \xmark & \cmark 
& \bf 39.78 & 15.66 & 7.03 & 3.93 & \bf 1.30 
& \bf 42.78 & \bf 18.66 & \bf 9.39 & \bf 5.66 & \bf 1.50 
& \bf 43.78 & \bf 19.63 & \bf 10.31 & \bf 6.44 & \bf 1.66 \\
\hline\hline
PSMNet 
& Middlebury & \xmark & \xmark 
& 38.38 & \bf 13.59 & \bf 5.40 & \bf 2.84 & 1.19 
& 42.13 & 16.80 & 7.78 & 4.52 & 1.41 
& 44.68 & 18.56 & 8.98 & 5.36 & 1.61 \\
PSMNet{\em -vpp} & Middlebury & \xmark & \cmark 
& \bf 37.22 & 13.61 & 5.54 & 2.89 & \bf 1.18 
& \bf 40.22 & \bf 16.16 & \bf 7.52 & \bf 4.35 & \bf 1.36 
& \bf 41.35 & \bf 17.29 & \bf 8.52 & \bf 5.10 & \bf 1.49 \\
\hline\hline
GMStereo 
& Sceneflow & \xmark & \xmark 
& 32.91 & 10.58 & 4.79 & 3.01 & 1.13 
& 43.34 & 17.56 & 8.16 & 4.76 & 1.42 
& 50.94 & 24.86 & 13.76 & 8.87 & 1.93 \\
GMStereo{\em -vpp}$^*$  & Sceneflow & \xmark & \cmark 
& \bf 30.15 & \bf 9.51 & \bf 4.43 & \bf 2.86 & \bf 1.07 
& \bf 39.70 & \bf 15.52 & \bf 7.34 & \bf 4.43 & \bf 1.34 
& \bf 45.76 & \bf 21.12 & \bf 11.69 & \bf 7.61 & \bf 1.75 \\
\hline\hline
GMStereo 
& Mixdata & \xmark & \xmark 
& 35.00 & 10.44 & 4.62 & 3.13 & 1.18 
& 33.22 & 12.45 & 6.69 & 4.67 & 1.30 
& 37.56 & 14.26 & 7.27 & \bf 4.77 & \bf 1.40 \\
GMStereo{\em -vpp}$^*$ & Mixdata & \xmark & \cmark 
& \bf 31.59 & \bf 9.31 & \bf 4.42 & \bf 3.06 & \bf 1.13 
& \bf 32.03 & \bf 11.97 & \bf 6.50 & \bf 4.58 & \bf 1.29 
& \bf 34.92 & \bf 13.43 & \bf 7.08 & \bf 4.77 & \bf 1.40 \\
\hline\hline
CFNet 
& Sceneflow & \xmark & \xmark 
& 29.37 & 8.42 & 3.77 & 2.48 & 1.02 
& 41.49 & 16.24 & 7.80 & 4.83 & 1.44 
& 55.00 & 31.54 & 22.22 & 18.23 & 26.79 \\
CFNet{\em -vpp}$^*$ & Sceneflow & \xmark & \cmark 
& \bf 27.95 & \bf 8.27 & \bf 3.71 & \bf 2.43 & \bf 0.99 
& \bf 38.24 & \bf 14.96 & \bf 7.26 & \bf 4.51 & \bf 1.36 
& \bf 45.46 & \bf 23.41 & \bf 15.45 & \bf 12.22 & \bf 13.05 \\
\hline\hline
CFNet 
& Middlebury & \xmark & \xmark 
& 25.66 & \bf 5.85 & \bf 2.48 & \bf 1.71 & 0.90 
& 28.68 & 8.87 & \bf 3.92 & \bf 2.55 & 1.06 
& 31.41 & 10.92 & \bf 5.13 & \bf 3.24 & \bf 1.18 \\
CFNet{\em -vpp}$^*$ & Middlebury & \xmark & \cmark 
& \bf 24.53 & 5.87 & 2.56 & 1.77 & \bf 0.89 
& \bf 26.88 & \bf 8.55 & 4.00 & 2.68 & \bf 1.04 
& \bf 28.95 & \bf 10.46 & 5.25 & 3.48 & 1.19 \\
\hline\hline
HSMNet 
& Middlebury & \xmark & \xmark 
& 32.13 & 9.40 & 4.07 & 2.52 & 1.04 
& 36.08 & 12.80 & 5.93 & 3.56 & 1.20 
& 40.44 & 15.84 & 7.98 & 4.98 & 1.43 \\
HSMNet{\em -vpp} & Middlebury & \xmark & \cmark 
& \bf 30.08 & \bf 8.73 & \bf 3.83 & \bf 2.38 & \bf 1.00 
& \bf 33.85 & \bf 11.61 & \bf 5.34 & \bf 3.19 & \bf 1.13 
& \bf 36.38 & \bf 13.49 & \bf 6.71 & \bf 4.22 & \bf 1.30 \\
\hline\hline
CREStereo 
& ETH3D & \xmark & \xmark 
& 23.72 & 6.21 & 3.28 & 2.51 & 0.94 
& 24.26 & 8.10 & 4.43 & 3.20 & 1.03 
& 28.76 & 11.14 & 6.45 & 4.70 & 1.30 \\
CREStereo{\em -vpp}$^*$ & ETH3D & \xmark & \cmark 
& \bf 22.10 & \bf 5.86 & \bf 3.24 & \bf 2.49 & \bf 0.90 
& \bf 22.51 & \bf 7.64 & \bf 4.25 & \bf 3.11 & \bf 0.98 
& \bf 25.05 & \bf 10.16 & \bf 6.06 & \bf 4.46 & \bf 1.20 \\
\hline\hline
{LEAStereo} 
& Sceneflow & \xmark & \xmark 
& 46.70 & 21.98 & 12.28 & 8.22 & 1.88 
& 53.54 & 28.17 & 16.36 & 10.75 & 2.11 
& 64.89 & 41.42 & 28.96 & 22.16 & 3.98 \\
LEAStereo{\em -vpp} & Sceneflow & \xmark & \cmark 
& \bf 39.99 & \bf 16.55 & \bf 8.13 & \bf 5.05 & \bf 1.42 
& \bf 46.65 & \bf 22.53 & \bf 12.26 & \bf 7.85 & \bf 1.76 
& \bf 51.04 & \bf 27.93 & \bf 17.62 & \bf 12.81 & \bf 2.59 \\
\hline\hline
{LEAStereo} 
& KITTI12 & \xmark & \xmark 
& 40.33 & 14.56 & 6.92 & 4.30 & 1.44 
& 43.61 & 17.07 & 7.74 & 4.43 & 1.50 
& 49.29 & 23.26 & 13.16 & 8.94 & 2.25 \\
LEAStereo{\em -vpp} & KITTI12 & \xmark & \cmark 
& \bf 33.37 & \bf 10.29 & \bf 4.72 & \bf 3.01 & \bf 1.18 
& \bf 36.67 & \bf 13.00 & \bf 5.88 & \bf 3.58 & \bf 1.33 
& \bf 38.82 & \bf 15.78 & \bf 8.50 & \bf 5.80 & \bf 1.64 \\
\hline\hline
HITNet 
& Sceneflow & \xmark & \xmark 
& 36.53 & 13.24 & 6.41 & 4.16 & 1.30 
& 44.52 & 19.27 & 10.38 & 6.94 & 1.68 
& 46.78 & 21.46 & 12.11 & 8.27 & 1.91 \\
HITNet{\em -vpp}$^*$ & Sceneflow & \xmark & \cmark 
& \bf 34.83 & \bf 12.93 & \bf 6.35 & \bf 4.09 & \bf 1.25 
& \bf 41.87 & \bf 18.22 & \bf 9.85 & \bf 6.56 & \bf 1.60 
& \bf 41.33 & \bf 18.68 & \bf 10.51 & \bf 7.22 & \bf 1.73 \\
\hline\hline
CoEx 
& Sceneflow & \xmark & \xmark 
& \bf 30.59 & \bf 9.28 & \bf 4.15 & \bf 2.75 & \bf 1.06 
& 35.12 & 13.09 & 6.64 & 4.40 & 1.30 
& 39.20 & 15.78 & 8.54 & 5.87 & 1.57 \\
CoEx{\em -vpp}$^*$ & Sceneflow & \xmark & \cmark 
& 30.99 & 10.38 & 4.67 & 2.83 & \bf 1.06 
& \bf 33.62 & \bf 12.78 & \bf 6.46 & \bf 4.16 & \bf 1.24 
& \bf 35.01 & \bf 13.94 & \bf 7.43 & \bf 4.95 & \bf 1.38 \\
\hline\hline
ELFNet 
& Sceneflow & \xmark & \xmark
& 45.42 & 23.60 & 14.33 & 9.93 & 2.14 
& 48.35 & 25.96 & 16.00 & 11.11 & 2.07 
& 58.49 & 36.09 & 25.11 & 19.24 & 3.56 \\
ELFNet{\em -vpp}$^*$ & Sceneflow & \xmark & \cmark 
& \bf 37.46 & \bf 15.88 & \bf 8.35 & \bf 5.40 & \bf 1.40 
& \bf 41.72 & \bf 19.69 & \bf 11.15 & \bf 7.48 & \bf 1.63 
& \bf 45.50 & \bf 23.75 & \bf 14.67 & \bf 10.50 & \bf 2.21 \\
\hline\hline
PCWNet 
& Sceneflow & \xmark & \xmark 
& 42.95 & 16.33 & 7.96 & 4.99 & 1.45 
& 50.21 & 22.83 & 11.80 & 7.25 & 1.76 
& 55.93 & 30.04 & 18.43 & 13.08 & 2.72 \\
PCWNet{\em -vpp}$^*$ & Sceneflow & \xmark & \cmark 
& \bf 39.08 & \bf 13.78 & \bf 6.61 & \bf 4.21 & \bf 1.33 
& \bf 45.24 & \bf 19.19 & \bf 9.75 & \bf 6.15 & \bf 1.59 
& \bf 48.09 & \bf 23.43 & \bf 13.80 & \bf 9.81 & \bf 2.21 \\
\hline\hline
PCWNet 
& KITTI & \xmark & \xmark 
& 32.73 & 10.01 & 4.46 & 2.83 & 1.16 
& 38.26 & 13.33 & 5.67 & 3.40 & 1.34 
& 42.00 & 16.00 & 7.43 & 4.46 & 1.49 \\
PCWNet{\em -vpp}$^*$ & KITTI & \xmark & \cmark 
& \bf 29.24 & \bf 8.70 & \bf 4.01 & \bf 2.65 & \bf 1.10 
& \bf 32.89 & \bf 10.97 & \bf 5.06 & \bf 3.28 & \bf 1.24 
& \bf 34.07 & \bf 12.52 & \bf 6.55 & \bf 4.48 & \bf 1.43 \\
\hline\hline
{PCVNet} 
& Sceneflow & \xmark & \xmark 
& 26.42 & 7.07 & 3.20 & 2.19 & 0.93 
& 27.65 & 9.07 & 4.34 & 2.84 & 1.03 
& 35.93 & 15.20 & 8.72 & 6.11 & 1.69 \\
PCVNet{\em -vpp} & Sceneflow & \xmark & \cmark 
& \bf 23.98 & \bf 6.07 & \bf 2.82 & \bf 1.97 & \bf 0.87 
& \bf 24.36 & \bf 7.72 & \bf 3.77 & \bf 2.54 & \bf 0.95 
& \bf 28.51 & \bf 11.23 & \bf 6.24 & \bf 4.35 & \bf 1.23 \\
\hline\hline
DLNR 
& Middlebury & \xmark & \xmark 
& 27.93 & 10.43 & \bf 7.13 & \bf 6.09 & 1.55 
& 44.38 & 31.10 & 26.65 & 24.11 & 4.44 
& 61.11 & 50.30 & 45.85 & 43.00 & 9.01 \\
DLNR{\em -vpp} & Middlebury & \xmark & \cmark 
& \bf 27.09 & \bf 10.29 & 7.22 & 6.18 & \bf 1.53 
& \bf 42.75 & \bf 30.73 & \bf 26.51 & \bf 24.01 & \bf 4.30 
& \bf 58.51 & \bf 48.66 & \bf 44.53 & \bf 41.86 & \bf 8.72 \\
\hline\hline
NMRF 
& Sceneflow & \xmark & \xmark
& 28.96 & 8.51 & 3.85 & \bf 2.59 & 1.05 
& 37.21 & 14.11 & 6.79 & 4.34 & 1.35 
& 40.88 & 16.81 & 8.76 & 5.78 & 1.57 \\
NMRF{\em -vpp}$^*$ & Sceneflow & \xmark & \cmark
& \bf 27.78 & \bf 8.37 & \bf 3.84 & 2.60 & \bf 1.03 
& \bf 35.39 & \bf 13.40 & \bf 6.56 & \bf 4.25 & \bf 1.32 
& \bf 36.57 & \bf 15.03 & \bf 8.01 & \bf 5.41 & \bf 1.48 \\
\hline\hline
NMRF 
& KITTI & \xmark & \xmark
& 29.27 & 6.70 & \bf 2.77 & \bf 1.92 & 1.03 
& 30.18 & 9.18 & \bf 4.24 & \bf 2.80 & \bf 1.17 
& 32.41 & 11.44 & \bf 5.87 &  \bf4.02 & \bf 1.34 \\
NMRF{\em -vpp}$^*$ & KITTI & \xmark & \cmark
& \bf 27.45 & \bf 6.48 & 2.83 & 2.01 & \bf 1.01 
& \bf 29.59 & \bf 9.17 & 4.40 & 2.98 & 1.19 
& \bf 29.97 & \bf 10.82 & 5.90 & 4.23 & \bf 1.34 \\
\hline

\hline
\end{tabular}}
\caption{\textbf{VPP with more off-the-shelf networks.} Results on DSEC \citep{Gehrig21ral} using 16-line LiDAR. Entries marked with $^*$ use $\alpha=0.2$ for blending.}
\label{tab:Results_DSEC}\vspace{-0.3cm}
\end{table*}

\subsection{Ablation Study}

We start by studying the impact of VPP under different settings in Tab. \ref{tab:ablation_adaptive}. These experiments are carried out on Midd-A at full resolution, involving PSMNet \citep{chang2018psmnet}, RAFT-Stereo \citep{lipson2021raft}, and rSGM \citep{spangenberg2014large}. The first row (A) reports the error rates achieved by the original networks/algorithm without applying any virtual pattern projection. In (B), we enable VPP by setting the hyper-parameters according to our previous ablation studies \citep{Bartolomei_2023_ICCV} -- \ie, a $3\times3$ patch-based random pattern {referred to as} (vi) with alpha-blending $\alpha=0.4$ and ``FGD" occlusion setting -- appreciating the notable decrease in error rates. {The patch-based random pattern (vi) remains the best trade-off between speed and pattern distinctiveness since the probability of overlapping random patterns is low, given the realistic assumption of sparse depth hints.} Starting from these settings, we evaluated different patches with a grid search, studying the impact given by using (C) distance-based patches or (D) adaptive patches, observing that only the latter yields a further improvement, optimal with a $7 \times 7$ patch. Finally, we combine the two in (E), finding that the best choice is (D) alone. Therefore, as hyper-parameters for the remaining experiments, we will assume those found in \cite{Bartolomei_2023_ICCV} using $7 \times 7$ adaptive patches (D).

\subsection{Comparison with Existing Approaches}

We now assess the performance of VPP and its two main competitors. 
{If not differently specified, we randomly sample 5\% depth points from GT, as done by \cite{poggi2019guided}, for datasets with dense ground-truth annotations and without sparse depth hints -- \ie, Middlebury, ETH3D.}

\textbf{Guided/VPP variants.} Given the flexibility of VPP and Guided Stereo, we evaluate their application to stereo networks under three different settings, respectively:

\begin{itemize}
    \item Without retraining the stereo network (\textbf{-gd/-vpp})
    \item After a brief fine-tuning of the pre-trained model, enhanced by Guided Stereo or VPP frameworks (\textbf{-gd-ft}/\textbf{-vpp-ft})
    \item By training from scratch a new model, enhanced by Guided Stereo or VPP (\textbf{-gd-tr}/\textbf{-vpp-tr})
\end{itemize}
For \textit{-ft} variants, two epochs of finetuning are carried out on FlyingThings \citep{mayer2016large}, with learning rate 1e-4 and 1e-5 for PSMNet and RAFT-Stereo, respectively.
For \textit{-tr} variants, PSMNet and RAFT-Stereo are trained for 10 and 20 epochs, with learning rates $1e^{-3}$ and $1e^{-4}$, respectively. In any case, PSMNet and RAFT-Stereo process $384\times512$ and $360\times720$ crops, respectively, with batch size 2 on a single 3090 GPU. 

The three variants allow for evaluating Guided Stereo and VPP when deployed with the least effort -- i.e., by simply taking a pre-trained stereo network off the shelf -- or with deeper intervention by the developer, either through a short fine-tuning on the pre-existing model or, with major efforts, by retraining it from scratch.

\begin{figure}[t]
    \centering
    \includegraphics[width=0.34\textwidth]{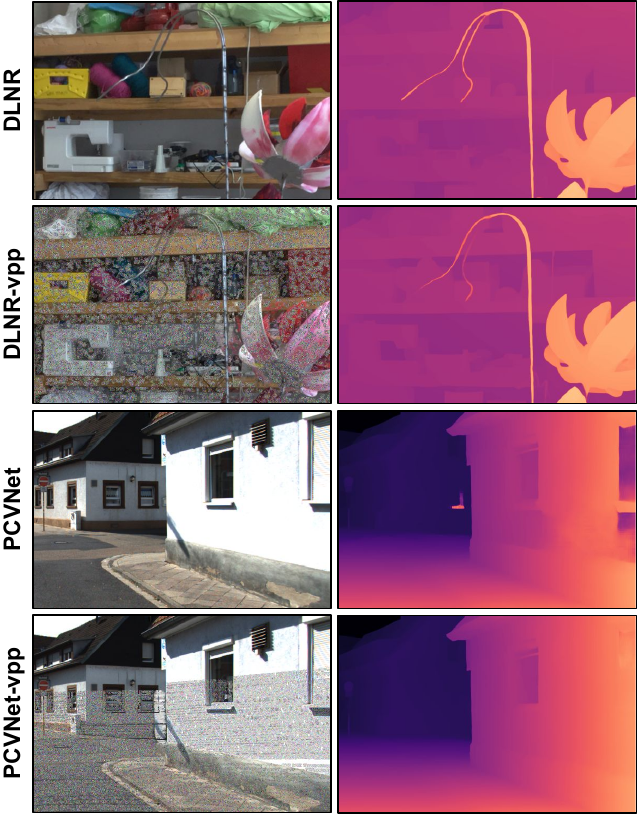}\\ 
    \caption{\textbf{Stereo networks DLNR and PCVNet with and without VPP.} In both cases, VPP preserves most fine details, and there is a significant improvement in the presence of uniform regions (e.g., as evident with the exterior white walls).}
    \label{fig:thin}
\end{figure}

\textbf{Comparison on Middlebury/ETH3D.}
Tab. \ref{tab:Results_Middlebury_ETH} reports the outcome on Midd-14, Midd-21 and ETH3D datasets using the rSGM algorithm and models trained on synthetic data only. 
From it, we can highlight one of the most evident advantages of our proposal, i.e., its remarkable capability to boost cross-domain generalization on all datasets without any retraining (-vpp). In contrast, Guided Stereo yields only marginal improvements under this setting (-gd), and LidarStereoNet needs training from scratch to process depth points since they are concatenated to the RGB images. In light of this evidence, the VPP paradigm is undoubtedly the best choice to boost the accuracy of \textit{off-the-shelf} networks.

Fine-tuning the networks shortly to get acquainted better with the sparse depth cues is beneficial for Guided Stereo (-gd-ft) and VPP (-vpp-ft) despite the latter frequently -- always with RAFTStereo -- outperforming the former even without any fine-tuning (-vpp). When training from scratch the networks, Guided Stereo (-gd-tr) is, sometimes, marginally more effective than  VPP (-vpp-tr). This fact again emphasizes how VPP is much more training independent, a remarkable factor of merit of our proposal. Indeed, acting at the image level turns out more robust than concatenating depth to RGB (LidarStereoNet) or acting on cost volumes (Guided Stereo) and, not of little importance, allows for a seamless integration even with closed source code.

\textbf{Depth Sparsity vs Accuracy.} Fig. \ref{fig:raft_density_curve} plots the error rate ($>2$) on Midd-14 varying the density of sparse depth points from 0\% to 5\%. From it, we observe that VPP generally yields almost optimal performance nearly with 1\% or less density and, with few exceptions -- the -tr configurations with higher density -- achieves much lower error rates.

\textbf{VPP with other Guided Methods.} Since we can seamlessly use VPP with other image-guided methods, Tab. \ref{tab:Results_all_together} reports the results of such joint deployment on the Midd-14. Except with rSGM, benefiting the joint deployment with Guided Stereo, VPP alone is always more effective.
Deploying VPP with LidarStereoNet also yields slight improvements.

\begin{figure}[t]
    \centering
    \includegraphics[width=0.7\linewidth]{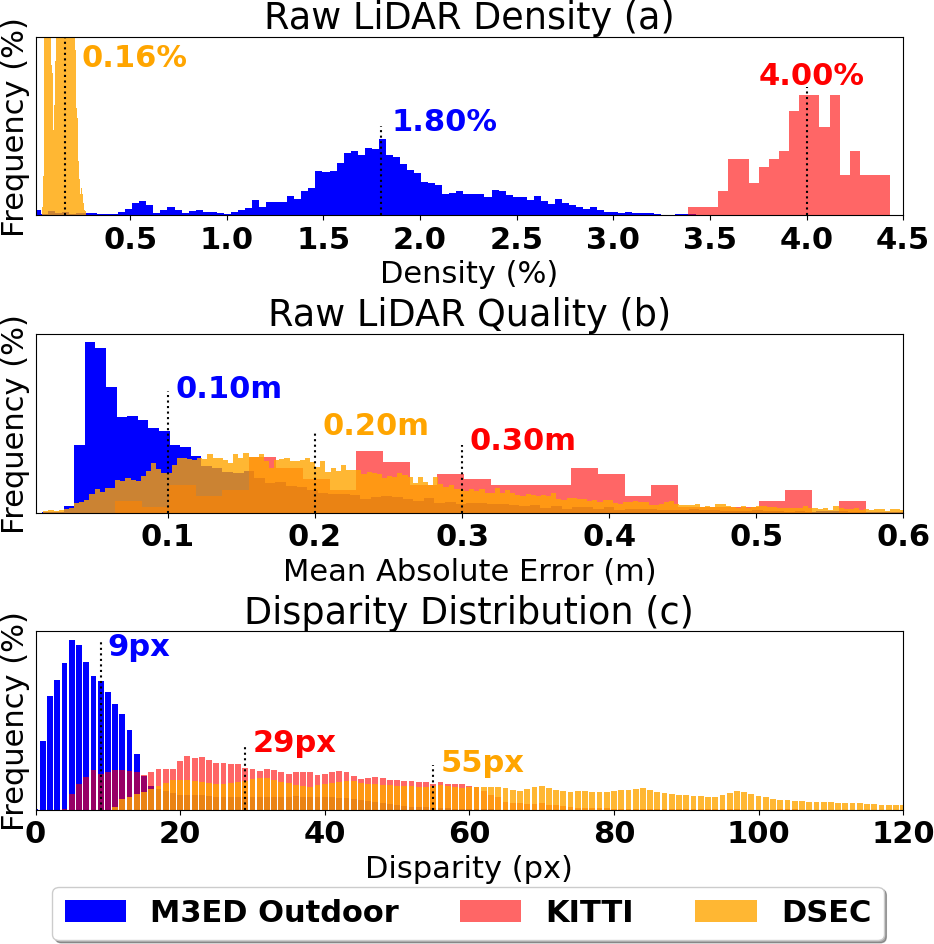}
    \caption{\textbf{Analysis of outdoor datasets.} We compare the density of raw depth hints (a), the noisiness w.r.t ground-truth (b), and disparity distribution (c) of three different datasets: KITTI \citep{Menze2015CVPR}, DSEC \citep{Gehrig21ral}, and M3ED \citep{Chaney_2023_CVPR} outdoor sequences.}
    \label{fig:lidar_quality}
\end{figure}

\begin{figure*}[t]
    \centering
    \renewcommand{\tabcolsep}{1pt}
    \begin{tabular}{ccccc}

        RGB \& VPP & LEAStereo & LEAStereo{\em -vpp} & PCWNet & PCWNet{\em -vpp} \\ 
        
        \includegraphics[trim=0cm 0cm 0cm 0cm, clip, width=0.185\linewidth]{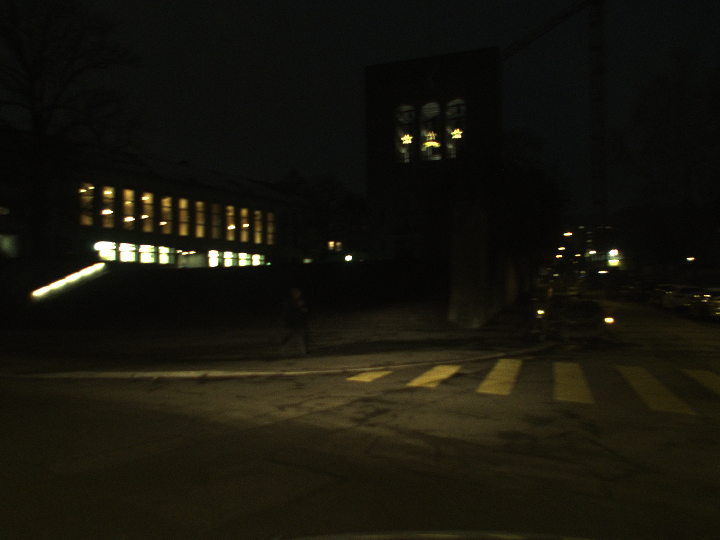} & 
        \includegraphics[trim=0cm 0cm 0cm 0cm, clip, width=0.185\linewidth]{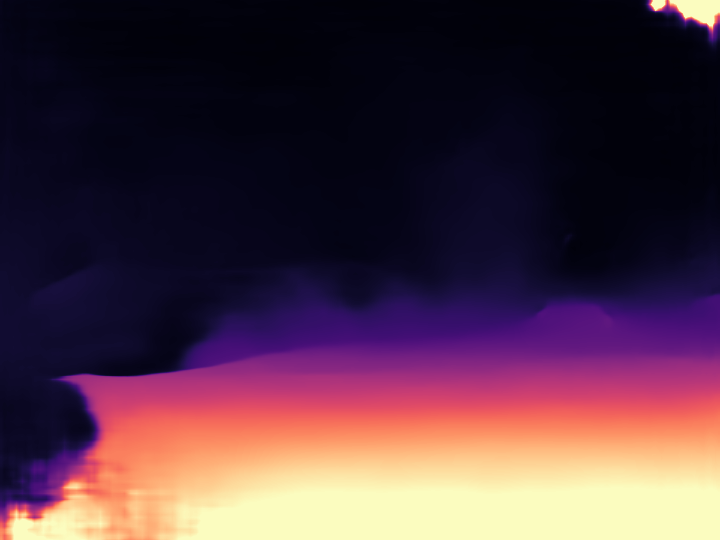} & 
        \includegraphics[trim=0cm 0cm 0cm 0cm, clip, width=0.185\linewidth]{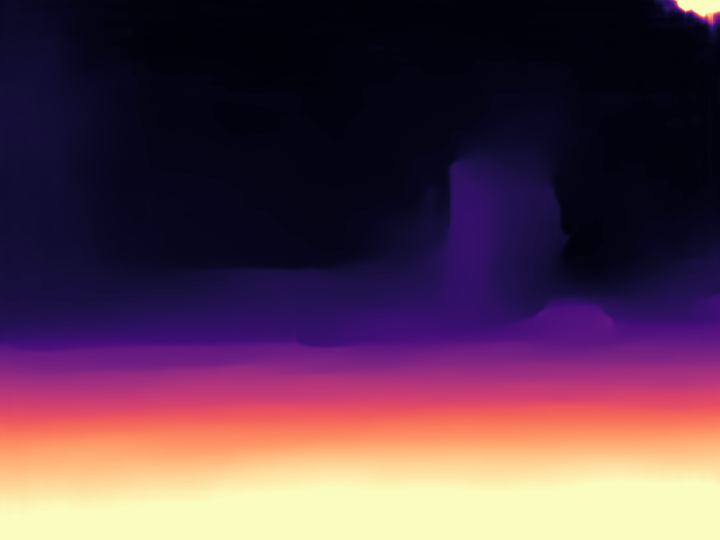} & 
        \includegraphics[trim=0cm 0cm 0cm 0cm, clip, width=0.185\linewidth]{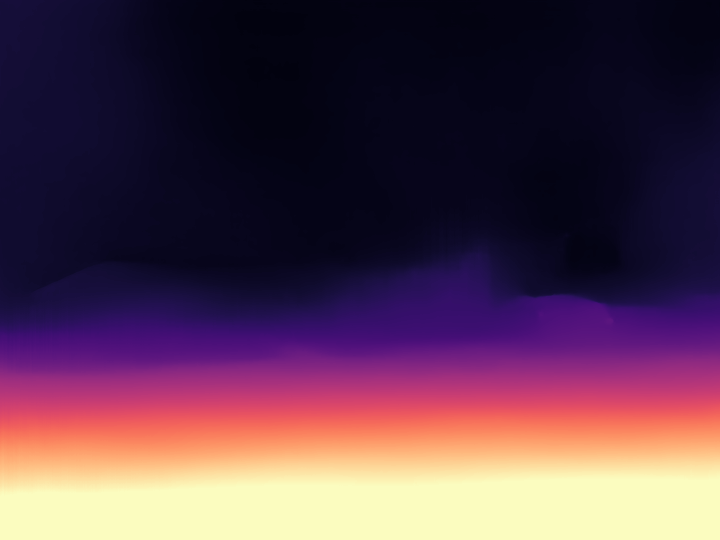} & 
        \includegraphics[trim=0cm 0cm 0cm 0cm, clip, width=0.185\linewidth]{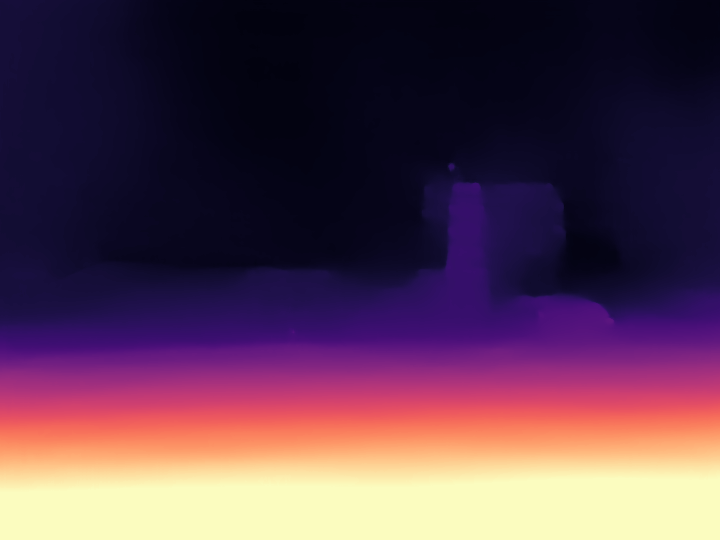} \\
        
        \begin{overpic}[trim=0cm 0cm 0cm 0cm, clip, width=0.185\linewidth]{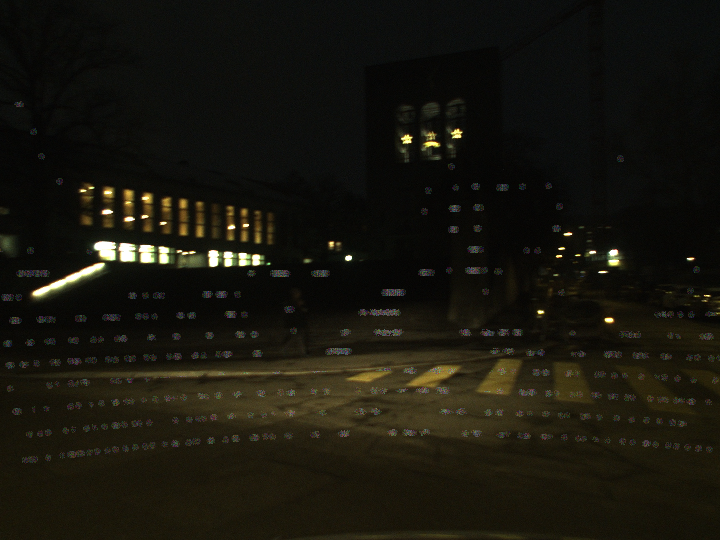}
        \end{overpic} &
        \begin{overpic}[trim=0cm 0cm 0cm 0cm, clip, width=0.185\linewidth]{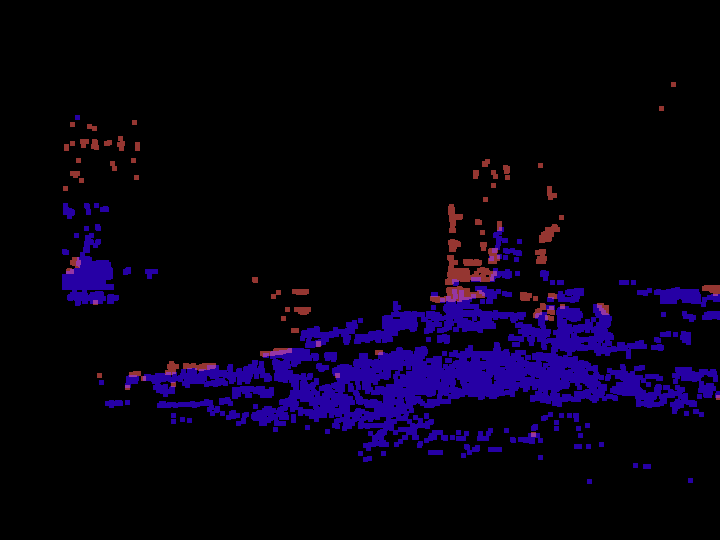}
          \put(5,2){\footnotesize\textcolor{white}{\textbf{Error Rate $>$ 3: 11.73\%}}}
        \end{overpic} &
        \begin{overpic}[trim=0cm 0cm 0cm 0cm, clip, width=0.185\linewidth]{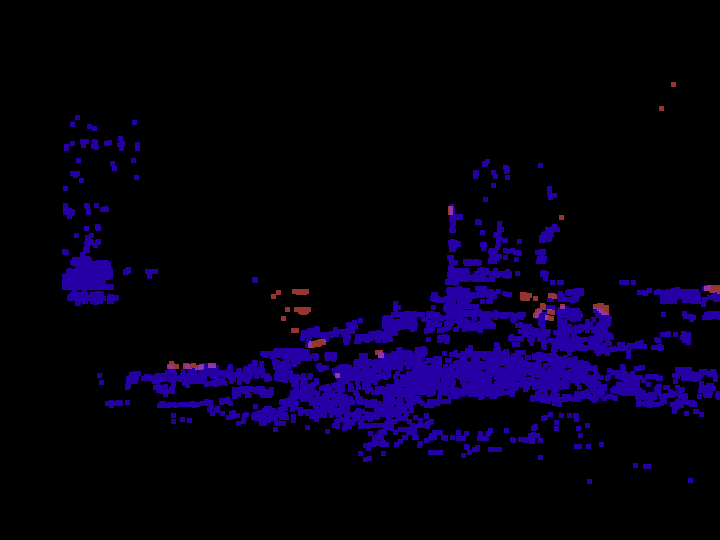}
          \put(5,2){\footnotesize\textcolor{white}{\textbf{Error Rate $>$ 3: 2.55\%}}}
        \end{overpic} & 
        \begin{overpic}[trim=0cm 0cm 0cm 0cm, clip, width=0.185\linewidth]{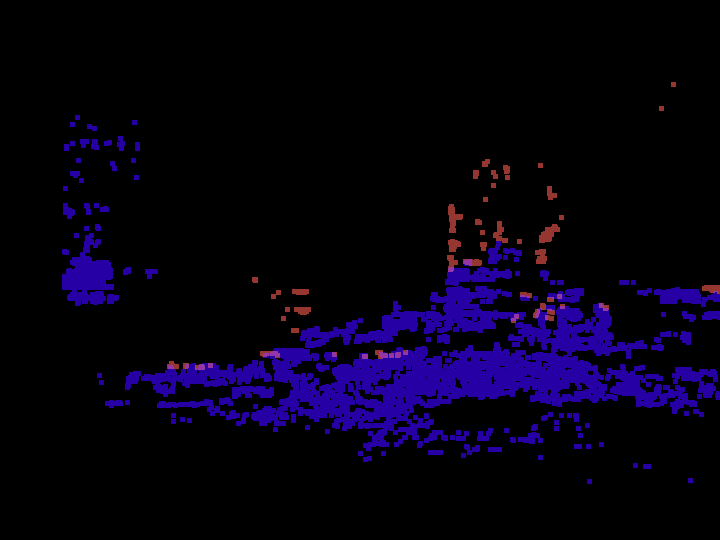} 
          \put(5,2){\footnotesize\textcolor{white}{\textbf{Error Rate $>$ 3: 5.03\%}}}
        \end{overpic} &
        \begin{overpic}[trim=0cm 0cm 0cm 0cm, clip, width=0.185\linewidth]{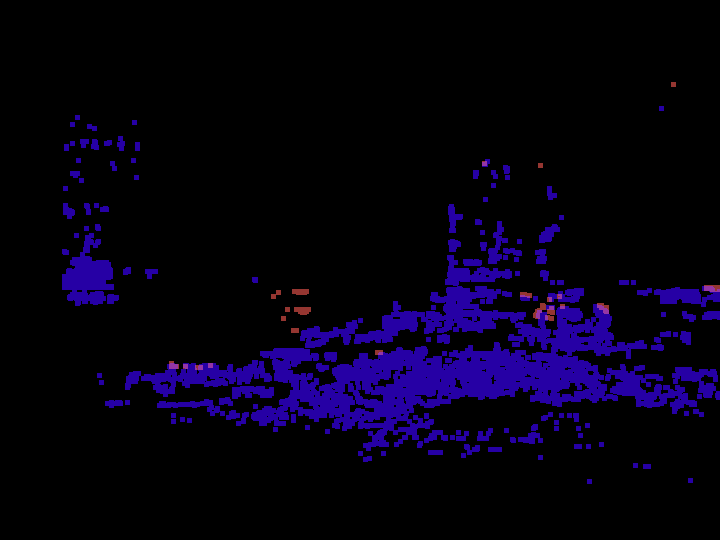} 
         \put(5,2){\footnotesize\textcolor{white}{\textbf{Error Rate $>$ 3: 1.46\%}}}
        \end{overpic}\vspace{0.1cm}
         
    \end{tabular}
    \caption{\textbf{Qualitative result on DSEC \citep{Gehrig21ral} night split.} Passive stereo networks (\eg, LEAStereo \citep{cheng2020hierarchical} and PCWNet \citep{shen2022pcw}) often fail when dealing with low ambient light, while VPP allows perceiving even challenging objects, such as the far background walls.}
    \label{fig:qualitative_dsec_night}
\end{figure*}

\textbf{Results after Fine-Tuning on Real Data.} Similarly to Tab. \ref{tab:Results_Middlebury_ETH}, Tab. \ref{tab:Results_Middlebury_2021} reports results after fine-tuning PSMNet and LidarStereoNet on Midd-14 for $\sim4000$ steps \citep{teed2020raft}. For RAFT-Stereo, we use the official Middlebury weights provided by the authors. We can observe a similar trend, with most VPP variants consistently outperforming those based on Guided Stereo and LidarStereoNet. This behaviour underlines how VPP improves cross-domain generalization as well as domain specialization.

 \textbf{Comparison on KITTI.} To prove the robustness of VPP with noisy depth data inferred with a real sensor outdoor and at long-range, where a physical pattern would be useless, Tab. \ref{tab:Results_KITTI_2015} reports experimental results on the KITTI 2015 dataset using the raw output of a Velodyne LiDAR. 
 The results confirm the previous trends: VPP constantly outperforms Guided Stereo, with any network and configuration, and LidarStereoNet.

\begin{table*}[t]
\centering
\renewcommand{\tabcolsep}{8pt}
\scalebox{0.53}{
\begin{tabular}{|l|l|cc|rrrr|r|rrrr|r|rrrr|r|}

\multicolumn{4}{c}{} & \multicolumn{5}{c}{M3ED Outdoor Day} & \multicolumn{5}{c}{M3ED Outdoor Night} & \multicolumn{5}{c}{M3ED Indoor} \\ 
\hline
 \multirow{2}{*}{Model} &  & \multicolumn{2}{c|}{Depth Points} & \multicolumn{4}{c|}{Error Rate (\%)} & avg. & \multicolumn{4}{c|}{Error Rate (\%)} & avg. & \multicolumn{4}{c|}{Error Rate (\%)} & avg. \\
 & Model name & Train & Test & $>1$ & $>2$ & $>3$ & $>4$ & (px) & $>1$ & $>2$ & $>3$ & $>4$ & (px) & $>1$ & $>2$ & $>3$ & $>4$ & (px) \\
 \hline\hline
rSGM 
& - & \xmark & \xmark 
& 44.51 & 13.50 & 6.63 & 3.75 & 1.23 
& 79.04 & 45.38 & 21.22 & 11.27 & 2.36 
& 62.51 & 40.57 & 30.40 & 24.69 & 6.46 \\
rSGM{\em -vpp} & - & \xmark & \cmark 
& \bf 5.88 & \bf 1.74 & \bf 0.99 & \bf 0.66 & \bf 0.33 
& \bf 6.39 & \bf 2.76 & \bf 1.51 & \bf 0.98 & \bf 0.35 
& \bf 24.55 & \bf 13.79 & \bf 9.30 & \bf 7.03 & \bf 1.58 \\
\hline\hline 
 RAFT-Stereo
 & Sceneflow & \xmark & \xmark 
& 43.04 & 8.87 & 4.94 & 2.69 & 1.14 
& 89.44 & 39.75 & 14.42 & 11.29 & 13.36 
& 52.74 & 29.96 & 21.70 & 16.92 & 3.06 \\
RAFT-Stereo{\em -vpp} & Sceneflow & \xmark & \cmark 
& \bf 6.97 & \bf 1.95 & \bf 1.15 & \bf 0.78 & \bf 0.36 
& \bf 5.90 & \bf 1.76 & \bf 0.75 & \bf 0.41 & \bf 0.28 
& \bf 26.27 & \bf 15.24 & \bf 10.78 & \bf 8.38 & \bf 1.99 \\
\hline\hline
 RAFT-Stereo
 & Middlebury & \xmark & \xmark 
& 41.08 & 9.12 & 5.28 & 3.18 & 1.65 
& 88.90 & 35.94 & 10.70 & 7.24 & 4.02 
& 50.87 & 28.04 & 19.99 & 15.53 & 3.11 \\
RAFT-Stereo{\em -vpp} & Middlebury & \xmark & \cmark 
& \bf 6.38 & \bf 1.89 & \bf 1.21 & \bf 0.87 & \bf 0.36 
& \bf 5.16 & \bf 1.51 & \bf 0.73 & \bf 0.43 & \bf 0.27 
& \bf 24.98 & \bf 14.17 & \bf 9.76 & \bf 7.45 & \bf 1.78 \\
\hline\hline
 RAFT-Stereo
 & ETH3D & \xmark & \xmark 
& 43.81 & 8.90 & 4.97 & 2.72 & 1.14 
& 87.66 & 35.72 & 10.57 & 7.52 & 6.04 
& 51.95 & 29.19 & 21.01 & 16.39 & 3.00 \\
RAFT-Stereo{\em -vpp} & ETH3D & \xmark & \cmark 
& \bf 6.94 & \bf 1.97 & \bf 1.15 & \bf 0.78 & \bf 0.36 
& \bf 5.75 & \bf 1.69 & \bf 0.71 & \bf 0.39 & \bf 0.27 
& \bf 26.12 & \bf 15.20 & \bf 10.74 & \bf 8.35 & \bf 2.37 \\
\hline\hline
PSMNet 
& Sceneflow & \xmark & \xmark 
& 59.54 & 22.68 & 9.34 & 4.53 & 1.80 
& 91.78 & 61.08 & 24.83 & 9.25 & 3.64 
& 59.77 & 36.86 & 26.86 & 21.55 & 4.24 \\
PSMNet{\em -vpp} & Sceneflow & \xmark & \cmark 
& \bf 25.34 & \bf 8.41 & \bf 2.12 & \bf 1.06 & \bf 0.82 
& \bf 27.16 & \bf 10.13 & \bf 2.15 & \bf 1.01 & \bf 0.90 
& \bf 27.63 & \bf 16.01 & \bf 11.24 & \bf 8.78 & \bf 2.21 \\
\hline\hline
PSMNet 
& Middlebury & \xmark & \xmark 
& 54.69 & 19.26 & 7.94 & 3.83 & 1.74 
& 83.05 & 42.72 & 14.82 & 6.01 & 4.03 
& 56.44 & 34.06 & 24.95 & 19.88 & 3.87 \\
PSMNet{\em -vpp} & Middlebury & \xmark & \cmark 
& \bf 23.97 & \bf 8.89 & \bf 2.03 & \bf 1.04 & \bf 0.86 
& \bf 25.50 & \bf 10.22 & \bf 1.87 & \bf 0.97 & \bf 1.06 
& \bf 27.80 & \bf 16.17 & \bf 11.33 & \bf 8.83 & \bf 2.22 \\
\hline\hline
GMStereo 
& Sceneflow & \xmark & \xmark 
& 42.37 & 10.57 & 5.01 & 2.61 & 1.12 
& 65.72 & 29.31 & 9.67 & 4.77 & 1.76 
& 55.80 & 33.24 & 24.60 & 19.73 & 3.56 \\
GMStereo{\em -vpp}$^*$  & Sceneflow & \xmark & \cmark 
& \bf 8.70 & \bf 2.42 & \bf 1.26 & \bf 0.77 & \bf 0.47 
& \bf 10.04 & \bf 2.81 & \bf 1.25 & \bf 0.65 & \bf 0.43 
& \bf 29.67 & \bf 17.26 & \bf 11.95 & \bf 9.01 & \bf 1.80 \\
\hline\hline
GMStereo 
& Mixdata & \xmark & \xmark 
& 33.55 & 7.99 & 4.50 & 2.33 & 1.01 
& 81.00 & 21.99 & 4.87 & 2.13 & 1.67 
& 48.34 & 27.69 & 19.92 & 15.30 & 2.78 \\
GMStereo{\em -vpp}$^*$ & Mixdata & \xmark & \cmark 
& \bf 8.72 & \bf 2.61 & \bf 1.39 & \bf 0.85 & \bf 0.44 
& \bf 10.04 & \bf 3.16 & \bf 1.40 & \bf 0.69 & \bf 0.41 
& \bf 26.67 & \bf 15.10 & \bf 10.26 & \bf 7.71 & \bf 1.62 \\
\hline\hline
CFNet 
& Sceneflow & \xmark & \xmark 
& 46.04 & 10.84 & 6.21 & 3.91 & 2.26 
& 91.02 & 63.93 & 47.12 & 44.85 & 76.28 
& 58.71 & 36.75 & 28.06 & 23.41 & 11.56 \\
CFNet{\em -vpp}$^*$ & Sceneflow & \xmark & \cmark 
& \bf 7.85 & \bf 2.21 & \bf 1.35 & \bf 0.97 & \bf 0.46 
& \bf 12.09 & \bf 7.77 & \bf 6.87 & \bf 6.50 & \bf 5.03 
& \bf 26.91 & \bf 15.62 & \bf 10.81 & \bf 8.33 & \bf 2.90 \\
\hline\hline
CFNet 
& Middlebury & \xmark & \xmark 
& 40.76 & 8.14 & 4.48 & 2.57 & 1.32 
& 82.64 & 27.44 & 6.37 & 3.53 & 2.01 
& 49.97 & 28.23 & 20.32 & 15.82 & 3.20 \\
CFNet{\em -vpp}$^*$ & Middlebury & \xmark & \cmark 
& \bf 6.03 & \bf 1.56 & \bf 0.98 & \bf 0.72 & \bf 0.43 
& \bf 5.12 & \bf 1.74 & \bf 1.12 & \bf 0.86 & \bf 0.35 
& \bf 23.93 & \bf 13.45 & \bf 9.09 & \bf 6.87 & \bf 1.49 \\
\hline\hline
HSMNet 
& Middlebury & \xmark & \xmark 
& 39.38 & 11.28 & 5.26 & 2.64 & 1.11 
& 74.93 & 32.11 & 9.67 & 3.06 & 1.73 
& 57.39 & 34.30 & 23.72 & 18.22 & 3.41 \\
HSMNet{\em -vpp} & Middlebury & \xmark & \cmark 
& \bf 12.19 & \bf 2.62 & \bf 1.17 & \bf 0.72 & \bf 0.56 
& \bf 9.92 & \bf 2.09 & \bf 0.78 & \bf 0.44 & \bf 0.49 
& \bf 33.06 & \bf 16.79 & \bf 10.68 & \bf 7.79 & \bf 1.93 \\
\hline\hline
CREStereo 
& ETH3D & \xmark & \xmark 
& 40.29 & 8.99 & 5.27 & 2.92 & 1.13 
& 90.53 & 31.79 & 6.31 & 2.97 & 1.91 
& 51.60 & 29.23 & 20.96 & 16.25 & 2.93 \\
CREStereo{\em -vpp}$^*$ & ETH3D & \xmark & \cmark 
& \bf 10.34 & \bf 3.28 & \bf 1.87 & \bf 1.18 & \bf 0.48 
& \bf 10.38 & \bf 3.19 & \bf 1.23 & \bf 0.62 & \bf 0.35 
& \bf 28.69 & \bf 16.39 & \bf 11.22 & \bf 8.46 & \bf 1.72 \\
\hline\hline
{LEAStereo} 
& Sceneflow & \xmark & \xmark 
& 53.18 & 20.40 & 10.68 & 6.04 & 1.71 
& 88.43 & 62.78 & 38.88 & 28.29 & 11.87 
& 62.06 & 38.04 & 28.06 & 22.41 & 5.14 \\
LEAStereo{\em -vpp} & Sceneflow & \xmark & \cmark 
& \bf 6.05 & \bf 1.78 & \bf 1.09 & \bf 0.77 & \bf 0.38 
& \bf 5.53 & \bf 2.44 & \bf 1.42 & \bf 0.90 & \bf 0.45 
& \bf 25.16 & \bf 13.81 & \bf 9.15 & \bf 6.86 & \bf 1.47 \\
\hline\hline
{LEAStereo} 
& KITTI12 & \xmark & \xmark 
& 40.57 & 12.84 & 7.34 & 4.56 & 2.13 
& 75.45 & 49.43 & 32.11 & 25.00 & 15.88 
& 56.51 & 34.05 & 24.28 & 18.87 & 4.04 \\
LEAStereo{\em -vpp} & KITTI12 & \xmark & \cmark 
& \bf 6.77 & \bf 1.27 & \bf 0.76 & \bf 0.55 & \bf 0.44 
& \bf 7.61 & \bf 2.31 & \bf 1.60 & \bf 1.26 & \bf 0.60 
& \bf 25.20 & \bf 13.55 & \bf 8.94 & \bf 6.56 & \bf 1.61 \\
\hline\hline
HITNet 
& Sceneflow & \xmark & \xmark 
& 57.81 & 18.77 & 9.39 & 4.76 & 1.47 
& 89.86 & 56.53 & 22.08 & 10.06 & 3.27 
& 56.88 & 34.75 & 26.02 & 21.14 & 4.50 \\
HITNet{\em -vpp}$^*$ & Sceneflow & \xmark & \cmark 
& \bf 21.92 & \bf 7.83 & \bf 3.34 & \bf 1.46 & \bf 0.74 
& \bf 21.39 & \bf 7.64 & \bf 2.76 & \bf 1.10 & \bf 0.67 
& \bf 29.73 & \bf 17.11 & \bf 12.04 & \bf 9.30 & \bf 2.14 \\
\hline\hline
CoEx 
& Sceneflow & \xmark & \xmark 
& 50.50 & 17.44 & 7.27 & 3.58 & 1.37 
& 87.22 & 53.00 & 22.21 & 8.86 & 4.29 
& 56.64 & 33.90 & 24.81 & 19.76 & 4.88 \\
CoEx{\em -vpp}$^*$ & Sceneflow & \xmark & \cmark 
& \bf 22.39 & \bf 6.46 & \bf 2.31 & \bf 1.26 & \bf 0.76 
& \bf 20.93 & \bf 5.95 & \bf 1.95 & \bf 0.81 & \bf 0.74 
& \bf 31.08 & \bf 17.14 & \bf 11.85 & \bf 9.16 & \bf 2.30 \\
\hline\hline
ELFNet 
& Sceneflow & \xmark & \xmark
& 60.09 & 23.93 & 11.59 & 5.83 & 1.93 
& 90.65 & 69.69 & 43.46 & 28.40 & 15.69 
& 57.11 & 36.44 & 28.74 & 24.12 & 6.96 \\
ELFNet{\em -vpp}$^*$ & Sceneflow & \xmark & \cmark 
& \bf 16.25 & \bf 3.79 & \bf 1.85 & \bf 1.33 & \bf 0.65 
& \bf 14.75 & \bf 3.93 & \bf 2.10 & \bf 1.68 & \bf 1.66 
& \bf 27.72 & \bf 16.00 & \bf 11.66 & \bf 9.41 & \bf 2.87 \\
\hline\hline
PCWNet 
& Sceneflow & \xmark & \xmark 
& 49.14 & 11.03 & 5.63 & 3.28 & 1.31 
& 88.83 & 48.30 & 19.37 & 10.17 & 2.43 
& 55.94 & 33.07 & 23.96 & 18.82 & 3.70 \\
PCWNet{\em -vpp}$^*$ & Sceneflow & \xmark & \cmark 
& \bf 10.48 & \bf 2.47 & \bf 1.43 & \bf 1.02 & \bf 0.56 
& \bf 8.99 & \bf 1.90 & \bf 0.82 & \bf 0.49 & \bf 0.38 
& \bf 27.86 & \bf 14.93 & \bf 9.90 & \bf 7.35 & \bf 1.60 \\
\hline\hline
PCWNet 
& KITTI & \xmark & \xmark 
& 33.36 & 8.27 & 5.37 & 3.39 & 1.76 
& 63.67 & 20.20 & 7.20 & 4.83 & 2.18 
& 52.86 & 29.49 & 20.45 & 15.61 & 3.23 \\
PCWNet{\em -vpp}$^*$ & KITTI & \xmark & \cmark 
& \bf 14.03 & \bf 2.51 & \bf 1.65 & \bf 1.32 & \bf 0.79 
& \bf 14.80 & \bf 1.80 & \bf 0.81 & \bf 0.56 & \bf 0.51 
& \bf 24.69 & \bf 12.87 & \bf 8.53 & \bf 6.29 & \bf 1.54 \\ 
\hline\hline
{PCVNet} 
& Sceneflow & \xmark & \xmark 
& 42.84 & 10.38 & 5.74 & 3.32 & 1.25 
& 86.96 & 45.19 & 24.00 & 19.15 & 13.25 
& 55.67 & 34.29 & 25.84 & 20.91 & 6.19 \\
PCVNet{\em -vpp} & Sceneflow & \xmark & \cmark 
& \bf 8.55 & \bf 2.33 & \bf 1.31 & \bf 0.88 & \bf 0.44 
& \bf 8.57 & \bf 2.53 & \bf 1.53 & \bf 1.17 & \bf 0.73 
& \bf 28.20 & \bf 16.48 & \bf 11.55 & \bf 8.90 & \bf 2.76 \\
\hline\hline
DLNR 
& Middlebury & \xmark & \xmark 
& 54.04 & 26.19 & 21.82 & 18.72 & 6.63 
& 90.96 & 76.47 & 68.15 & 65.46 & 38.39 
& 53.64 & 32.57 & 24.26 & 19.43 & 3.99 \\
DLNR{\em -vpp} & Middlebury & \xmark & \cmark 
& \bf 23.03 & \bf 17.81 & \bf 15.82 & \bf 14.35 & \bf 4.75 
& \bf 48.85 & \bf 44.29 & \bf 41.70 & \bf 39.75 & \bf 19.73 
& \bf 28.06 & \bf 17.12 & \bf 12.34 & \bf 9.80 & \bf 2.26 \\
\hline\hline
NMRF 
& Sceneflow & \xmark & \xmark
& 42.58 & 9.23 & 4.96 & 2.69 & 1.15 
& 84.26 & 37.34 & 11.22 & 6.16 & 2.12 
& 53.01 & 30.77 & 22.48 & 17.92 & 3.28 \\ 
NMRF{\em -vpp}$^*$ & Sceneflow & \xmark & \cmark
& \bf 10.17 & \bf 2.70 & \bf 1.53 & \bf 1.01 & \bf 0.50 
& \bf 10.71 & \bf 2.69 & \bf 1.18 & \bf 0.72 & \bf 0.49 
& \bf 28.98 & \bf 16.67 & \bf 11.61 & \bf 8.90 & \bf 1.85 \\
\hline\hline
NMRF 
& KITTI & \xmark & \xmark
& 40.19 & 20.30 & 14.90 & 11.63 & 5.36 
& 73.95 & 34.72 & 22.21 & 17.83 & 7.22 
& 48.20 & 27.07 & 19.69 & 15.36 & 3.12 \\
NMRF{\em -vpp}$^*$ & KITTI & \xmark & \cmark
& \bf 22.89 & \bf 13.95 & \bf 10.10 & \bf 7.94 & \bf 3.37 
& \bf 23.56 & \bf 14.02 & \bf 10.10 & \bf 7.10 & \bf 2.38 
& \bf 26.18 & \bf 14.43 & \bf 9.83 & \bf 7.39 & \bf 1.57 \\
\hline
\end{tabular}}
\caption{\textbf{VPP with more off-the-shelf networks.} Results on M3ED \citep{Chaney_2023_CVPR} using 64-line LiDAR. Entries marked with $^*$ use $\alpha=0.2$ for blending.}
\label{tab:Results_M3ED}
\end{table*}

\begin{figure*}[t]
    \centering
    \renewcommand{\tabcolsep}{1pt}
    \begin{tabular}{ccccc}

        Grayscale \& VPP & ELFNet & ELFNet{\em -vpp} & HITNet & HITNet{\em -vpp} \\ 
        
        \includegraphics[trim=10cm 0cm 0cm 2cm, clip, width=0.185\linewidth]{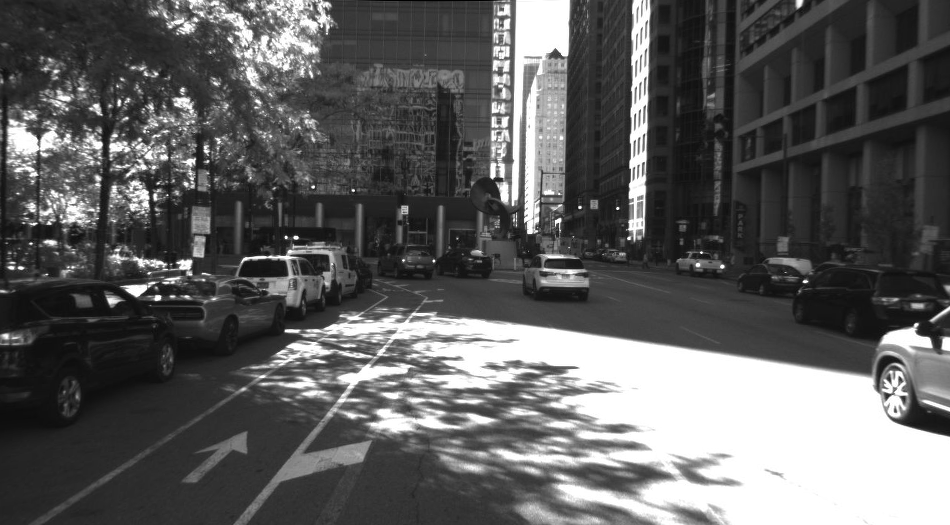} & 
        \includegraphics[trim=10cm 0cm 0cm 2cm, clip, width=0.185\linewidth]{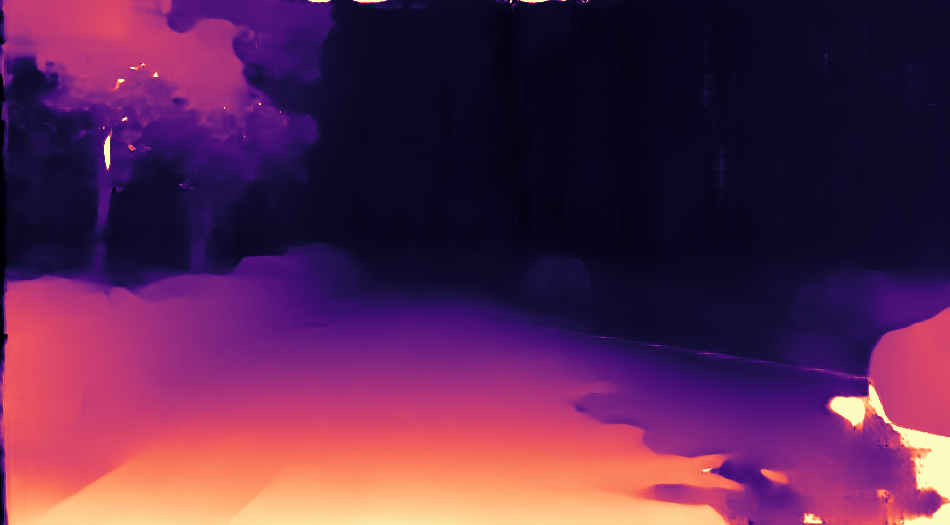} & 
        \includegraphics[trim=10cm 0cm 0cm 2cm, clip, width=0.185\linewidth]{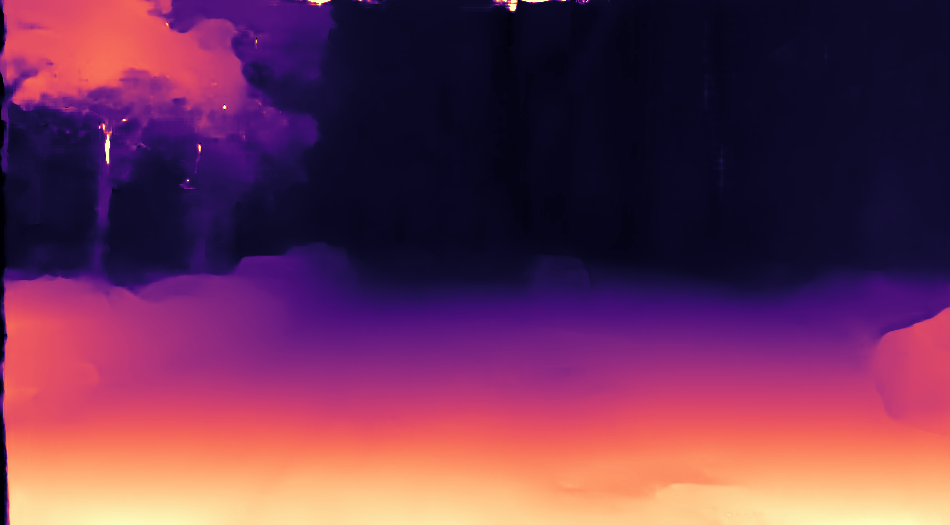} & 
        \includegraphics[trim=10cm 0cm 0cm 2cm, clip, width=0.185\linewidth]{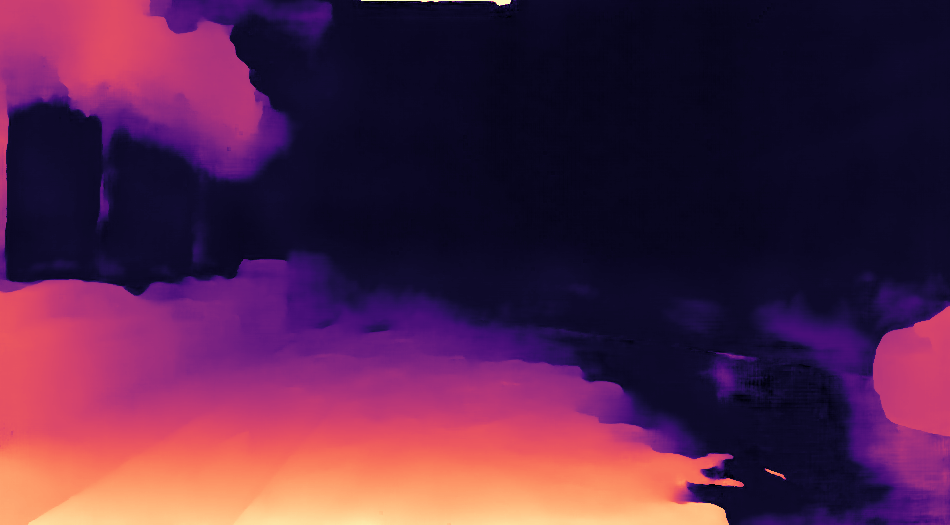} & 
        \includegraphics[trim=10cm 0cm 0cm 2cm, clip, width=0.185\linewidth]{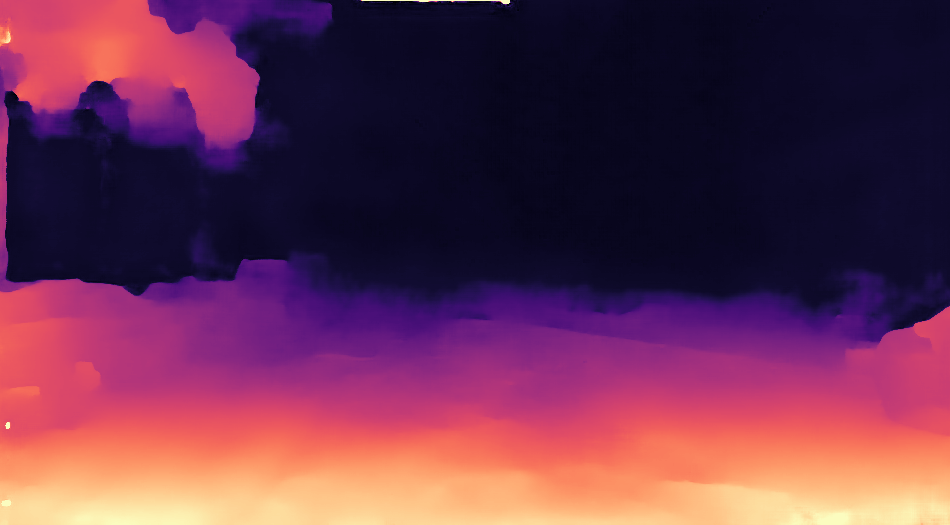} \\
        
        \begin{overpic}[trim=10cm 0cm 0cm 2cm, clip, width=0.185\linewidth]{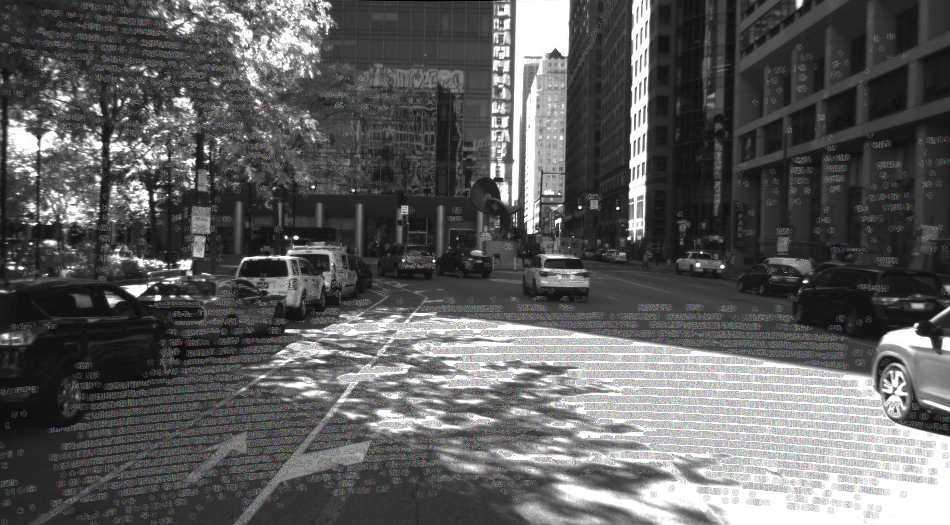}
        \end{overpic} &
        \begin{overpic}[trim=10cm 0cm 0cm 2cm, clip, width=0.185\linewidth]{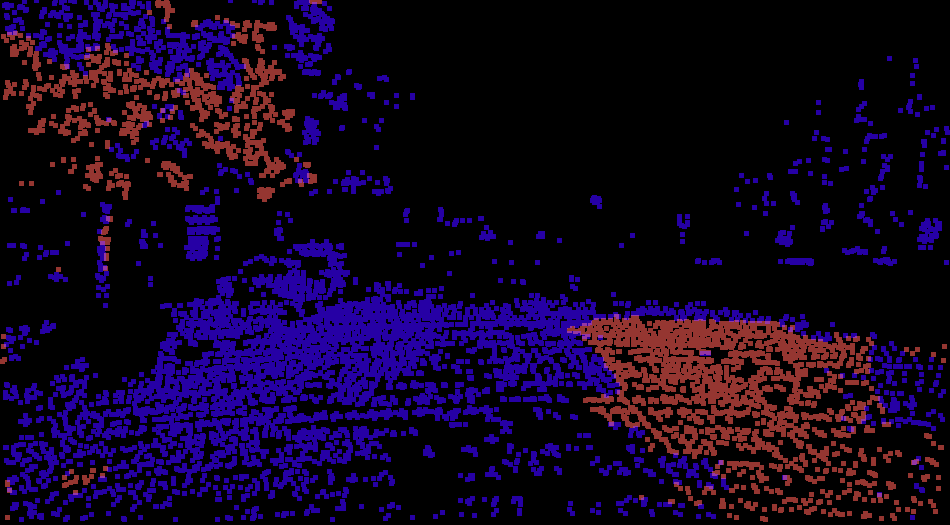}
          \put(5,2){\footnotesize\textcolor{white}{\textbf{Error Rate $>$ 3: 30.47\%}}}
        \end{overpic} &
        \begin{overpic}[trim=10cm 0cm 0cm 2cm, clip, width=0.185\linewidth]{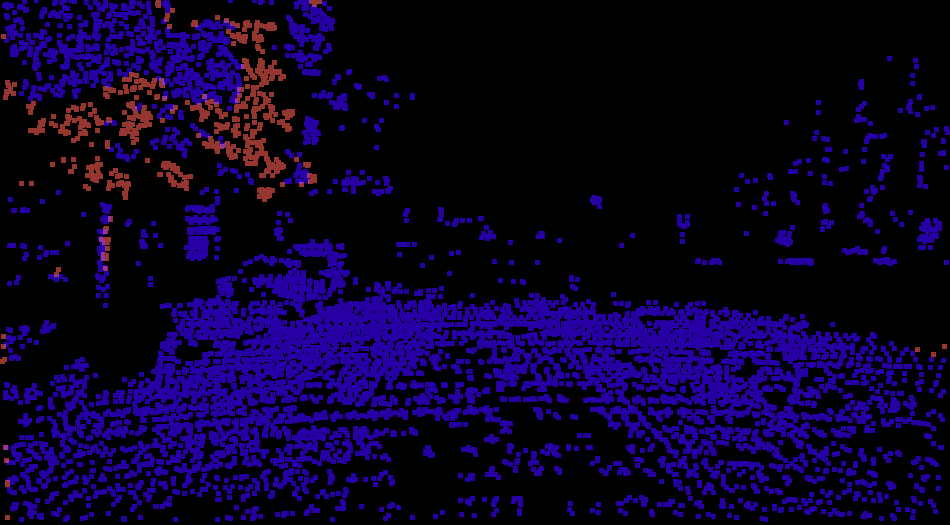}
          \put(5,2){\footnotesize\textcolor{white}{\textbf{Error Rate $>$ 3: 7.40\%}}}
        \end{overpic} & 
        \begin{overpic}[trim=10cm 0cm 0cm 2cm, clip, width=0.185\linewidth]{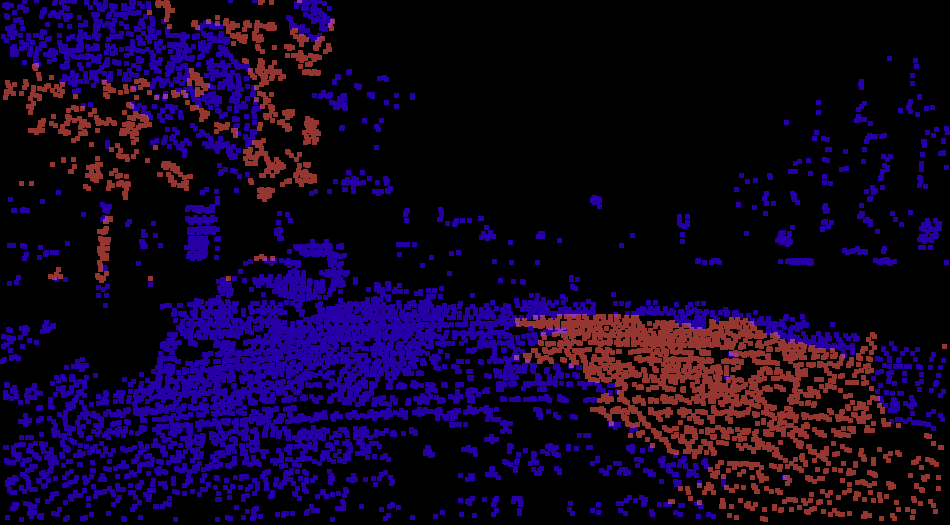} 
          \put(5,2){\footnotesize\textcolor{white}{\textbf{Error Rate $>$ 3: 30.84\%}}}
        \end{overpic} &
        \begin{overpic}[trim=10cm 0cm 0cm 2cm, clip, width=0.185\linewidth]{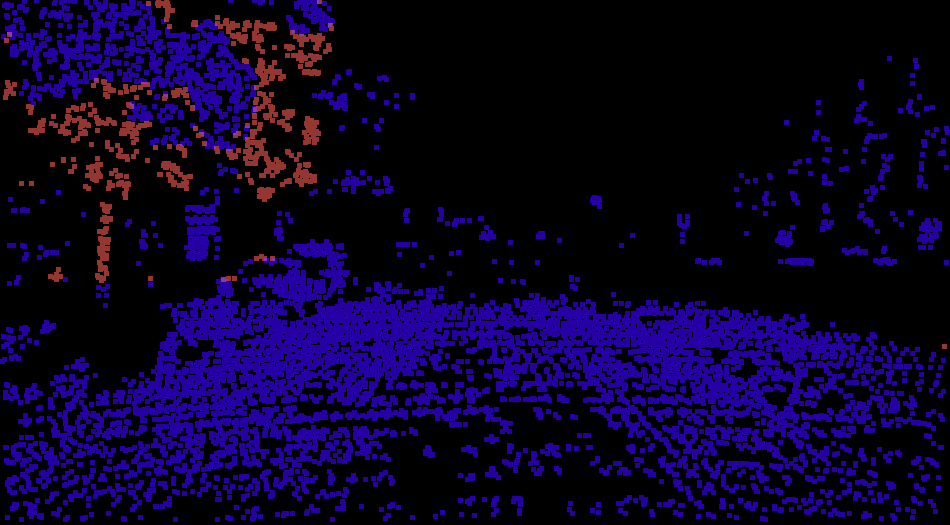} 
         \put(5,2){\footnotesize\textcolor{white}{\textbf{Error Rate $>$ 3: 8.54\%}}}
        \end{overpic}\vspace{0.1cm}
         
    \end{tabular}
    \caption{\textbf{Qualitative result on M3ED \citep{Chaney_2023_CVPR} day split.} Both ELFNet \citep{lou2023elfnet} and HITNet \citep{Tankovich_2021_CVPR} struggle with large texture-less areas, for example, created by scenes with high dynamic range. Our framework can leverage sparse depth points to solve this issue, whereas projected patterns would be ineffective.}
    \label{fig:qualitative_m3ed_day}
\end{figure*}

\subsection{VPP with More Off-the-shelf Networks}

Tab. \ref{tab:roundtable} collects results obtained deploying VPP with state-of-the-art stereo models \citep{lipson2021raft,yin2019hierarchical,Tankovich_2021_CVPR,cheng2020hierarchical,Shen_2021_CVPR,li2022practical,xu2022unifying,coexstereo,lou2023elfnet,shen2022pcw,zeng2023parameterized,zhao2023high,guan2024neural} taken off-the-shelf, running the weights provided by the authors. 
Again, VPP sensibly boosts the accuracy of any model with rare exceptions, either trained on synthetic or real data. Fig. \ref{fig:thin} shows the outcome with two networks trained on the same synthetic data: VPP guarantees a correct reconstruction of uniform areas while preserving fine details.

\subsection{Additional Evaluation with Raw Depth Data}

In addition to the previous evaluation with raw data on KITTI 2015, we extend our experiments with other unfiltered depth hints sourced from off-the-shelf LiDAR sensors using two additional datasets: DSEC \citep{Gehrig21ral} and M3ED \citep{Chaney_2023_CVPR}. 

\begin{table*}[t]
    \centering
    \scalebox{0.69}{
    \begin{tabular}{|c|c|cccc|c|cccc|c|cccc|c|}
        \multicolumn{2}{c}{} & \multicolumn{5}{c}{Vanilla Images} & \multicolumn{5}{c}{Active Stereo} & \multicolumn{5}{c}{VPP Stereo} \\
        \hline
         \multirow{2}{*}{Model} & & \multicolumn{4}{c|}{Error Rate (\%)} & avg. & \multicolumn{4}{c|}{Error Rate (\%)} & avg. & \multicolumn{4}{c|}{Error Rate (\%)} & avg. \\
        & Model name & $>1$ & $>2$ & $>3$ & $>4$ & (px) & $>1$ & $>2$ & $>3$ & $>4$ & (px) & $>1$ & $>2$ & $>3$ & $>4$ & (px) \\
        \hline\hline
        rSGM 
        & - & 39.88 & 26.79 & 20.04 & 15.87 & 2.88 & 11.32 & 6.51 & 5.14 & 4.45 & 1.11 & \bf 7.65 & \bf 4.44 & \bf 3.48 & \bf 3.00 & \bf 0.83 \\ 
        \hline\hline
        RAFT-Stereo 
        & Sceneflow & 29.18 & 19.82 & 14.13 & 10.60 & 1.61 & 7.83 & 3.96 & 2.76 & 2.18 & 0.51 & \bf 4.12 & \bf 2.34 & \bf 1.70 & \bf 1.37 & \bf 0.44 \\ 
        \hline\hline
        RAFT-Stereo 
        & Middlebury & 25.82 & 16.57 & 11.30 & 8.31 & 1.51 & 6.27 & 3.82 & 2.95 & 2.49 & 0.65 & \bf 4.20 & \bf 2.34 & \bf 1.67 & \bf 1.35 & \bf 0.48 \\
        \hline\hline
        RAFT-Stereo 
        & ETH3D & 28.16 & 18.78 & 13.73 & 10.83 & 1.68 & 7.64 & 3.82 & 2.68 & 2.13 & 0.50 & \bf 4.08 & \bf 2.33 & \bf 1.70 & \bf 1.37 & \bf 0.45 \\
        \hline\hline
        PSMNet 
        & Sceneflow & 30.46 & 19.53 & 14.69 & 12.06 & 2.53 & 8.15 & 3.62 & 2.47 & 1.94 & 0.67 & \bf 7.19 & \bf 2.92 & \bf 1.96 & \bf 1.55 & \bf 0.60 \\
        \hline\hline
        PSMNet 
        & Middlebury & 29.41 & 17.74 & 12.49 & 9.61 & 2.03 & 8.23 & 3.70 & 2.59 & 2.07 & 0.73 & \bf 7.76 & \bf 3.15 & \bf 2.06 & \bf 1.62 & \bf 0.58 \\ 
        \hline\hline
        GMStereo$^*$ 
        & Sceneflow & 33.34 & 21.24 & 14.83 & 10.82 & 1.67 & 7.22 & 3.24 & 2.21 & 1.71 & 0.55 & \bf 4.43 & \bf 2.18 & \bf 1.48 & \bf 1.15 & \bf 0.45 \\ 
        \hline\hline
        GMStereo$^*$ 
        & Mixdata & 22.29 & 11.98 & 7.48 & 5.11 & 1.08 & \bf 5.53 & 2.46 & 1.67 & 1.28 & \bf 0.54 & 6.43 & \bf 1.88 & \bf 1.21 & \bf 0.90 & 0.55 \\ 
        \hline\hline
        CFNet$^*$ 
        & Sceneflow & 30.20 & 20.04 & 14.53 & 11.35 & 1.84 & 10.88 & 8.02 & 6.95 & 6.36 & 2.78 & \bf 5.37 & \bf 3.16 & \bf 2.37 & \bf 1.93 & \bf 0.60 \\
        \hline\hline
        CFNet$^*$ 
        & Middlebury & 27.57 & 16.76 & 11.39 & 8.35 & 1.42 & 6.34 & 3.87 & 2.94 & 2.44 & \bf 0.58 & \bf 5.42 & \bf 3.29 & \bf 2.49 & \bf 2.04 & 0.59 \\ 
        \hline\hline
        HSMNet 
        & Middlebury & 29.90 & 15.73 & 10.61 & 7.80 & 1.44 & 17.82 & 7.08 & 4.27 & 3.08 & 0.92 & \bf 13.79 & \bf 5.46 & \bf 3.33 & \bf 2.41 & \bf 0.77 \\
        \hline\hline
        CREStereo$^*$ 
        & ETH3D & 23.16 & 14.40 & 9.31 & 6.20 & 1.07 & 4.14 & 2.38 & 1.74 & 1.41 & \bf 0.35 & \bf 3.34 & \bf 1.88 & \bf 1.35 & \bf 1.07 & 0.36 \\
        \hline\hline
        {LEAStereo} 
        & Sceneflow & 33.40 & 21.44 & 16.20 & 13.20 & 2.90 & 11.82 & 7.16 & 5.70 & 5.03 & 1.32 & \bf 8.48 & \bf 5.54 & \bf 4.70 & \bf 4.29 & \bf 1.21 \\
        \hline\hline
        {LEAStereo} 
        & KITTI12 & 37.13 & 25.26 & 20.09 & 17.40 & 3.75 & 10.79 & 6.64 & 5.40 & 4.72 & 0.99 & \bf 9.60 & \bf 5.63 & \bf 4.50 & \bf 3.93 & \bf 0.97 \\ 
        \hline\hline
        HITNet$^*$ 
        & Sceneflow & 27.20 & 18.16 & 14.03 & 11.34 & 1.82 & 8.55 & 4.38 & 3.21 & 2.62 & 0.67 & \bf 5.85 & \bf 3.00 & \bf 2.14 & \bf 1.74 & \bf 0.49 \\
        \hline\hline
        CoEx$^*$ 
        & Sceneflow & 29.17 & 17.43 & 12.66 & 10.03 & 1.98 & 15.76 & 7.82 & 5.55 & 4.49 & 1.15 & \bf 8.80 & \bf 3.39 & \bf 2.19 & \bf 1.67 & \bf 0.60 \\
        \hline\hline
        ELFNet$^*$ 
        & Sceneflow & 36.38 & 26.42 & 21.92 & 19.25 & 4.80 & 14.65 & 9.97 & 8.15 & 7.20 & 1.78 & \bf 9.46 & \bf 6.64 & \bf 5.67 & \bf 5.13 & \bf 1.34 \\ 
        \hline\hline
        PCWNet$^*$ 
        & Sceneflow & 26.51 & 16.95 & 12.94 & 10.29 & 1.97 & 11.12 & 6.31 & 4.89 & 4.18 & 1.07 & \bf 5.37 & \bf 2.56 & \bf 1.79 & \bf 1.44 & \bf 0.50 \\ 
        \hline\hline
        PCWNet$^*$ 
        & KITTI & 31.57 & 21.88 & 17.99 & 15.69 & 3.04 & 9.62 & 6.38 & 5.27 & 4.64 & 0.95 & \bf 8.73 & \bf 5.49 & \bf 4.08 & \bf 3.42 & \bf 0.84 \\ 
        \hline\hline
        {PCVNet} 
        & Sceneflow & 34.09 & 24.32 & 18.48 & 14.72 & 2.04 & 8.00 & 4.14 & 2.87 & 2.20 & 0.55 & \bf 4.99 & \bf 2.62 & \bf 1.83 & \bf 1.44 & \bf 0.47 \\ 
        \hline\hline
        DLNR 
        & Middlebury & 28.19 & 19.37 & 14.58 & 11.83 & 2.57 & 27.44 & 24.58 & 23.07 & 21.93 & 8.06 & \bf 5.10 & \bf 2.99 & \bf 2.13 & \bf 1.74 & \bf 0.55 \\ 
        \hline\hline
        NMRF$^*$ 
        & Sceneflow & 29.84 & 20.31 & 14.99 & 11.34 & 1.53 & 5.73 & 2.88 & 2.08 & 1.71 & 0.43 & \bf 4.03 & \bf 2.27 & \bf 1.64 & \bf 1.34 & \bf 0.42 \\ 
        \hline\hline
        NMRF$^*$ 
        & KITTI & 28.58 & 18.85 & 14.21 & 11.58 & 2.28 & 8.07 & 5.34 & 4.45 & 4.00 & \bf 0.97 & \bf 6.76 & \bf 4.59 & \bf 3.85 & \bf 3.45 & 1.09 \\ 
        \hline
    \end{tabular}}
    \caption{\textbf{VPP versus Active Stereo.} Results on SIMSTEREO \citep{jospin2022active} dataset sampling 2\% of depth points from simulated noisy ground-truth \citep{handa2014benchmark}. Entries marked with $^*$ use $\alpha=0.2$ for blending.}
    \label{tab:simstereo_results}
\end{table*}
\begin{figure*}[t]
    \centering
    \renewcommand{\tabcolsep}{1pt}
    \begin{tabular}{ccccc}

        Active & VPP & DLNR & DLNR{\em -active} & DLNR{\em -vpp} \\ 
        
        \includegraphics[trim=0cm 0cm 0cm 0cm, clip, width=0.185\linewidth]{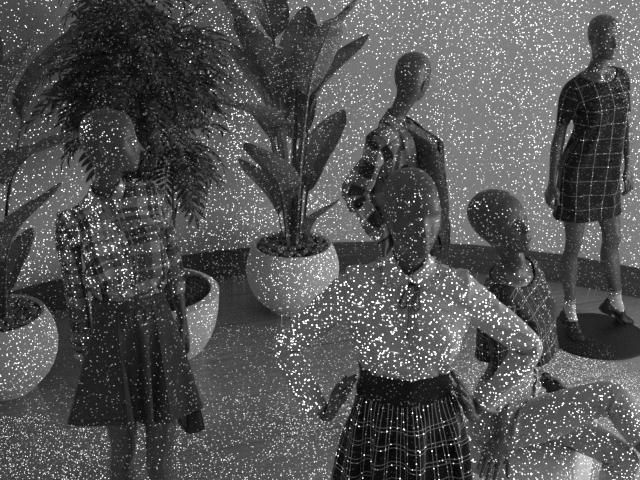} & 
        \includegraphics[trim=0cm 0cm 0cm 0cm, clip, width=0.185\linewidth]{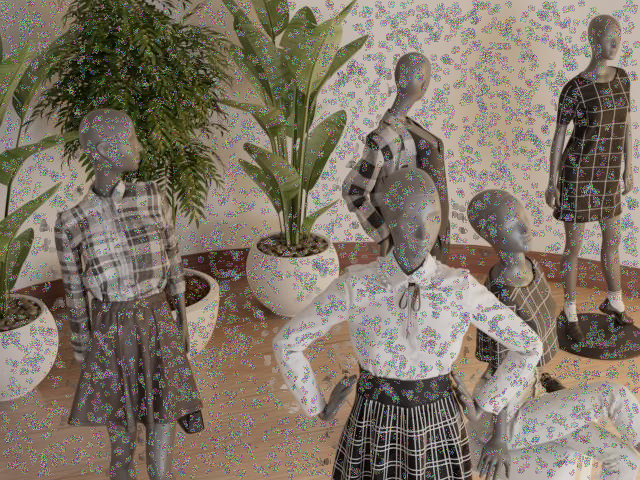} & 
        \includegraphics[trim=0cm 0cm 0cm 0cm, clip, width=0.185\linewidth]{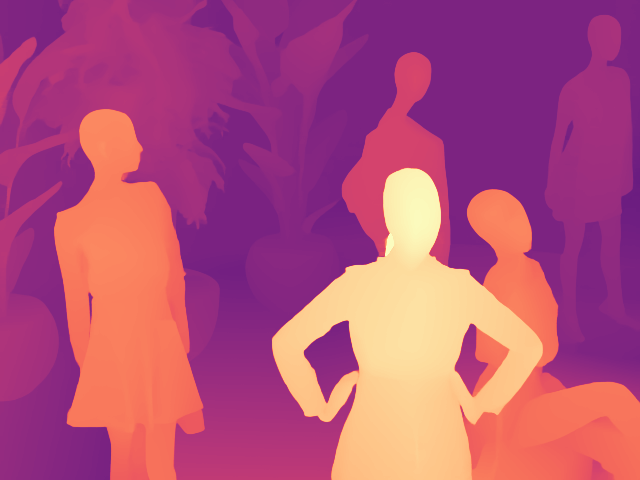} & 
        \includegraphics[trim=0cm 0cm 0cm 0cm, clip, width=0.185\linewidth]{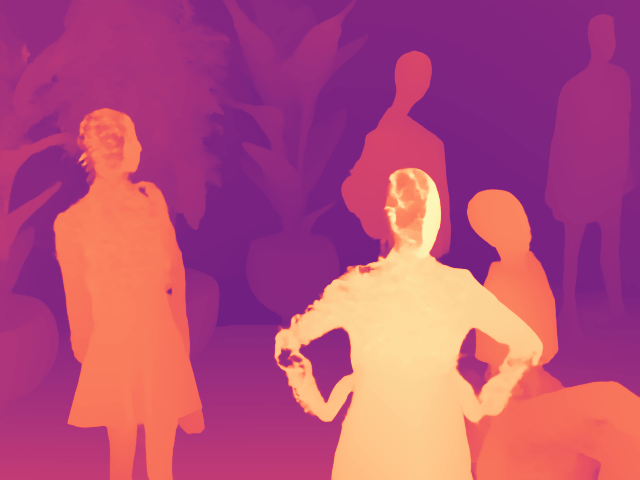} & 
        \includegraphics[trim=0cm 0cm 0cm 0cm, clip, width=0.185\linewidth]{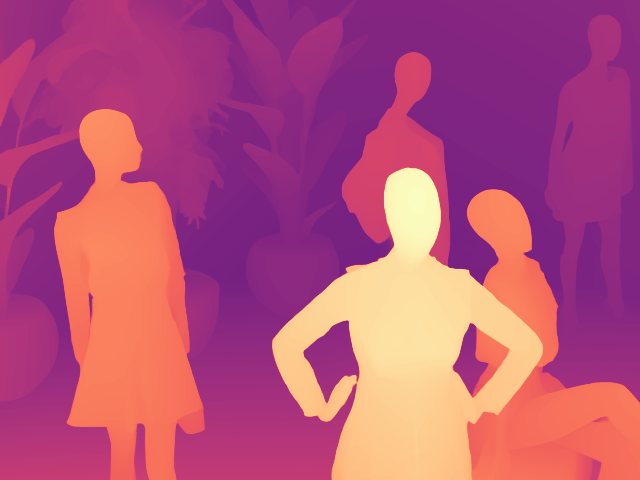} \\
        
        \begin{overpic}[trim=0cm 0cm 0cm 0cm, clip, width=0.185\linewidth]{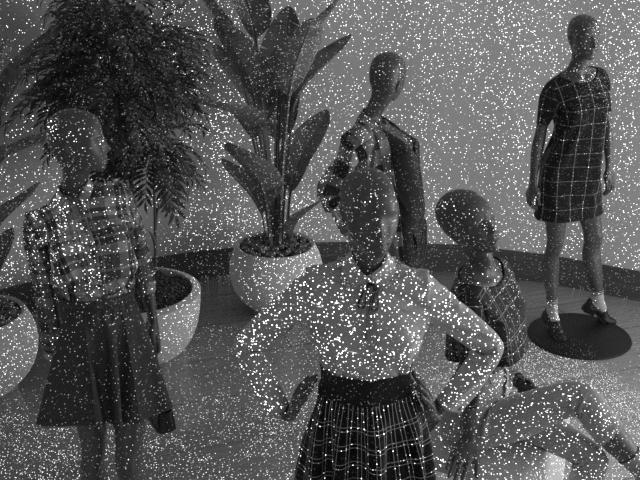}
        \end{overpic} &
        \begin{overpic}[trim=0cm 0cm 0cm 0cm, clip, width=0.185\linewidth]{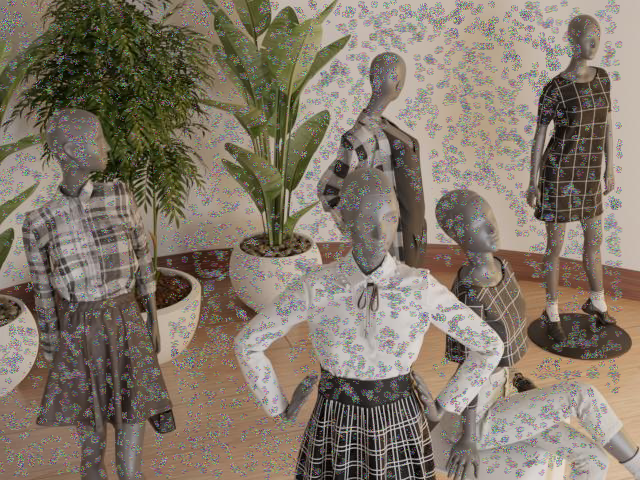}
        \end{overpic} &
        \begin{overpic}[trim=0cm 0cm 0cm 0cm, clip, width=0.185\linewidth]{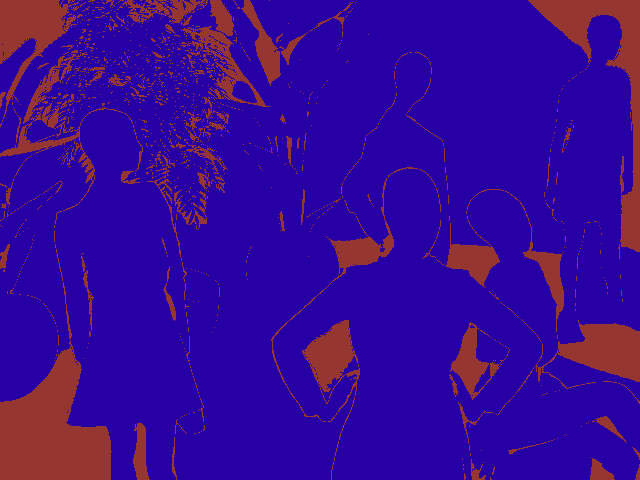}
          \put(5,2){\footnotesize\textcolor{white}{\textbf{Error Rate $>$ 1: 14.90\%}}}
        \end{overpic} & 
        \begin{overpic}[trim=0cm 0cm 0cm 0cm, clip, width=0.185\linewidth]{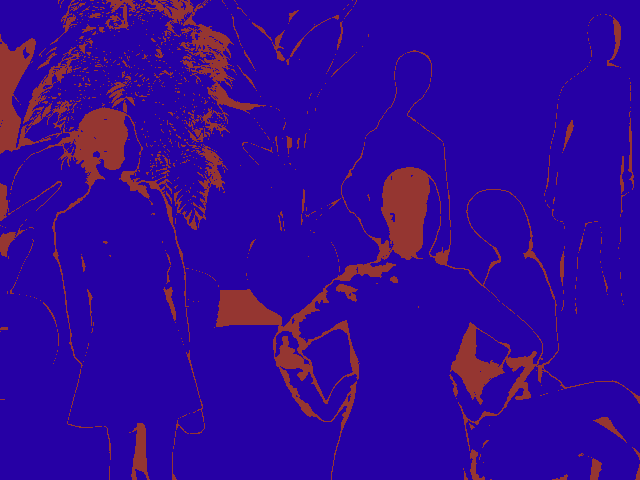}
          \put(5,2){\footnotesize\textcolor{white}{\textbf{Error Rate $>$ 1: 10.80\%}}}
        \end{overpic} &
        \begin{overpic}[trim=0cm 0cm 0cm 0cm, clip, width=0.185\linewidth]{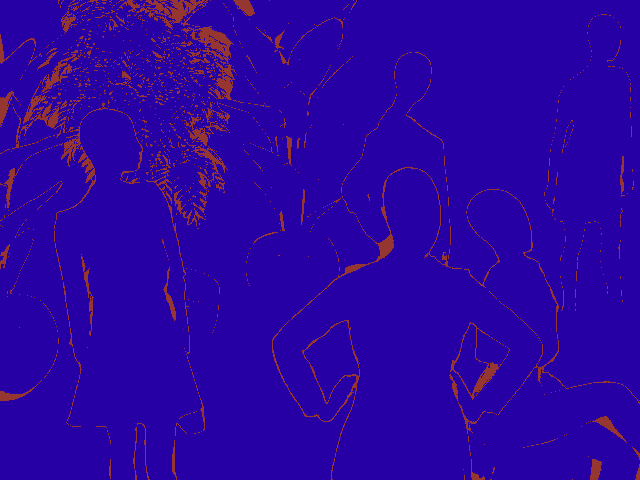}
         \put(5,2){\footnotesize\textcolor{white}{\textbf{Error Rate $>$ 1: 5.06\%}}}
        \end{overpic}\vspace{0.1cm}
         
    \end{tabular}
    \caption{\textbf{Qualitative result on SIMSTEREO \citep{jospin2022active}.} Some networks, such as DLNR \citep{zhao2023high}, often produce artefacts when fed with stereo pairs acquired with pattern projection active. In contrast, the same networks can seamlessly take advantage of our virtual patterns.}
    \label{fig:qualitative_simstereo}
\end{figure*}

\textbf{Analysis of Depth Hints.} Fig. \ref{fig:lidar_quality} plots various indicators of the quality of the sparse depth seeds and the datasets themselves: we measure a) the density of sparse depth hints, b) the mean absolute error of sparse hints compared to ground-truth data, and c) the disparity distribution of stereo frames.
In a), as expected, we appreciate how depth hints provided in DSEC are very sparse with -- a median density of 0.16\% -- while other datasets collected with 64-line LiDAR achieve a median of 1.80 and 4.00\% respectively (because of the different image resolution) beneficial to the effectiveness of VPP and other fusion frameworks as highlighted in Fig. \ref{fig:raft_density_curve}.
Interestingly, in b), we found a notable difference in the quality of raw depth hints across the datasets, ranging from a median error of 0.1m for M3ED to a median of 0.3m for KITTI 2015. 
In the remainder, our experiments will confirm 
that a higher density and depth quality translates into a higher performance boost.
Finally, c) exposes the significantly different disparity distributions characterizing M3ED with respect to KITTI and DSEC, mainly due to its shorter baseline.

\textbf{Results on DSEC.} Tab. \ref{tab:Results_DSEC} reports the performance by off-the-shelf networks and rSGM algorithm within the three DSEC splits -- Day, Afternoon, Night --, from easier (\ie, \textit{Day}) to harder (\ie, \textit{Night}) light conditions.
Even if the depth hints density is extremely low, VPP improves the accuracy of stereo networks in nearly all cases: the gain in accuracy is less appreciable in \textit{Day} split -- nonetheless, $>1$ error rate is decreased over 2\% in almost cases -- in contrast, as light conditions are getting worse (\ie, \textit{Afternoon}, \textit{Night}), our framework can reduce the very same error rate over 5\%, with some cases exciding 10\%.
Fig. \ref{fig:qualitative_dsec_night} exhibits a challenging frame from \textit{Night} split and predictions from LEAStereo \citep{cheng2020hierarchical} and PCWNet \citep{shen2022pcw} stereo networks: even with very few virtual patches, our solution allows to detect challenging obstacles such as the barely visible wall in the middle, that otherwise would have been missed processing the vanilla stereo pair.

\textbf{Results on M3ED.} Tab. \ref{tab:Results_M3ED} reports the accuracy by off-the-shelf networks and rSGM algorithm within two outdoor splits featuring different light conditions (\ie, \textit{Day}, \textit{Night}) and the \textit{Indoor} split in M3ED.
On the one hand, M3ED is slightly more challenging than DSEC due to a smaller baseline and open scenes without street lights. On the other hand, the depth hints are less noisy and 10 times denser than in the DSEC dataset; as a result, the overall improvement given by our proposal is an order of magnitude higher than in DSEC.
Specifically, although M3ED is a very challenging dataset -- \ie, $>1$ error rate is, in most cases, over 40\% for \textit{Outdoor Day} split and over 70\% for \textit{Outdoor Night} split -- when stereo networks and rSGM algorithm are used in combination with VPP we observe an outstanding improvement in any metrics: all models benefit from our framework, with $>1$ error rate reduction respectively over 30\% for \textit{Outdoor Day} split and 50\% for \textit{Outdoor Night} split.
Also, the \textit{Indoor} split results confirm the previous trend, decreasing the $>1$ error rate by more than 20\% in all cases.
Fig. \ref{fig:qualitative_m3ed_day} plots a frame from \textit{Outdoor Day} with the predictions by ELFNet \citep{lou2023elfnet} and HITNet \citep{Tankovich_2021_CVPR} stereo networks alone or when assisted by VPP, clearly confirming the benefits yielded by our method in low-textured regions.

\begin{table*}[t]
    \centering
    \scalebox{0.67}{
        \begin{tabular}{|c|c|cccc|c|cccc|c|cccc|c|}
            \multicolumn{2}{c}{} & \multicolumn{5}{c}{Vanilla Images} & \multicolumn{5}{c}{Active Stereo} & \multicolumn{5}{c}{VPP Stereo} \\
            \hline
             \multirow{2}{*}{Model} & & \multicolumn{4}{c|}{Error Rate (\%)} & avg. & \multicolumn{4}{c|}{Error Rate (\%)} & avg. & \multicolumn{4}{c|}{Error Rate (\%)} & avg. \\
            & Model name & $>1$ & $>2$ & $>3$ & $>4$ & (px) & $>1$ & $>2$ & $>3$ & $>4$ & (px) & $>1$ & $>2$ & $>3$ & $>4$ & (px) \\
            \hline\hline
            rSGM 
            & - & 73.13 & 57.74 & 49.52 & 44.00 & 15.36 & 64.10 & 44.68 & 34.27 & 27.80 & 7.77 & \bf 33.69 & \bf 18.82 & \bf 12.06 & \bf 9.09 & \bf 2.38 \\ 
            \hline\hline
            {RAFT-Stereo} 
            & Sceneflow & 63.31 & 44.48 & 34.25 & 27.21 & 4.69 & 54.89 & 36.34 & 26.18 & 19.63 & 3.50 & \bf 34.92 & \bf 20.54 & \bf 13.53 & \bf 10.36 & \bf 2.18 \\ 
            \hline\hline
            RAFT-Stereo 
            & Middlebury & 59.29 & 39.70 & 29.54 & 23.06 & 4.49 & 52.26 & 34.21 & 24.16 & 17.70 & 3.30 & \bf 33.76 & \bf 19.23 & \bf 12.35 & \bf 9.38 & \bf 2.10 \\
            \hline\hline
            RAFT-Stereo 
            & ETH3D & 59.78 & 41.32 & 32.11 & 25.79 & 4.65 & 53.90 & 35.82 & 25.85 & 19.35 & 3.47 & \bf 34.65 & \bf 20.29 & \bf 13.34 & \bf 10.21 & \bf 2.18 \\ 
            \hline\hline
            PSMNet 
            & Sceneflow & 69.28 & 52.23 & 43.09 & 36.70 & 8.33 & 57.58 & 39.19 & 28.98 & 22.37 & 4.37 & \bf 35.87 & \bf 20.92 & \bf 13.82 & \bf 10.45 & \bf 2.61 \\
            \hline\hline
            PSMNet 
            & Middlebury & 67.99 & 50.97 & 41.78 & 35.47 & 7.38 & 55.58 & 37.49 & 27.31 & 20.82 & 4.24 & \bf 36.53 & \bf 21.78 & \bf 14.47 & \bf 10.87 & \bf 2.65 \\
            \hline\hline
            GMStereo$^*$ 
            & Sceneflow & 66.81 & 50.59 & 41.90 & 35.69 & 7.10 & 55.50 & 37.46 & 27.63 & 21.75 & 4.05 & \bf 36.96 & \bf 21.77 & \bf 14.44 & \bf 11.09 & \bf 2.37 \\  
            \hline\hline
            GMStereo$^*$ 
            & Mixdata & 58.58 & 40.71 & 30.50 & 23.73 & 4.21 & 50.18 & 33.45 & 23.72 & 17.56 & 3.28 & \bf 34.65 & \bf 19.45 & \bf 12.57 & \bf 9.62 & \bf 2.12 \\   
            \hline\hline
            CFNet$^*$ 
            & Sceneflow & 68.13 & 51.38 & 43.01 & 37.45 & 15.91 & 60.51 & 42.21 & 32.69 & 26.64 & 8.14 & \bf 35.26 & \bf 20.22 & \bf 13.09 & \bf 9.79 & \bf 2.20 \\ 
            \hline\hline
            CFNet$^*$ 
            & Middlebury & 63.30 & 46.35 & 36.69 & 29.93 & 5.62 & 51.72 & 33.73 & 23.55 & 17.06 & 3.19 & \bf 33.10 & \bf 18.55 & \bf 11.90 & \bf 9.00 & \bf 2.03 \\ 
            \hline\hline
            HSMNet
            & Middlebury & 67.60 & 49.39 & 38.60 & 31.35 & 5.48 & 57.69 & 38.15 & 27.00 & 20.10 & 4.02 & \bf 40.83 & \bf 22.73 & \bf 14.23 & \bf 10.31 & \bf 2.55 \\ 
            \hline\hline
            CREStereo$^*$ 
            & ETH3D & 63.18 & 44.57 & 34.00 & 26.92 & 4.38 & 52.78 & 34.65 & 24.54 & 18.06 & 3.37 & \bf 35.76 & \bf 20.63 & \bf 13.10 & \bf 9.68 & \bf 2.14 \\ 
            \hline\hline
            {LEAStereo} 
            & Sceneflow & 70.74 & 55.10 & 46.96 & 41.29 & 12.60 & 64.25 & 43.72 & 32.52 & 25.77 & 5.50 & \bf 33.24 & \bf 17.92 & \bf 11.10 & \bf 8.26 & \bf 1.99 \\ 
            \hline\hline
            {LEAStereo}
            & KITTI12 & 67.58 & 51.18 & 42.46 & 36.68 & 10.16 & 56.76 & 37.77 & 27.21 & 20.51 & 3.99 & \bf 34.57 & \bf 19.45 & \bf 12.20 & \bf 8.84 & \bf 2.14 \\ 
            \hline\hline
            HITNet$^*$ 
            & Sceneflow & 68.03 & 51.94 & 43.60 & 37.62 & 10.72 & 56.38 & 38.39 & 28.35 & 21.65 & 3.89 & \bf 37.04 & \bf 21.44 & \bf 14.04 & \bf 10.65 & \bf 2.41 \\
            \hline\hline
            CoEx$^*$  
            & Sceneflow & 66.40 & 49.61 & 40.94 & 34.93 & 12.36 & 55.83 & 37.78 & 27.23 & 20.50 & 3.78 & \bf 37.91 & \bf 21.34 & \bf 13.99 & \bf 10.75 & \bf 2.46 \\ 
            \hline\hline
            ELFNet$^*$ 
            & Sceneflow & 68.21 & 53.20 & 45.48 & 39.95 & 13.65 & 59.25 & 42.26 & 33.12 & 27.48 & 8.15 & \bf 36.37 & \bf 21.37 & \bf 15.07 & \bf 12.36 & \bf 4.41 \\ 
            \hline\hline
            PCWNet$^*$ 
            & Sceneflow & 66.75 & 50.53 & 42.01 & 35.51 & 7.49 & 56.45 & 37.86 & 27.89 & 21.18 & 3.69 & \bf 35.69 & \bf 19.58 & \bf 12.23 & \bf 9.22 & \bf 2.11 \\ 
            \hline\hline
            PCWNet$^*$ 
            & KITTI & 65.56 & 47.77 & 38.33 & 31.99 & 7.23 & 54.20 & 34.65 & 23.67 & 16.61 & 3.12 & \bf 33.23 & \bf 16.97 & \bf 10.50 & \bf 7.84 & \bf 1.90 \\ 
            \hline\hline
            {PCVNet} 
            & Sceneflow & 66.36 & 50.55 & 41.60 & 35.61 & 17.40 & 56.58 & 39.00 & 28.88 & 22.35 & 4.13 & \bf 36.73 & \bf 21.66 & \bf 14.20 & \bf 10.74 & \bf 2.34 \\ 
            \hline\hline
            DLNR 
            & Middlebury & 57.44 & 39.14 & 28.87 & 22.12 & 4.05 & 53.09 & 35.83 & 25.80 & 19.13 & 3.53 & \bf 35.14 & \bf 20.15 & \bf 12.73 & \bf 9.52 & \bf 2.17 \\  
            \hline\hline
            NMRF$^*$ 
            & Sceneflow & 63.03 & 45.08 & 35.59 & 28.87 & 5.16 & 54.34 & 36.31 & 26.36 & 20.23 & 3.91 & \bf 36.04 & \bf 21.09 & \bf 14.00 & \bf 10.76 & \bf 2.30 \\ 
            \hline\hline
            NMRF$^*$ 
            & KITTI & 61.86 & 45.93 & 37.32 & 31.46 & 6.99 & 51.14 & 33.21 & 23.45 & 17.29 & 3.28 & \bf 34.99 & \bf 19.24 & \bf 12.43 & \bf 9.40 & \bf 2.11 \\ 
            \hline
        \end{tabular}
    }
    \caption{\textbf{VPP versus Active Stereo.} Results on M3ED-active \citep{Chaney_2023_CVPR} split using from 64-line LiDAR. Entries marked with $^*$ use $\alpha=0.2$ for blending.}
    \label{tab:m3ed_active_results}
\end{table*}

\begin{figure*}[t]
    \centering
    \renewcommand{\tabcolsep}{1pt}
    \begin{tabular}{ccccc}

        Active & VPP & RAFT-Stereo & RAFT-Stereo{\em -active} & RAFT-Stereo{\em -vpp} \\ 
        
        \includegraphics[trim=0cm 0cm 0cm 0cm, clip, width=0.185\linewidth]{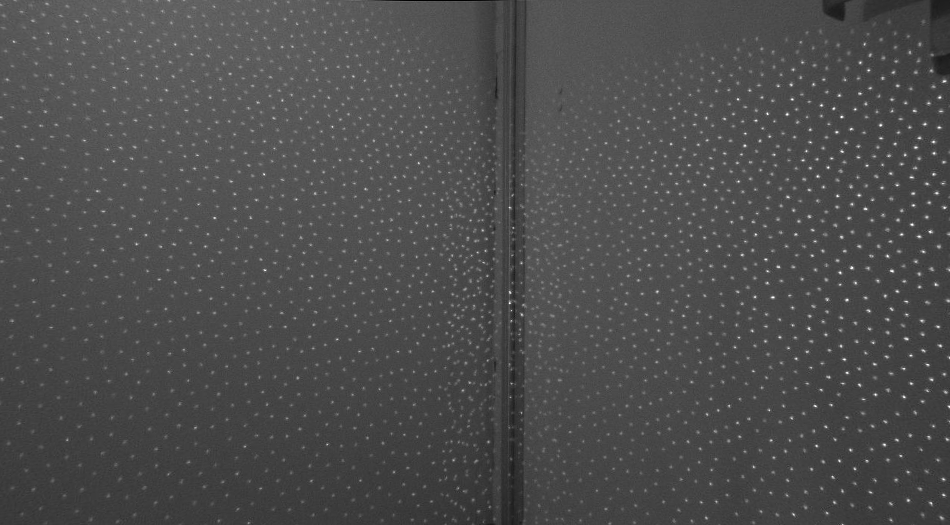} & 
        \includegraphics[trim=0cm 0cm 0cm 0cm, clip, width=0.185\linewidth]{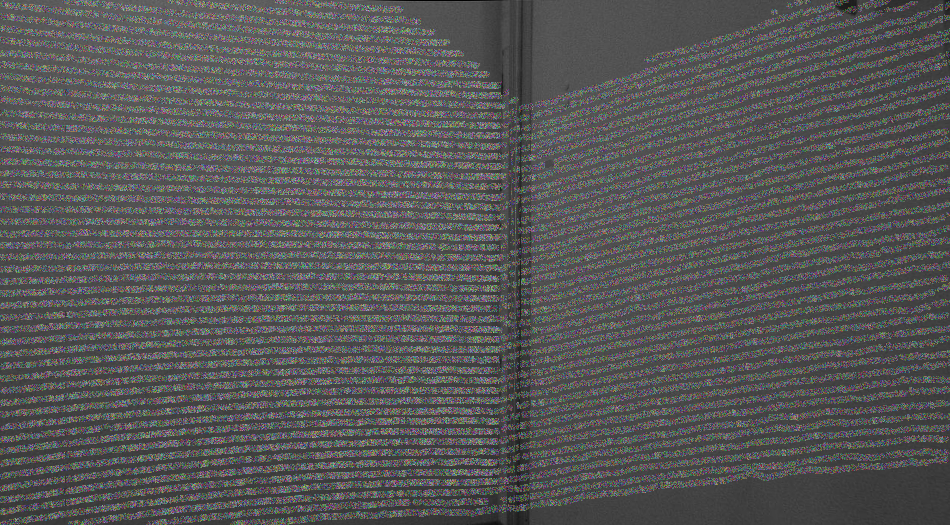} & 
        \includegraphics[trim=0cm 0cm 0cm 0cm, clip, width=0.185\linewidth]{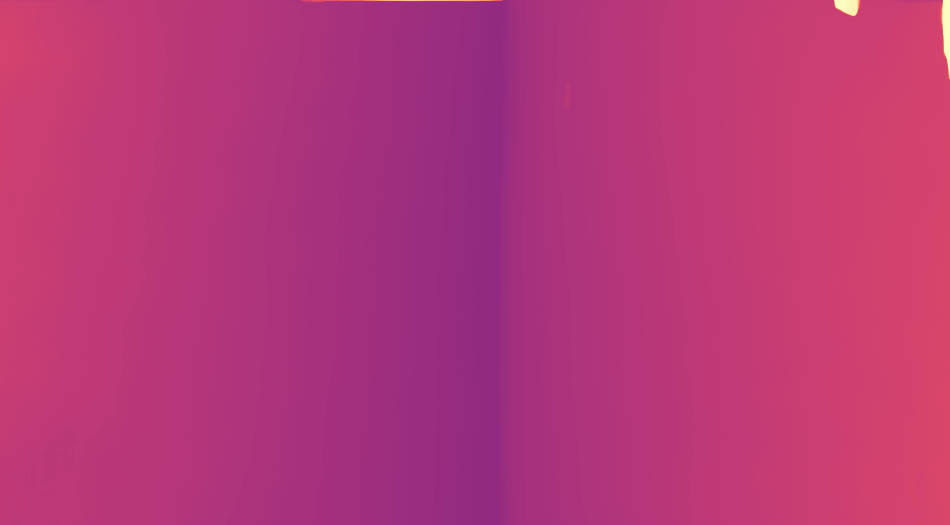} & 
        \includegraphics[trim=0cm 0cm 0cm 0cm, clip, width=0.185\linewidth]{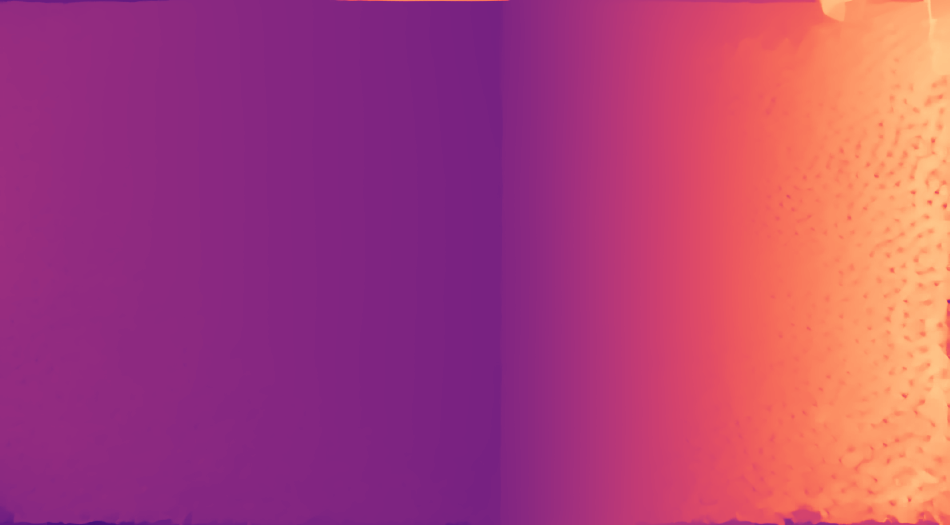} & 
        \includegraphics[trim=0cm 0cm 0cm 0cm, clip, width=0.185\linewidth]{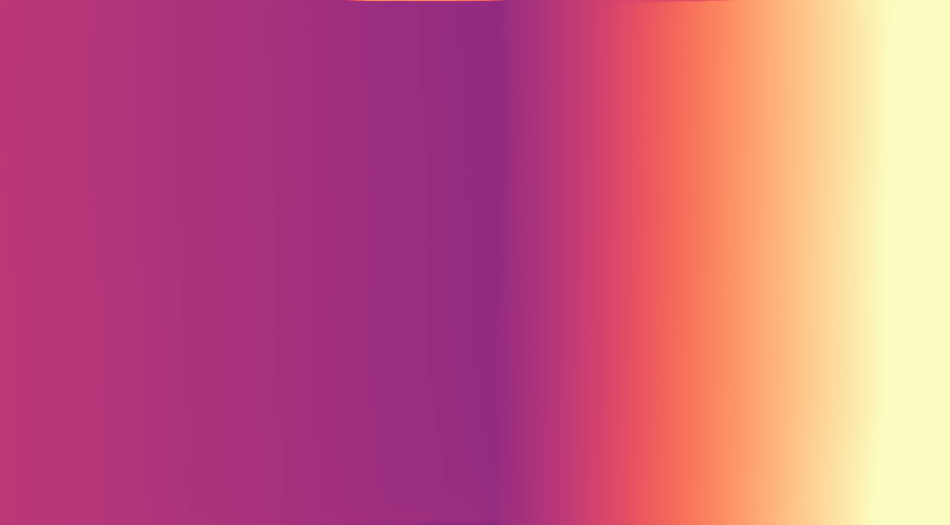} \\
        
        \begin{overpic}[trim=0cm 0cm 0cm 0cm, clip, width=0.185\linewidth]{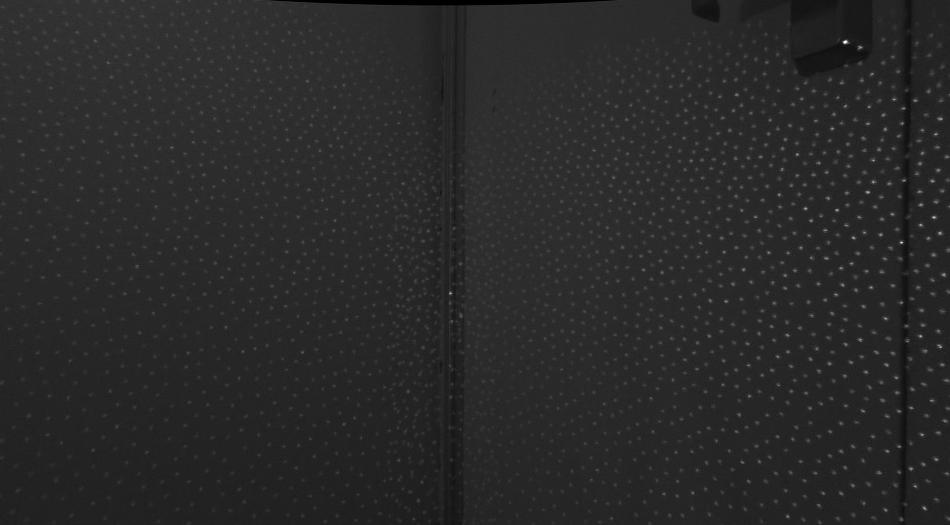}
        \end{overpic} &
        \begin{overpic}[trim=0cm 0cm 0cm 0cm, clip, width=0.185\linewidth]{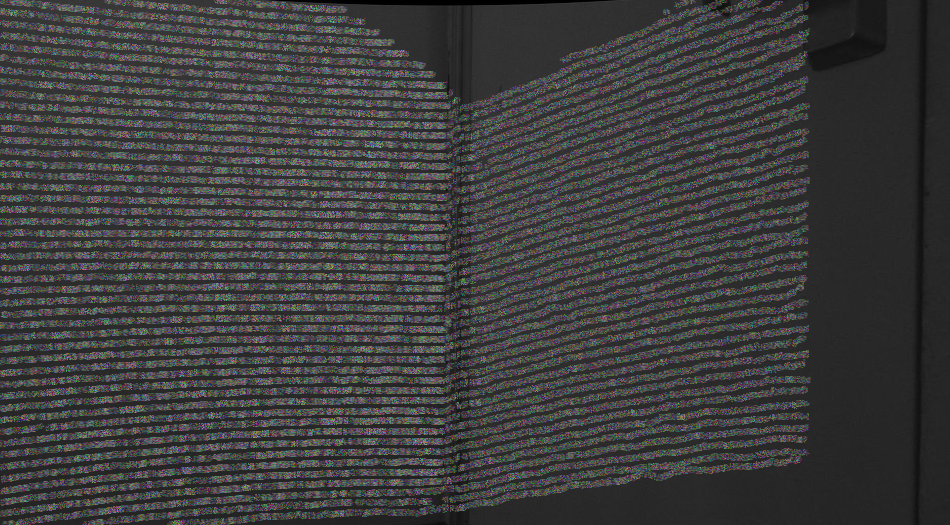}
        \end{overpic} &
        \begin{overpic}[trim=0cm 0cm 0cm 0cm, clip, width=0.185\linewidth]{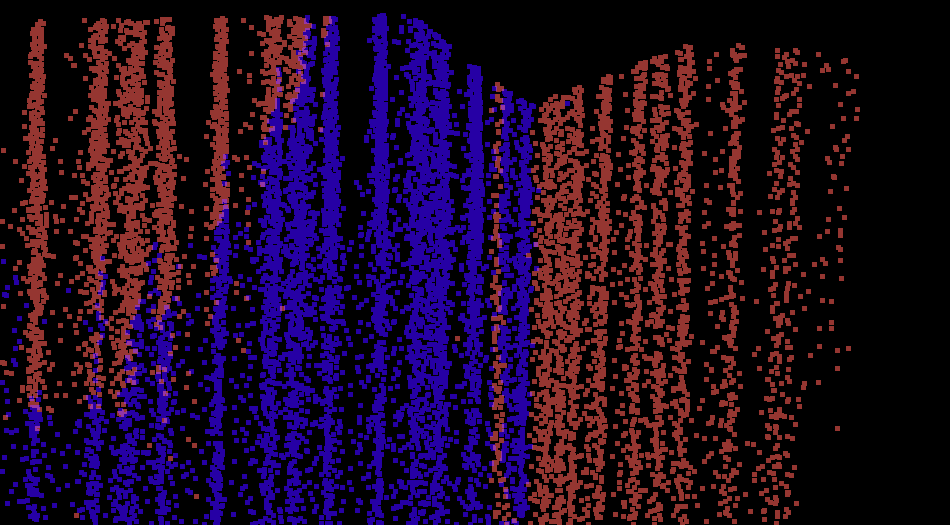}
          \put(5,2){\footnotesize\textcolor{white}{\textbf{Error Rate $>$ 3: 47.68\%}}}
        \end{overpic} & 
        \begin{overpic}[trim=0cm 0cm 0cm 0cm, clip, width=0.185\linewidth]{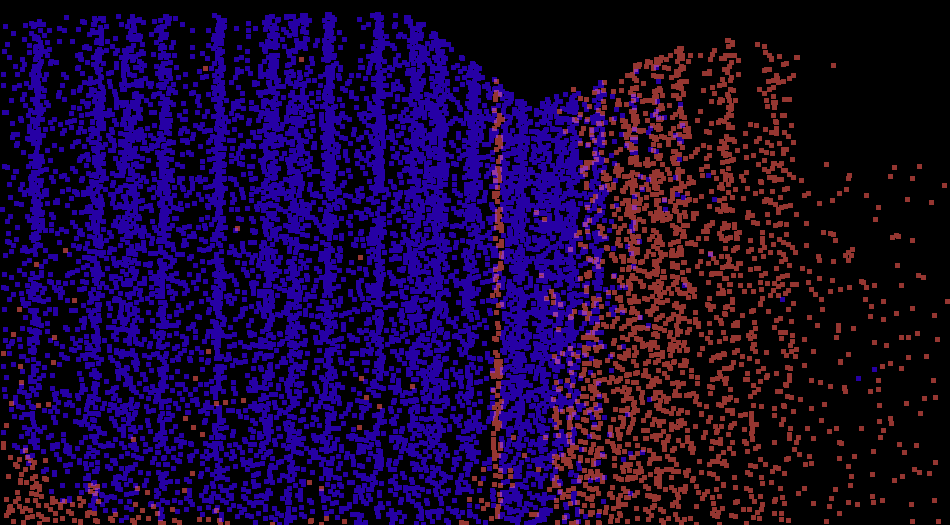}
          \put(5,2){\footnotesize\textcolor{white}{\textbf{Error Rate $>$ 3: 21.61\%}}}
        \end{overpic} &
        \begin{overpic}[trim=0cm 0cm 0cm 0cm, clip, width=0.185\linewidth]{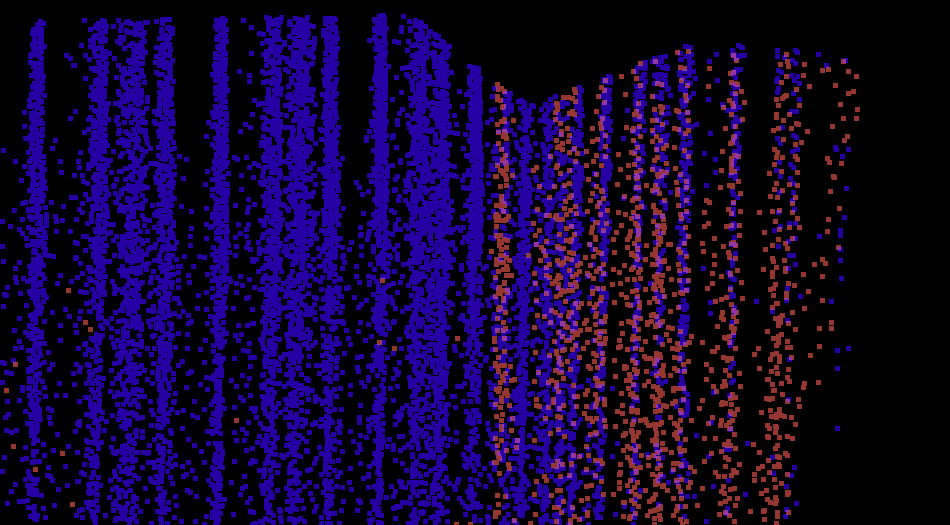}
         \put(5,2){\footnotesize\textcolor{white}{\textbf{Error Rate $>$ 3: 13.93\%}}}
        \end{overpic}\vspace{0.1cm}
         
    \end{tabular}
    \caption{\textbf{Qualitative result on M3ED-active \citep{Chaney_2023_CVPR} .} Virtual and physical projected patterns help to improve the accuracy of a recent stereo network (\ie, RAFT-Stereo \citep{lipson2021raft}) when dealing with large uniform areas.}
    \label{fig:qualitative_m3ed_active}
\end{figure*}

\subsection{Comparison with Active Stereo}

To conclude, we embark on a quantitative comparison between the effectiveness of VPP against the traditional active stereo setup -- typically composed of an IR stereo camera and an IR projector -- using one synthetic dataset \citep{jospin2022active} and the M3ED-active \citep{Chaney_2023_CVPR} split.

\textbf{Results on SIMSTEREO.} Tab. \ref{tab:simstereo_results} collects our findings in an ideal scenario where IR and RGB cameras are perfectly aligned -- \ie, RGB and IR images frame the very same scene, for which a dense, synthetic ground-truth disparity map is provided: as expected, the IR active pattern is beneficial in reducing error metrics, with only a few exceptions -- CFNet \citep{Shen_2021_CVPR} and DLNR \citep{zhao2023high}; this finding is consistent with those by \cite{jospin2022active}, where they propose to adapt networks to the active pattern via fine-tuning.
In contrast, our framework can be tuned (\eg, using a different amount of alpha-blending) to these networks without any additional training.

Furthermore, our proposal addresses different issues regarding active stereo \citep{grunnet2018projectors}: firstly, traditional active stereo systems use IR spectrum to not interfere with human vision and consequently require an additional RGB camera to capture visible spectrum, while VPP can directly exploit RGB spectrum to create more complex patterns; secondly, the IR projected pattern fades proportionally to the square of the distance and it suffers from strong ambient lights, whereas our proposal completely bypasses these issues.
In addition to these advantages, our framework generally performs better than active stereo, sometimes halving the error metrics.
Fig. \ref{fig:qualitative_simstereo} visually shows the difference between traditional IR projected patterns and the virtual patterns on the same frame, followed by predictions using DLNR \citep{zhao2023high} stereo network. Feeding the network with traditionally patterned images reduces the $>1$ error compared to passive stereo yet introduces some notable artifacts. In contrast, feeding the same network with virtually patterned frames, most artifacts disappear and the error metric is more than halved compared to active stereo. 

\textbf{Results on M3ED-active.} To further support our findings even on real data, we extend the previous experiment to the \textit{Passive} and \textit{Active} splits from M3ED \citep{Chaney_2023_CVPR}. 
We recall that, although the images on the two splits do not coincide exactly -- i.e., the active pattern is visible only once every two frames, and the robot is moving during acquisition -- we assume a negligible motion occurs between adjacent passive/active frames, because of the very short time delay between the two: accordingly, we feel a direct comparison between methods running on the two distinct splits is fair enough, although not perfect.
Tab. \ref{tab:m3ed_active_results} highlights that our framework surpasses traditional active pattering in all tested cases, halving all error metrics in most experiments.
Although active stereo already reduces most of the error metrics with respect to passive stereo, VPP confirms its undisputed superiority with dramatically larger improvements. 
Fig \ref{fig:qualitative_m3ed_active} shows a qualitative example with a large uniform area, texturized both by physically projected patterns (fisrt column) or VPP (second column): not surprisingly, passive stereo almost fails even with RAFT-Stereo \citep{lipson2021raft}, whereas both patterns help to reconstruct the walls correctly, with VPP resulting the absolute winner by a massive margin.

\section{Limitations and Failure Cases}

Our evaluation does not reveal noticeable failure cases or limitations. However, selecting the correct device to infer robust depth hints in the desired target domains is crucial, especially for safety-critical applications. In the worst-case scenario, if the depth sensor fails -- e.g., due to extreme environmental conditions, challenging object/material, or any other reason yielding sporadic or continuous breakdown of the deployed sensing technology -- the perception module would perform identically as the underlying stereo matcher, like a passive system without benefiting from virtual patterns. 
Finally, the sparse depth hints should be spread over the whole image for optimal performance, as for any other fusion method. Nonetheless, as the experimental analysis highlights, our proposal can seamlessly deal with very sparse, fluctuating, and noisy depth hints.

\section{Conclusion}

This paper proposes a novel paradigm for combining stereo with sparse depth hints. Inspired by active stereo but unaffected by its inherent severe limitations, our Virtual Pattern Projection (VPP) exploits such depth hints to generate coherently hallucinated stereo pairs to facilitate visual correspondence.
Extensive experiments on several standard stereo datasets prove that VPP outperforms existing stereo-depth fusion techniques, works seamlessly with any state-of-the-art deep stereo network and algorithm, and is more robust and effective than conventional active stereo technologies. Additionally, it remarkably increases the cross-domain generalization ability of deep stereo networks.
In light of this evidence, our paradigm has the potential to become a standard component in the perception pipelines of several applications, ranging from autonomous driving to robotics.

{\section*{Acknowledgements} 
This study was carried out within the MOST – Sustainable Mobility National Research Center and received funding from the European Union Next-GenerationEU – PIANO NAZIONALE DI RIPRESA E RESILIENZA (PNRR) – MISSIONE 4 COMPONENTE 2, INVESTIMENTO 1.4 – D.D. 1033 17/06/2022, CN00000023. This manuscript reflects only the authors’ views and opinions, neither the European Union nor the European Commission can be considered responsible for them.

We acknowledge the CINECA award under the ISCRA initiative, for the availability of high-performance computing resources and support.} 

\section*{Data Availability Statement}
The datasets and benchmark results supporting the findings of this study are publicly available. The KITTI 2012, 2015 and derived datasets can be accessed at \url{https://www.cvlibs.net/datasets/kitti/eval_stereo.php}. The Middlebury 2014 and 2021 datasets are available at \url{https://vision.middlebury.edu/stereo/data/}. 
The ETH3D dataset can be downloaded at \url{https://www.eth3d.net/datasets}.
The DSEC dataset can be accessed at \url{https://dsec.ifi.uzh.ch/}. 
The M3ED dataset is available at \url{https://m3ed.io/}. The SIMSTEREO dataset can be accessed at \url{https://ieee-dataport.org/open-access/active-passive-simstereo}.

All the pre-processed data used in our experiments are also available at \url{https://github.com/bartn8/vppstereo/?tab=readme-ov-file#floppy_disk-datasets}.

\bibliography{egbib}

\end{document}